\documentclass{article}

    \PassOptionsToPackage{numbers, compress}{natbib}

 \usepackage[preprint]{neurips_2026}

\usepackage[utf8]{inputenc} 
\usepackage[T1]{fontenc}    
\usepackage{hyperref}       
\usepackage{url}            
\usepackage{booktabs}       
\usepackage{amsfonts}       
\usepackage{nicefrac}       
\usepackage{microtype}      
\usepackage{xcolor}         

\usepackage[table]{xcolor}
\usepackage{adjustbox}
\usepackage{titletoc}
\usepackage{amssymb}
\usepackage{multirow}
\usepackage{makecell}
\usepackage{amsmath}
\usepackage{caption}
\usepackage{subcaption}
\usepackage{wrapfig}
\usepackage{enumitem}
\usepackage{float}

\title{VGenST-Bench: A Benchmark for Spatio-Temporal Reasoning via Active Video Synthesis}

\author{%
  Jinho Park$^{1}$ \quad Youbin Kim$^{1}$ \quad Hogun Park$^{1}$ \quad Eunbyung Park$^{2\dagger}$ \\[0.5em]
  $^{1}$Department of Artificial Intelligence, Sungkyunkwan University \\
  $^{2}$Department of Artificial Intelligence, Yonsei University \\
  $^{\dagger}$Corresponding author \\[0.5em]    
  \url{https://zinosii.github.io/VGenST-Bench/} \\
}

\begin{document}

\maketitle

\begin{figure}[h]

\vspace{-4ex}

\centering
\includegraphics[width=\textwidth]{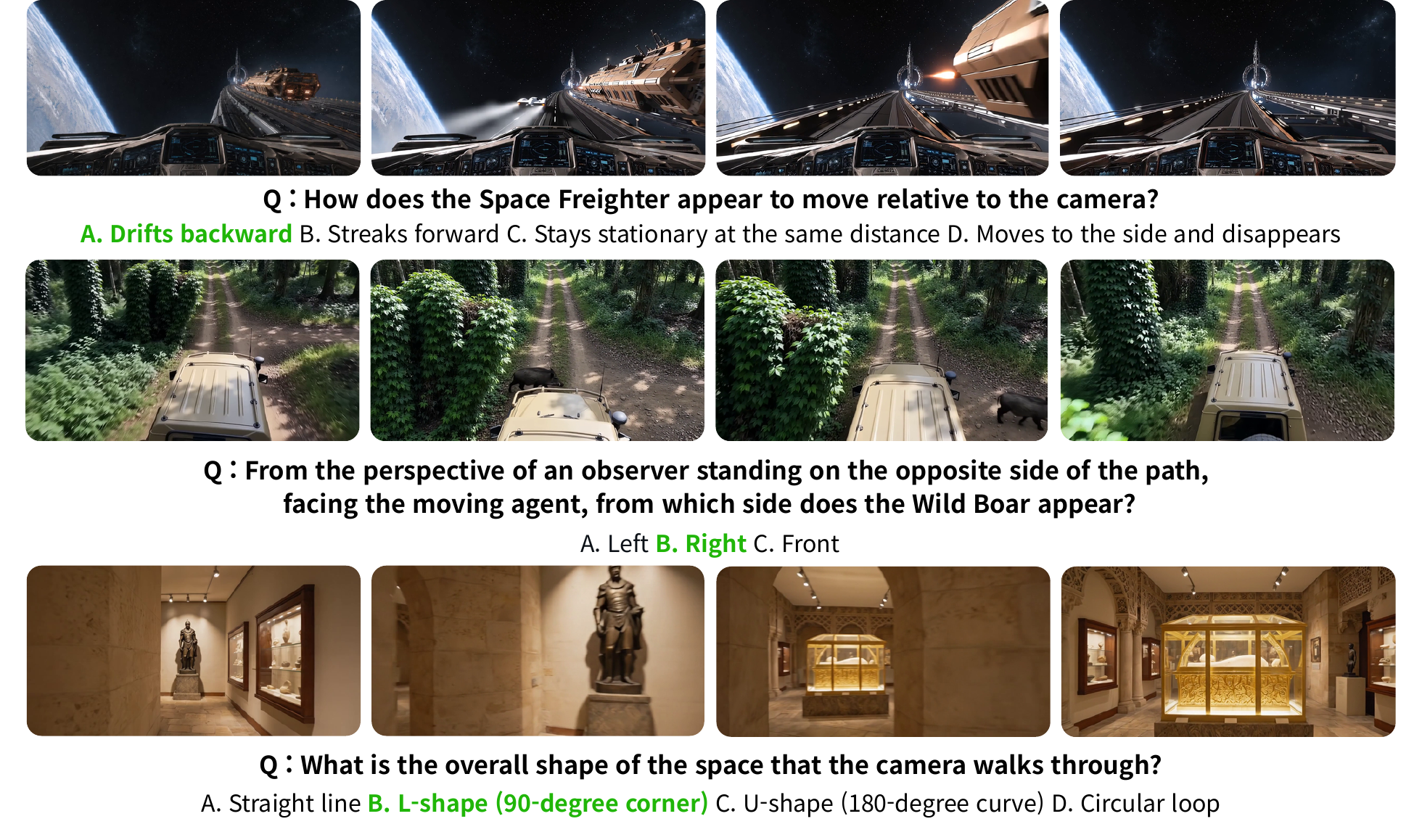}
\vspace{-4ex}
\caption{\textbf{Examples of VGenST-Bench.} 
Each example contains a generated video and a multiple-choice question targeting a specific spatio-temporal reasoning. Correct answers are highlighted.}

\label{fig:teaser}
\end{figure}

\begin{abstract}
 Spatio-temporal reasoning is a core capability for Multimodal Large Language Models (MLLMs) operating in the real world. As such, evaluating it precisely has become an essential challenge. However, existing spatio-temporal reasoning benchmark datasets primarily rely on static image sets or passively curated video data, which limits the evaluation of fine-grained reasoning capabilities. In this paper, we introduce \textbf{VGenST-Bench}, a video benchmark that employs generative models to actively synthesize highly controlled and diverse evaluation scenarios. To construct VGenST-Bench, we propose a multi-agent pipeline incorporating a human quality control stage, ensuring the quality of all generated videos and QA pairs. We establish a comprehensive $3 \times 2 \times 2$ video taxonomy, encompassing \textit{Spatial Scale}, \textit{Perspective}, and \textit{Scene Dynamics} to span diverse scenarios. Furthermore, we design a hierarchical task suite that decouples low-level visual perception from high-level spatio-temporal reasoning. By shifting the paradigm from passive curation to active synthesis, VGenST-Bench enables fine-grained diagnosis of spatio-temporal understanding in MLLMs.
\end{abstract}

\section{Introduction}

\begin{figure}[t]
\centering
\includegraphics[width=1.0\textwidth]{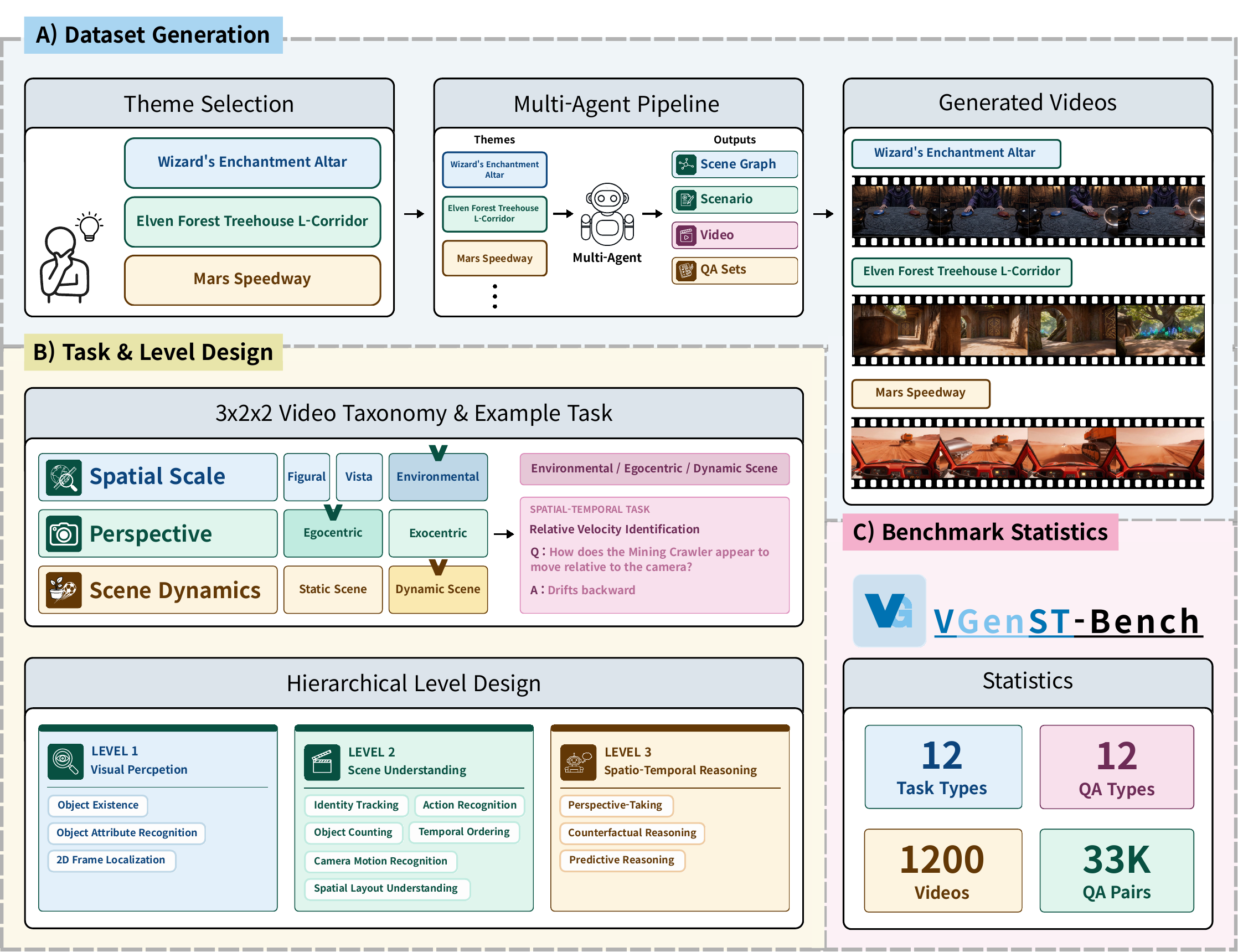}

\vspace{-1ex}

\caption{\textbf{Overview of VGenST-Bench.} \textbf{A) Dataset generation.} Given input video themes, our multi-agent pipeline jointly synthesizes videos paired with scene graphs, scenarios, and QA sets. \textbf{B) Task \& level design.} Videos are organized along a $3 \times 2 \times 2$ taxonomy over \textit{Spatial scale}, \textit{Perspective}, and \textit{Scene dynamics}, with one spatio-temporal task assigned per cell. QA pairs follow a three-level hierarchy: (L1) Visual perception, (L2) Scene understanding, and (L3) Spatio-temporal reasoning. \textbf{C) Benchmark statistics.} VGenST-Bench comprises 1,200 videos and 33K QA pairs spanning 12 task types and 12 QA types.}

\vspace{-2ex}

\label{fig:overview}
\end{figure}

Multimodal Large Language Models (MLLMs) have rapidly advanced beyond basic perceptual tasks such as image recognition and captioning, and are now being deployed in physically grounded applications, including robotics \cite{driess2023palm,zitkovich2023rt,kim2024openvla} and autonomous driving \cite{xu2024drivegpt4,tian2024drivevlm}. These deployments position MLLMs as a foundation toward world models that can understand and predict the dynamics of physical environments \cite{hu2023gaia,intelligence2025pi}.
However, despite this progress, current MLLMs still exhibit notable challenges in understanding how objects and scenes evolve over time and across viewpoints. In particular, \textit{spatio-temporal reasoning}, the ability to perceive and infer the positions, orientations, and attributes of objects across time and changing perspectives, remains a major challenge \cite{lin2025ost,lin2025mmsi,li2025sti}.

To evaluate these capabilities, numerous benchmarks have been proposed \cite{cai2025holistic,zheng2025multimodal,liu2025spatial}. However, existing efforts predominantly focus on static image-based spatial reasoning, which cannot capture dynamic spatio-temporal relationships \cite{kamath2023s,zhang2025sphere,wang2025spatial457,jia2025omnispatial}. Recent video-based benchmarks have begun to address this gap, but they share a common reliance on \textit{passive curation}, collecting clips from the web or using existing datasets, which gives rise to three recurring limitations.

\textbf{(i) Susceptibility to data contamination.} Modern MLLMs ingest vast volumes of publicly available video and image data during pretraining, making evaluations on passively curated benchmarks vulnerable to train-test overlap. Such contamination is pervasive in multimodal settings and systematically inflates reported performance, leaving the reliability of current MLLM evaluations questionable \cite{song2024both,chen2024we,sainz2023nlp}. \textbf{(ii) Shortcut exploitation.} Beyond contamination, passively curated benchmarks inherit distributional regularities from their source data that allow models to substitute linguistic priors, single-frame cues, or static scene context for genuine spatio-temporal reasoning \cite{cores2024lost, krojer2025shortcut}. Recent studies show that standard video-language benchmarks fail to isolate temporal understanding \cite{buch2022revisiting,lei2023revealing,cai2024temporalbench}, suggesting that much of the reported progress on spatio-temporal reasoning may reflect exploitation of shortcuts rather than the capability these benchmarks purport to measure. 
\textbf{(iii) Limited scalability and narrow coverage.} Constructing video benchmarks from web sources requires extensive manual effort to collect, filter, and annotate clips that contain the desired reasoning scenarios \cite{zhou2025mlvu, yu2019activitynet, lei2020tvqa+}. As an alternative, recent benchmarks repurpose existing 3D scene datasets \cite{dai2017scannet,baruch2021arkitscenes} as their data source \cite{yin2025spatial,yang2025thinking, gong2025space10comprehensivebenchmarkmultimodal, linghu2024multi}, but these usually cover only a narrow range of 3D environments, making it difficult to extend evaluation to diverse spatial scales, perspectives, or scene dynamics.

Recent advances in video generative models have demonstrated remarkable capabilities in synthesizing high-fidelity video \cite{wiedemer2025video, seedance2025seedance, wan2025wanopenadvancedlargescale}. This enables a fundamentally different approach to benchmark construction—actively synthesizing precisely controlled evaluation scenarios rather than passively curating them from existing sources. This motivates our question: \textit{Can actively synthesized videos serve as a reliable testbed for spatio-temporal reasoning in MLLMs?}

In this work, we introduce \textbf{VGenST-Bench}, a benchmark leveraging \textbf{V}ideo \textbf{Gen}erative models to evaluate \textbf{S}patio-\textbf{T}emporal reasoning in MLLMs. To the best of our knowledge, VGenST-Bench is the first benchmark built on photorealistic videos synthesized by video generative models for this purpose. To construct this benchmark, we design a multi-agent pipeline that generates benchmark-ready evaluation videos and questions, followed by a final human quality-control stage. The detailed pipeline design is provided in Fig.~\ref{fig:pipeline}.

Grounded in cognitive studies of spatial cognition and event perception \cite{hegarty2006spatial,montello1993scale,klatzky1998allocentric}, VGenST-Bench is organized under a $3 \times 2 \times 2$ taxonomy along three orthogonal axes: \textbf{Spatial scale}, \textbf{Perspective}, and \textbf{Scene dynamics}. This taxonomy yields 12 video categories covering a broad range of spatio-temporal reasoning scenarios. For each category, we design a dedicated spatio-temporal reasoning task tailored to its characteristic combination of three axes. We further pair each task with a three-level question hierarchy spanning \textbf{(L1) Visual perception}, \textbf{(L2) Scene understanding}, and \textbf{(L3) Spatio-temporal reasoning}. This hierarchy enables fine-grained diagnosis of where models succeed or fail along the perception-to-reasoning. Extensive experiments on a diverse set of proprietary and open-source models reveal that performance degrades sharply from L1 to L3, and even the strongest model falls substantially short of human performance. These results highlight the effectiveness of VGenST-Bench in revealing the spatio-temporal reasoning limitations of current MLLMs.

In summary, our work makes the following key contributions:
\begin{itemize}
    \item \textbf{Video benchmark with active synthesis paradigm:} We propose VGenST-Bench, the first benchmark to evaluate spatio-temporal reasoning in MLLMs using actively synthesized video, organized under a $3 \times 2 \times 2$ taxonomy with 12 reasoning tasks and a three-level question hierarchy.
    \item \textbf{Benchmark construction pipeline:} We design a multi-agent generation pipeline that jointly synthesizes scene graphs, scenarios, videos, and QA sets, followed by a human quality-control stage. This pipeline enables controllable construction of evaluation scenarios at scale, overcoming the passive curation bottleneck of prior video benchmarks.
    \item \textbf{Comprehensive experiments on MLLMs:} We conduct in-depth diagnostic experiments on a diverse set of proprietary and open-source MLLMs, providing systematic insights into the spatio-temporal reasoning capabilities of current models along our taxonomy and question hierarchy.

\end{itemize}

\begin{table}[t]
\centering
\small
\begin{adjustbox}{max width=\textwidth}
\begin{tabular}{lcccccc}
\toprule
\textbf{Benchmark} & \textbf{Venue/Year} & \textbf{Modality} & \textbf{Reasoning Type} & \textbf{QA Pairs (\#)} & \textbf{Data Scale (\#)} & \textbf{Data Source} \\ 

\midrule

MME~\cite{fu2023mme}&NeurIPS'25&I&S&2.3K&1.1K  &Real image datasets \\
3DSRBench~\cite{ma20253dsrbench}&ICCV'25&I&S&6.9K&2.7K  &Real image datasets \\
SpatialViz-Bench~\cite{wang2025spatialviz}&ICLR'26&I&S&1.1K&1.1K  &Programmatic generated images \\
Spatial457~\cite{wang2025spatial457}&CVPR'25&I&S&23K&1.0K  &Rendered synthetic 3D scenes \\
VSI-Bench~\cite{yang2025thinking}&CVPR'25&V&S&5K&288 &3D indoor scene datasets \\
EgoExoBench~\cite{he2025egoexobench}&NeurIPS'25&V&S/T&7.3K&2.7K &Ego-exo paired video datasets \\
STI-Bench~\cite{li2025sti}&ICCV'25&V&S/T&2K&300 &Autonomous driving \& 3D indoor scene datasets \\

OST-Bench~\cite{lin2025ost}&NeurIPS'25&V&S/T&10K&1.4K &3D indoor scene datasets \\
\midrule

\rowcolor[HTML]{F2F2F2}
\textbf{VGenST-Bench (Ours)} & \textbf{2026} & \textbf{V} & \textbf{S/T} & \textbf{33K} & \textbf{1.2K } & \textbf{Video generative models} \\
\bottomrule
\end{tabular}
\end{adjustbox}

\vspace{1ex}

\caption{\textbf{Comparison of VGenST-Bench with recent MLLM benchmarks.} Our benchmark is the first spatio-temporal reasoning benchmark that leverages video generative models. I: image, V: video; S: spatial, T: temporal; S/T: spatio-temporal.}

\vspace{-5ex}

\label{tab:benchmark_comparison}
\end{table}

\section{Related Work}
\textbf{Spatio-temporal reasoning benchmarks for MLLMs.} 
Early benchmarks evaluate spatial understanding in MLLMs through static 2D images, probing object localization, relative position, and compositional spatial relations \cite{johnson2017clevr, liu2023visual, kamath2023s, chen2024spatialvlm, cheng2024spatialrgpt, tong2024eyes, fu2024blink, ma20253dsrbench, wang2024picture, jia2025omnispatial}. While valuable, static image-based evaluation possesses an inherent limitation, a fundamental inability to capture state transitions across the temporal dimension. To bridge this gap, recent studies have begun to incorporate video datasets \cite{liu2024tempcompass,li2024mvbench,fu2025video,zhou2025mlvu,yu2019activitynet,lei2020tvqa+} or repurpose existing 3D scene datasets \cite{lin2025ost,yang2025thinking,yin2025spatial,linghu2024multi,gong2025space10comprehensivebenchmarkmultimodal} to evaluate spatio-temporal reasoning. Although these sources provide visual richness, they are \textit{passively curated} from in-the-wild environments rather than \textit{actively designed} for reasoning evaluation. This reliance on public data not only limits the diversity and controllability of evaluation scenarios but also exposes the benchmarks to data contamination. A complementary line of work utilizes synthetic evaluation data \cite{yi2019clevrer, salewski2020clevr, li2023super, zhu2026video, ma20253dsrbench}. While affording precise ground-truth control, these benchmarks suffer from a visual realism gap, limiting their utility for evaluating modern MLLMs trained on photorealistic data. Most recently, a few works have begun to leverage video generative models for benchmark construction \cite{li2025videohallu, fu2025learning}, but they primarily target hallucination detection or physics plausibility rather than spatio-temporal reasoning. Compared with these prior works, as shown in Tab.~\ref{tab:benchmark_comparison}, VGenST-Bench is the first video spatio-temporal reasoning benchmark constructed entirely from video generative models, enabling controllable, diverse scenarios at scale. More comprehensive discussion is provided in Appendix~\ref{app:Extended Related Work}.

\begin{figure}[t]
  \centering
  \includegraphics[width=1.0\textwidth]{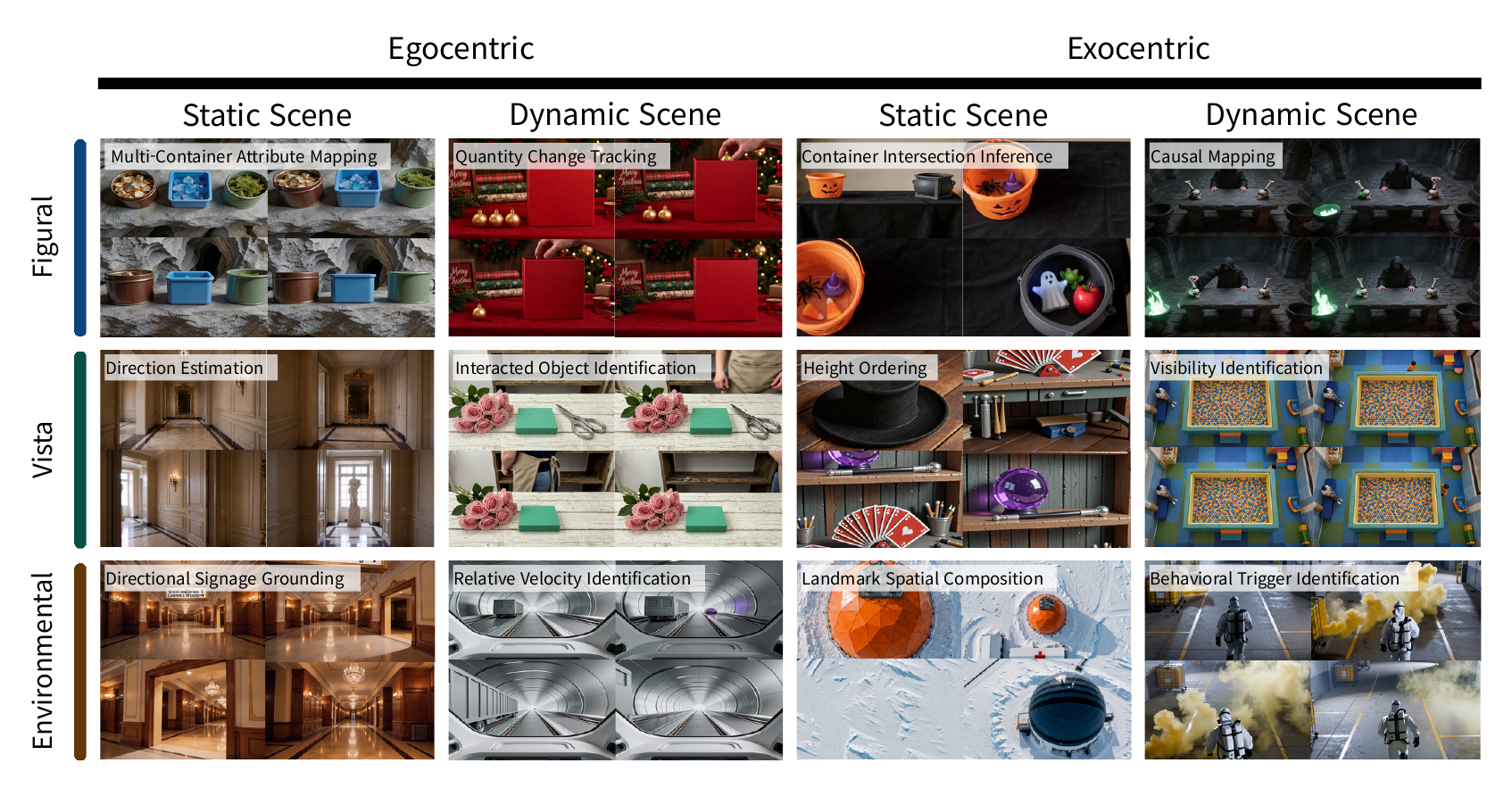}

  \vspace{-1.5ex}
  \caption{\textbf{Representative videos for the 12 tasks of VGenST-Bench.} Each cell of the $3 \times 2 \times 2$ taxonomy (Spatial scale $\times$ Perspective $\times$ Scene dynamics) is paired with one dedicated reasoning task. Rows correspond to spatial scales (Figural / Vista / Environmental); columns are grouped by perspective (Egocentric / Exocentric) and scene dynamics (Static / Dynamic). Each strip shows four sampled frames from a representative video for the task.}

  \vspace{-2ex}
\label{fig:task_examples}
\end{figure}

\section{VGenST-Bench}

\subsection{Video Taxonomy and Task Design}
\label{sec:Video Taxonomy and Task Design}

To systematically cover spatio-temporal reasoning scenarios, we organize VGenST-Bench under a $3 \times 2 \times 2$ taxonomy along three axes: \textbf{(i) Spatial scale} (\textit{figural}, \textit{vista}, \textit{environmental}), 
\textbf{(ii) Perspective} (\textit{egocentric}, \textit{exocentric}), and 
\textbf{(iii) Scene dynamics} (\textit{static}, \textit{dynamic}). These axes are motivated by cognitive studies of spatial cognition and event perception, which suggest that spatial reasoning varies with the scale of space, the reference frame used to encode spatial relations, and whether the scene involves static configurations or dynamic events. Each combination of axis values defines a distinct video category. We design one dedicated reasoning task per cell, yielding 12 tasks that together probe the full taxonomy (Tab.~\ref{tab:tasks_def}, with visual examples in Fig.~\ref{fig:task_examples}). 
Further details are provided in Appendix~\ref{app:vgenst_details}.

\begin{table}[t]
\centering
\footnotesize
\renewcommand{\arraystretch}{1.15}
\setlength{\tabcolsep}{4pt}
\begin{tabular}{ll|p{4.5cm}|p{4.5cm}}
\toprule
\rowcolor[HTML]{F2F2F2}
\multicolumn{2}{c|}{\textbf{Scale} $\times$ \textbf{Dynamics}} & \textbf{Egocentric} & \textbf{Exocentric} \\
\midrule
\multirow{2}{*}{\textbf{Figural}} 
& Static  & \textbf{MC}: Multi-Container Attribute Mapping & \textbf{CI}: Container Intersection Inference \\
& Dynamic & \textbf{QC}: Quantity Change Tracking & \textbf{CM}: Causal Mapping \\
\midrule
\multirow{2}{*}{\textbf{Vista}} 
& Static  & \textbf{DE}: Direction Estimation & \textbf{HO}: Height Ordering \\
& Dynamic & \textbf{IO}: Interacted Object Identification & \textbf{VI}: Visibility Identification \\
\midrule
\multirow{2}{*}{\textbf{Environmental}} 
& Static  & \textbf{DS}: Directional Signage Grounding & \textbf{LS}: Landmark Spatial Composition \\
& Dynamic & \textbf{RV}: Relative Velocity Identification & \textbf{BT}: Behavioral Trigger Identification \\
\bottomrule
\end{tabular}

\vspace{1ex}

\caption{Twelve tasks of VGenST-Bench, organized along the $3 \times 2 \times 2$ taxonomy (Spatial scale $\times$ Perspective $\times$ Scene dynamics). Each cell contains the task code (bold) and full name.}

\vspace{-1ex}
\label{tab:tasks_def}

\end{table}

\begin{table}[h]
\centering

\footnotesize
\begin{tabular}{c|c|c}
\toprule
\rowcolor[HTML]{F2F2F2}
\textbf{L1: Visual Perception} & \textbf{L2: Scene Understanding} & \textbf{L3: Spatio-Temporal Reasoning} \\
\midrule
Object Existence (OE) & Identity Tracking (IT) & Perspective-Taking (PT) \\
Object Attribute Recognition (OA) & Action Recognition (AR) & Counterfactual Reasoning (CR) \\
2D Frame Localization (FL) & Object Counting (OC) & Predictive Reasoning (PR) \\
& Temporal Ordering (TO) & \\
& Camera Motion Recognition (CM) & \\
& Spatial Layout Understanding (SL) & \\
\bottomrule
\end{tabular}

\vspace{1ex}

\caption{Twelve QA types of VGenST-Bench, organized along the three-level cognitive hierarchy.}

\label{tab:qa_types}

\end{table}

\subsection{QA Design}
\label{sec:QA Design}

\textbf{Level Design.} Orthogonal to the video taxonomy, each video is paired with QA pairs organized along a three-level cognitive hierarchy that progresses from low-level perception to high-level reasoning, inspired by recent hierarchical spatial reasoning benchmarks \cite{zhang2025sphere, wang2025spatial457, li2025unfolding}. As shown in Tab.~\ref{tab:qa_types}, the hierarchy comprises \textbf{(L1) Visual perception} (3 QA types), which probes the recognition of objects and their visual attributes from individual frames; \textbf{(L2) Scene understanding} (6 QA types), which assesses the integration of perceptual cues across frames into coherent spatial and temporal structures; and \textbf{(L3) Spatio-temporal reasoning} (3 QA types), which evaluates higher-order inference such as perspective-taking, counterfactual, and predictive reasoning. Full QA type definitions and examples are provided in Appendix~\ref{app:QA Type Definitions}.

\textbf{Task–QA Applicability.}
Not every QA type is applicable for each of the twelve tasks. For instance, Action Recognition is undefined for tasks with static scenes. We therefore define a \emph{task--QA applicability matrix} (Appendix~\ref{app:Task--QA Applicability Matrix}) that specifies which QA types are evaluated for each of the 12 tasks, ensuring that every QA pair is well-defined for its underlying video while preserving balanced coverage across the hierarchy.

\textbf{Question Reformulation for Robust Evaluation.}
A central concern in multiple-choice evaluation is that models may exploit option-level shortcuts rather than genuinely reasoning about the video \cite{lei2023revealing, krojer2025shortcut}. To mitigate this, each base multiple-choice question (MCQ) is expanded into three variants: (i) a \textit{None-of-these distractor}, which adds ``None of these'' as an additional incorrect option to test whether models commit to the correct choice when one is present; (ii) a \textit{None-of-these answer}, which replaces the correct option with ``None of these'' to test whether models can reject all listed options when appropriate; and (iii) an \textit{open-ended} variant that removes options entirely to assess reasoning without choice priors. After filtering, this protocol yields a total of 33K QA pairs across the benchmark. We provide more details in Appendix \ref{app:Question Reformulation Variants}.

\subsection{Dataset Construction}
\label{sec:Dataset Construction}

We construct VGenST-Bench through a multi-agent pipeline that sequentially synthesizes scene graphs, scenarios, videos, and QA pairs, followed by a human quality-control stage.

\textbf{Structuring Inputs to a Video Generator.} To make video generation \emph{benchmark-ready}, our pipeline conditions the video generator on three structured representations. \textbf{(1) Theme} provides the visual and semantic context of the video, the style of objects, and the overall environment (e.g., \textit{Cyberpunk Hacker's Neon Desk}, \textit{Wizard's Enchantment Altar}). \textbf{(2) Scene Graph} specifies the static spatial configuration required by the task: the objects to be rendered, their attributes (color, material, role), and their pairwise spatial relations. \textbf{(3) Scenario} lifts the scene graph into the temporal domain, specifying the reasoning goal, the camera setup, and a structured timeline of events. Together, these three representations fix the visual context and the spatio-temporal ground truth needed for both rendering and QA generation.

\begin{figure}[t]
  \centering
  \includegraphics[width=1.0\textwidth]{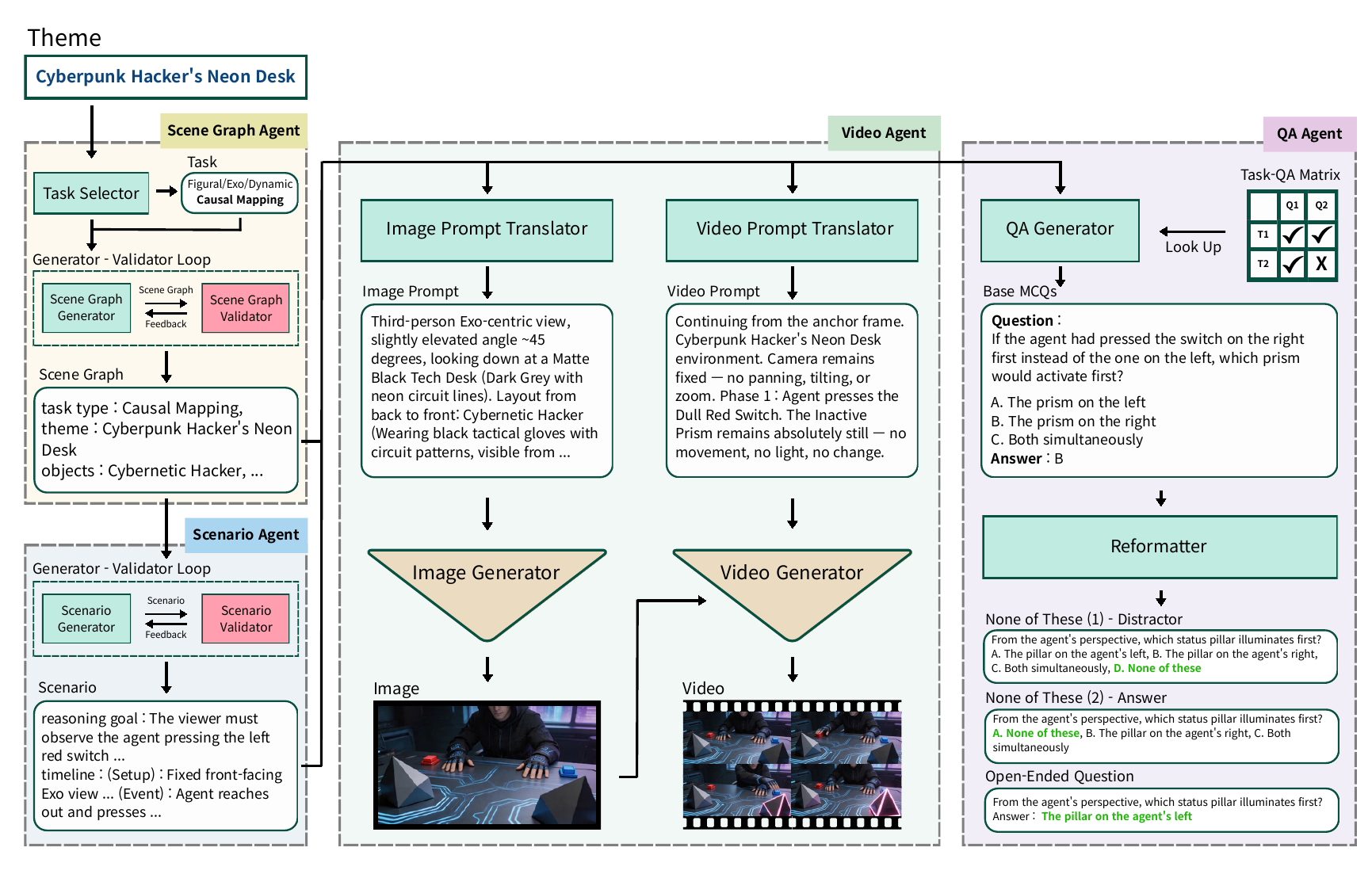}

\caption{\textbf{VGenST-Bench construction pipeline.} 
Starting from a theme, four agents operate in sequence. The \textbf{Scene Graph Agent} produces a structured scene graph specifying objects and spatial composition; the \textbf{Scenario Agent} expands it into a temporally grounded scenario with reasoning goal and timeline; the \textbf{Video Agent} synthesizes the corresponding image and video through generative models; and the \textbf{QA Agent} generates base MCQs from a task--QA applicability matrix and reformats each into three variants.}

\label{fig:pipeline}
\end{figure}

\textbf{Multi-Agent Pipeline.} Given an input theme, the pipeline produces the \textbf{scene graph, scenario, video, and QA pairs} through four agent modules as shown in Fig.~\ref{fig:pipeline}. \textbf{(1) Scene Graph Agent} produces a scene graph from a (theme, task) pair sampled by a Task Selector; a Validator iteratively rejects scene graphs missing required objects, attributes, or relations and returns feedback to the Generator. \textbf{(2) Scenario Agent} translates the validated scene graph into a temporal scenario; a Validator iteratively verifies that the timeline is sufficient to derive the ground-truth answer and contains no contradictions. \textbf{(3) Video Agent} renders the scenario in two stages: an Image Prompt Translator produces a first-frame prompt that an image generator turns into an anchor frame, and a Video Prompt Translator composes a video prompt that a video generator combines with the anchor frame to produce the final clip. This image-anchored design stabilizes scene composition and reduces visual drift of generated videos. We employ a diverse pool of contemporary image and video generative models, 
and primarily select the best output per scenario. \textbf{(4) QA Agent} generates base MCQs by looking up the task-QA applicability matrix and conditioning on the scene graph and scenario as ground-truth references; a Reformatter then expands each base MCQ into the three variants (Section~\ref{sec:QA Design}). We provide more details in Appendix~\ref{app:Construction Pipeline Details}.

\textbf{Human Quality Control.} 
All generated videos and base QA pairs undergo a two-stage human verification protocol with a pair of validators per task, retaining only items both validators mark valid. In \emph{Stage 1 (Video QC)}, the validator pair reviews the generated videos and rejects clips that fail visual fidelity or scenario adherence. In \emph{Stage 2 (QA QC)}, the same pair reviews each base MCQ on the surviving videos and rejects items with ambiguous or invalid answers. Further details are reported in Appendix~\ref{app:Human Quality Control}.

\section{Experiments}

\begin{table}[t]
\centering

\footnotesize
\setlength{\tabcolsep}{3pt}
\renewcommand{\arraystretch}{1.1}
\begin{tabular}{l|cccc|cccc|cccc|cc}
\toprule
& \multicolumn{4}{c|}{\textbf{Figural}}
& \multicolumn{4}{c|}{\textbf{Vista}}
& \multicolumn{4}{c|}{\textbf{Environmental}}
& \multicolumn{2}{c}{\textbf{Avg}} \\
\textbf{Model} & MC & QC & CI & CM & DE & IO & HO & VI & DS & RV & LS & BT & Macro & Micro \\
\midrule
\rowcolor[HTML]{F2F2F2}
\multicolumn{15}{l}{\textit{Baseline}} \\
\midrule
Human             & 100.0 & 100.0 & 100.0 & 99.5 & 98.0 & 99.5 & 100.0 & 97.0 & 100.0 & 99.0 & 96.1 & 99.0 & 99.0 & 99.0 \\
Random            & 0.8 & 0.8 & 1.0 & 2.1 & 0.1 & 0.9 & 0.6 & 0.1 & 1.1 & 1.0 & 0.1 & 1.2 & 0.8 & 0.8 \\

\midrule
\rowcolor[HTML]{F2F2F2}
\multicolumn{15}{l}{\textit{Proprietary Models}} \\
\midrule
GPT-5.4                   & \cellcolor[HTML]{D7E5FA}88.8 & \cellcolor[HTML]{CCDEF9}71.0 & \cellcolor[HTML]{D1E1F9}92.5 & \cellcolor[HTML]{D0E0F9}79.0 & \cellcolor[HTML]{CBDDF9}76.4 & \cellcolor[HTML]{CBDDF9}94.0 & \cellcolor[HTML]{D7E5FA}84.5 & \cellcolor[HTML]{CBDDF9}80.6 & \cellcolor[HTML]{CBDDF9}97.0 & \cellcolor[HTML]{D7E5FA}81.9 & \cellcolor[HTML]{D8E5FA}57.0 & \cellcolor[HTML]{D5E4FA}90.1 & \cellcolor[HTML]{D0E0F9}82.7 & \cellcolor[HTML]{D0E0F9}82.7 \\
GPT-5.4 mini              & \cellcolor[HTML]{F6F9FE}68.5 & \cellcolor[HTML]{F7FAFE}46.3 & \cellcolor[HTML]{E7EFFC}75.4 & \cellcolor[HTML]{DDE9FB}72.5 & \cellcolor[HTML]{E6EFFC}60.0 & \cellcolor[HTML]{DEE9FB}80.4 & \cellcolor[HTML]{E6EFFC}70.7 & \cellcolor[HTML]{E8F0FC}60.2 & \cellcolor[HTML]{ECF3FD}75.0 & \cellcolor[HTML]{E4EDFC}73.8 & \cellcolor[HTML]{F5F9FE}41.0 & \cellcolor[HTML]{E6EFFC}80.2 & \cellcolor[HTML]{E9F0FC}67.0 & \cellcolor[HTML]{E9F0FC}67.1 \\
GPT-5.4 nano              & \cellcolor[HTML]{FCFDFF}64.2 & \cellcolor[HTML]{EEF4FD}48.2 & \cellcolor[HTML]{EDF3FD}68.2 & \cellcolor[HTML]{F6F9FE}52.0 & \cellcolor[HTML]{E2ECFB}59.4 & \cellcolor[HTML]{F2F7FD}62.9 & \cellcolor[HTML]{FFFFFF}44.0 & \cellcolor[HTML]{F4F8FE}48.8 & \cellcolor[HTML]{FDFEFF}59.5 & \cellcolor[HTML]{F4F8FE}57.9 & \cellcolor[HTML]{F3F7FD}37.1 & \cellcolor[HTML]{E3ECFB}78.8 & \cellcolor[HTML]{F3F7FE}56.8 & \cellcolor[HTML]{F3F7FE}56.8 \\
Gemini 3.1 Flash-Lite     & \cellcolor[HTML]{D9E6FA}87.9 & \cellcolor[HTML]{CBDDF9}71.6 & \cellcolor[HTML]{CBDDF9}97.0 & \cellcolor[HTML]{D4E3FA}76.2 & \cellcolor[HTML]{D9E6FA}66.9 & \cellcolor[HTML]{CCDDF9}93.9 & \cellcolor[HTML]{D5E3FA}87.1 & \cellcolor[HTML]{D8E5FA}71.1 & \cellcolor[HTML]{D4E2FA}91.0 & \cellcolor[HTML]{DDE8FB}77.2 & \cellcolor[HTML]{E5EEFC}47.9 & \cellcolor[HTML]{D8E6FA}88.0 & \cellcolor[HTML]{D4E3FA}79.6 & \cellcolor[HTML]{D4E3FA}79.6 \\
Gemini 3 Flash            & \cellcolor[HTML]{CBDDF9}97.0 & \cellcolor[HTML]{CCDDF9}71.4 & \cellcolor[HTML]{CDDEF9}95.7 & \cellcolor[HTML]{CBDDF9}82.2 & \cellcolor[HTML]{D0E0F9}73.4 & \cellcolor[HTML]{CCDDF9}93.9 & \cellcolor[HTML]{CBDDF9}96.9 & \cellcolor[HTML]{DEE9FB}66.1 & \cellcolor[HTML]{CBDDF9}97.0 & \cellcolor[HTML]{CBDDF9}91.1 & \cellcolor[HTML]{CBDDF9}66.0 & \cellcolor[HTML]{CBDDF9}96.9 & \cellcolor[HTML]{CBDDF9}85.6 & \cellcolor[HTML]{CBDDF9}85.9 \\

\midrule
\rowcolor[HTML]{F2F2F2}
\multicolumn{15}{l}{\textit{Open-Source Models}} \\
\midrule
Qwen3.5-4B                & \cellcolor[HTML]{DDE9FB}81.4 & \cellcolor[HTML]{E6EEFC}51.6 & \cellcolor[HTML]{E7EFFC}72.9 & \cellcolor[HTML]{E5EEFC}61.4 & \cellcolor[HTML]{ECF2FD}50.4 & \cellcolor[HTML]{CCDDF9}91.9 & \cellcolor[HTML]{E9F1FC}63.6 & \cellcolor[HTML]{E5EEFC}59.5 & \cellcolor[HTML]{D1E1F9}88.0 & \cellcolor[HTML]{F9FBFE}53.2 & \cellcolor[HTML]{F2F7FD}35.6 & \cellcolor[HTML]{FFFFFF}58.9 & \cellcolor[HTML]{E6EFFC}64.0 & \cellcolor[HTML]{E7EFFC}63.6 \\
Qwen3.5-9B                & \cellcolor[HTML]{D9E6FA}85.2 & \cellcolor[HTML]{E1EBFB}54.4 & \cellcolor[HTML]{E3EDFB}75.8 & \cellcolor[HTML]{E1EBFB}63.3 & \cellcolor[HTML]{E8F0FC}52.0 & \cellcolor[HTML]{CBDDF9}92.1 & \cellcolor[HTML]{E6EFFC}67.2 & \cellcolor[HTML]{E1EBFB}62.1 & \cellcolor[HTML]{CDDEF9}90.9 & \cellcolor[HTML]{F5F8FE}56.3 & \cellcolor[HTML]{EEF4FD}38.0 & \cellcolor[HTML]{FAFCFE}63.9 & \cellcolor[HTML]{E2ECFB}66.8 & \cellcolor[HTML]{E3ECFB}66.4 \\
Qwen3.5-27B               & \cellcolor[HTML]{D5E4FA}87.6 & \cellcolor[HTML]{DEE9FB}57.0 & \cellcolor[HTML]{E0EBFB}77.9 & \cellcolor[HTML]{DEE9FB}65.6 & \cellcolor[HTML]{E4EDFC}54.4 & \cellcolor[HTML]{CCDDF9}91.4 & \cellcolor[HTML]{E4EDFC}69.5 & \cellcolor[HTML]{DEEAFB}64.7 & \cellcolor[HTML]{CBDDF9}91.9 & \cellcolor[HTML]{F2F6FD}58.2 & \cellcolor[HTML]{EAF2FC}41.2 & \cellcolor[HTML]{F6F9FE}64.8 & \cellcolor[HTML]{DFEAFB}68.7 & \cellcolor[HTML]{E0EBFB}68.3 \\
InternVL3.5-4B            & \cellcolor[HTML]{FFFFFF}62.9 & \cellcolor[HTML]{FFFFFF}38.3 & \cellcolor[HTML]{FFFFFF}55.2 & \cellcolor[HTML]{FFFFFF}46.9 & \cellcolor[HTML]{FFFFFF}41.4 & \cellcolor[HTML]{FFFFFF}55.0 & \cellcolor[HTML]{FCFDFF}47.6 & \cellcolor[HTML]{FFFFFF}42.0 & \cellcolor[HTML]{FFFFFF}60.5 & \cellcolor[HTML]{FFFFFF}51.2 & \cellcolor[HTML]{FCFDFF}32.7 & \cellcolor[HTML]{F3F7FE}70.2 & \cellcolor[HTML]{FFFFFF}50.3 & \cellcolor[HTML]{FFFFFF}50.3 \\
InternVL3.5-8B            & \cellcolor[HTML]{F1F6FD}72.6 & \cellcolor[HTML]{FBFCFE}42.0 & \cellcolor[HTML]{FDFEFF}57.6 & \cellcolor[HTML]{FEFFFF}48.7 & \cellcolor[HTML]{FBFCFE}46.6 & \cellcolor[HTML]{F8FAFE}60.2 & \cellcolor[HTML]{FAFBFE}51.1 & \cellcolor[HTML]{F6F9FE}49.2 & \cellcolor[HTML]{FBFDFF}64.0 & \cellcolor[HTML]{FEFEFF}53.8 & \cellcolor[HTML]{FFFFFF}32.6 & \cellcolor[HTML]{EBF2FD}76.4 & \cellcolor[HTML]{FAFCFE}54.6 & \cellcolor[HTML]{FAFCFE}54.5 \\
InternVL3.5-30B-A3B       & \cellcolor[HTML]{DBE7FA}86.4 & \cellcolor[HTML]{E3EDFC}55.7 & \cellcolor[HTML]{EAF1FC}71.6 & \cellcolor[HTML]{E8F0FC}62.3 & \cellcolor[HTML]{E4EDFC}59.0 & \cellcolor[HTML]{E4EDFC}74.9 & \cellcolor[HTML]{EAF1FC}64.7 & \cellcolor[HTML]{E1ECFB}63.5 & \cellcolor[HTML]{E6EEFC}77.8 & \cellcolor[HTML]{EAF1FC}66.8 & \cellcolor[HTML]{E8F0FC}45.7 & \cellcolor[HTML]{D5E3FA}90.4 & \cellcolor[HTML]{E4EDFC}68.2 & \cellcolor[HTML]{E4EDFC}68.1 \\
Gemma-4-26B-A4B-it        & \cellcolor[HTML]{E9F0FC}77.0 & \cellcolor[HTML]{DAE7FA}61.9 & \cellcolor[HTML]{DAE7FA}84.4 & \cellcolor[HTML]{E7EFFC}62.5 & \cellcolor[HTML]{EBF2FD}54.3 & \cellcolor[HTML]{E1EBFB}77.2 & \cellcolor[HTML]{EDF3FD}61.9 & \cellcolor[HTML]{DEE9FB}66.0 & \cellcolor[HTML]{E0EBFB}81.6 & \cellcolor[HTML]{EFF5FD}62.8 & \cellcolor[HTML]{EFF4FD}41.1 & \cellcolor[HTML]{E5EEFC}79.7 & \cellcolor[HTML]{E5EEFC}67.5 & \cellcolor[HTML]{E5EEFC}67.4 \\
Gemma-4-31B-it            & \cellcolor[HTML]{E1EBFB}82.0 & \cellcolor[HTML]{D2E2F9}66.9 & \cellcolor[HTML]{D4E3FA}89.4 & \cellcolor[HTML]{E0EBFB}67.5 & \cellcolor[HTML]{E4EDFC}59.3 & \cellcolor[HTML]{DAE7FA}82.2 & \cellcolor[HTML]{E8F0FC}66.8 & \cellcolor[HTML]{D8E5FA}71.1 & \cellcolor[HTML]{DAE6FA}86.6 & \cellcolor[HTML]{E9F0FC}67.8 & \cellcolor[HTML]{E7EFFC}46.2 & \cellcolor[HTML]{DDE9FB}84.6 & \cellcolor[HTML]{DEE9FB}72.5 & \cellcolor[HTML]{DEE9FB}72.5 \\
GLM-4.6V-Flash            & \cellcolor[HTML]{F7F9FE}68.0 & \cellcolor[HTML]{FCFDFF}40.0 & \cellcolor[HTML]{E3EDFC}76.9 & \cellcolor[HTML]{DEE9FB}69.2 & \cellcolor[HTML]{E8F0FC}56.6 & \cellcolor[HTML]{DDE8FB}80.5 & \cellcolor[HTML]{DDE9FB}78.6 & \cellcolor[HTML]{F0F5FD}52.8 & \cellcolor[HTML]{F0F5FD}70.5 & \cellcolor[HTML]{F4F8FE}59.3 & \cellcolor[HTML]{F0F5FD}40.1 & \cellcolor[HTML]{E9F1FC}76.9 & \cellcolor[HTML]{EAF1FC}64.1 & \cellcolor[HTML]{EBF2FC}63.7 \\
Kimi-K2.6                 & \cellcolor[HTML]{EBF2FC}71.2 & \cellcolor[HTML]{EFF4FD}45.0 & \cellcolor[HTML]{CBDDF9}89.1 & \cellcolor[HTML]{D0E0F9}76.7 & \cellcolor[HTML]{CDDEF9}73.1 & \cellcolor[HTML]{CCDDF9}87.6 & \cellcolor[HTML]{D1E1F9}86.1 & \cellcolor[HTML]{E6EFFC}56.4 & \cellcolor[HTML]{DEE9FB}79.8 & \cellcolor[HTML]{E9F0FC}64.2 & \cellcolor[HTML]{E3ECFB}49.8 & \cellcolor[HTML]{DEE9FB}80.9 & \cellcolor[HTML]{DAE6FA}71.7 & \cellcolor[HTML]{DBE7FA}71.0 \\
\bottomrule
\end{tabular}

\vspace{1ex}
\caption{\textbf{Main results on VGenST-Bench.} Tasks are grouped by spatial scale (Figural / Vista / Environmental). All multiple-choice questions are evaluated under \emph{circular evaluation.} Per-task scores are macro-averaged across applicable QA types. \textbf{Macro Avg}: unweighted mean across the 12 tasks. \textbf{Micro Avg}: sample-weighted mean across all questions. See Tab.~\ref{tab:tasks_def} for full task names.}

\vspace{-5ex}

\label{tab:main_results}

\end{table}

\subsection{Experimental Setup}
\label{sec:Experimental Setup}

\textbf{Evaluation Metric.} We report \emph{accuracy} as our primary metric across all questions. For the base MCQ, we adopt \emph{circular evaluation} \cite{liu2024mmbench} to control position bias and choice priors, ensuring that reported accuracy reflects genuine reasoning. Each question is evaluated under all $N$ cyclic permutations of its answer choices, and the model is scored as correct only if it answers correctly under every cycle. For the open-ended variant, we follow the LLM-as-judge protocol \cite{zheng2023judging}, using Claude-Sonnet-4.6 \cite{anthropic2025claude46} to judge whether the model's response is semantically equivalent to the ground-truth answer.

\textbf{Human Baseline.} We report human evaluation performance on VGenST-Bench. To conduct the human evaluation, we recruit 10 participants from diverse backgrounds, excluding computer-science majors to ensure that performance reflects general spatio-temporal reasoning capabilities. Each annotator answers a subset of 10 videos sampled across all 12 tasks (120 videos for each annotator), under the same circular evaluation protocol applied to model evaluation. 

\textbf{Evaluated Models.} We evaluate 15 MLLMs spanning two categories. \textbf{Proprietary Models}. We evaluate GPT-5.4, GPT-5.4-mini/nano \cite{openai2026gpt54}, 
and Gemini 3.1 Flash-Lite, Gemini 3 Flash \cite{google2026gemini3flash}.
\textbf{Open-source Models}. We evaluate
Qwen3.5 (4B/9B/27B) \cite{team2026qwen3}, 
InternVL3.5 (4B/8B/30B-A3B) \cite{wang2025internvl3}, 
Gemma-4 (26B-A4B-it/31B-it) \cite{google2026gemma4}, 
GLM-4.6V-Flash \cite{zai2025glm46v},
and Kimi-K2.6 \cite{moonshot2026kimik26}. For proprietary models, we use the official APIs provided by the respective vendors. 

For open-source models, we serve models locally using vLLM~\cite{kwon2023efficient}, running larger models on 8 $\times$ NVIDIA B200 GPUs and smaller models on 2 $\times$ NVIDIA RTX PRO 5000 GPUs. Across all models, we uniformly sample
K=8 frames per video as input.

\subsection{Main Results}
\label{sec:Main Results}

Tab.~\ref{tab:main_results} reports the main evaluation results across all evaluated models, organized by spatial scale (Figural / Vista / Environmental).
Each cell reports accuracy on base MCQ questions under circular evaluation. As expected, human annotators remain a clear upper bound, achieving 99.0\% on average and near-saturation across all twelve tasks. Even the strongest evaluated MLLM, \textit{Gemini 3 Flash}, achieves only 85.9\% on average---a gap of more than 13pp below human performance, and this gap widens substantially for compact and smaller models, with GPT-5.4 nano below 60\%. Among open-source models, the majority perform below 70\%, revealing limitations in their spatio-temporal reasoning. Notable exceptions are \textit{Kimi-K2.6} and \textit{Gemma-4-31B-it}, which achieve approximately 71.0\% and 72.5\% on average, respectively, narrowing the gap to top proprietary models and demonstrating that large-scale open-source models can be competitive. We provide detailed analysis in Section~\ref{sec:Detailed Analysis}.

\subsection{Detailed Analysis}
\label{sec:Detailed Analysis}

\begin{figure}[t]
\centering
\begin{minipage}[b]{0.6\textwidth}
\centering
\includegraphics[width=\textwidth]{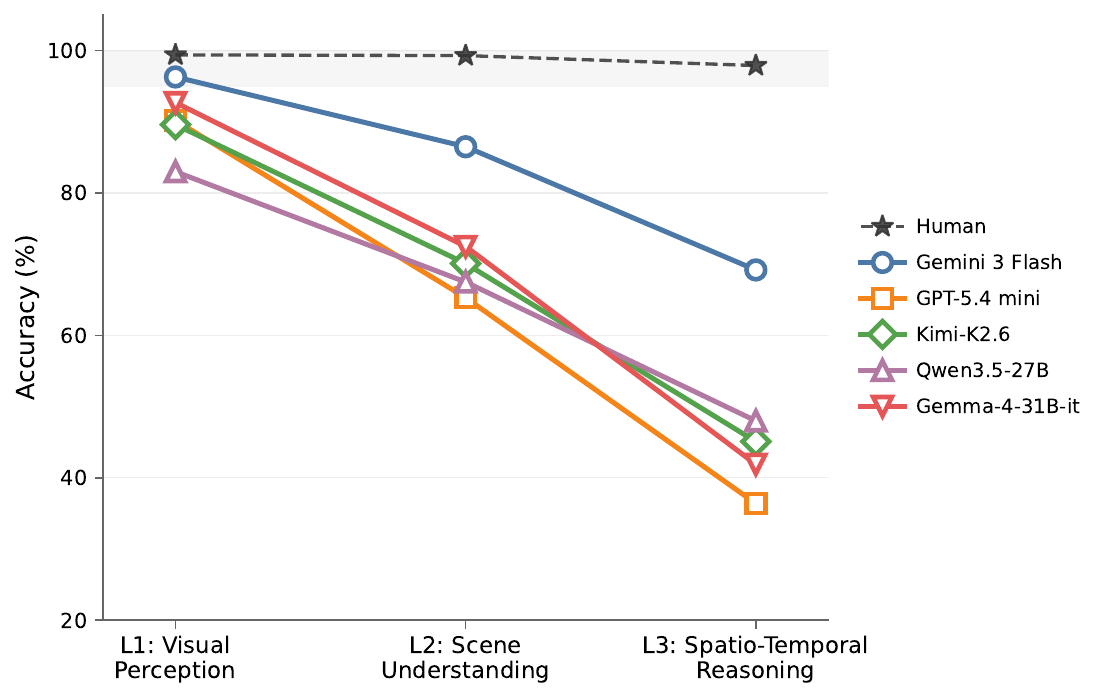}
\\\small (a) Accuracy trends across L1/L2/L3.
\end{minipage}
\hfill
\begin{minipage}[b]{0.39\textwidth}
\centering
\footnotesize
\setlength{\tabcolsep}{4pt}
\renewcommand{\arraystretch}{1.15}
\begin{tabular}{l|ccc|c}
\toprule
\textbf{Model} & \textbf{L1} & \textbf{L2} & \textbf{L3} & \textbf{$\Delta$} \\
\midrule
\rowcolor[HTML]{F2F2F2}
\multicolumn{5}{l}{\textit{Baseline}} \\
Human               & 99.4 & 99.3 & 97.9 &  1.5 \\
\midrule
\rowcolor[HTML]{F2F2F2}
\multicolumn{5}{l}{\textit{Proprietary}} \\
Gemini 3 Flash      & 96.3 & 86.5 & 69.2 & 27.1 \\
GPT-5.4 mini        & 90.2 & 65.3 & 36.4 & 53.8 \\
\midrule
\rowcolor[HTML]{F2F2F2}
\multicolumn{5}{l}{\textit{Open-source}} \\
Kimi-K2.6           & 89.6 & 70.1 & 45.1 & 44.5 \\
Qwen3.5-27B         & 83.0 & 67.5 & 48.0 & 35.0 \\
Gemma-4-31B-it      & 92.7 & 72.5 & 41.9 & 50.8 \\
\bottomrule
\end{tabular}
\vspace{1ex}
\\ \small (b) Detailed scores per model.
\end{minipage}
\caption{\textbf{Hierarchical Analysis: Accuracy across the three question levels.} 
(a) All models degrade consistently from L1 to L3, while humans remain near-ceiling. (b) Breakdown by model, with the L1$-$L3 gap ($\Delta$).}

\vspace{-3ex}

\label{fig:hierarchical_analysis}
\end{figure}

\textbf{Hierarchical Analysis.} We observe a consistent and substantial performance degradation from L1 to L3 as shown in Fig.~\ref{fig:hierarchical_analysis}. GPT-5.4 mini achieves 90.2\% on L1, drops to 65.3\% on L2, 
and further to 36.4\% on L3; this pattern persists across every model category.
In contrast, human annotators decline only marginally (99.4\% $\to$ 97.9\%), whereas other models exhibit dramatically larger drops. This degradation indicates that current MLLMs are competent at single-frame visual perception but exhibit systematic bottlenecks in higher-order reasoning that requires integrating spatial and temporal cues, validating our hierarchy as a diagnostic tool for separating perception from reasoning.

\begin{figure}[b]
\centering
\includegraphics[width=\textwidth]{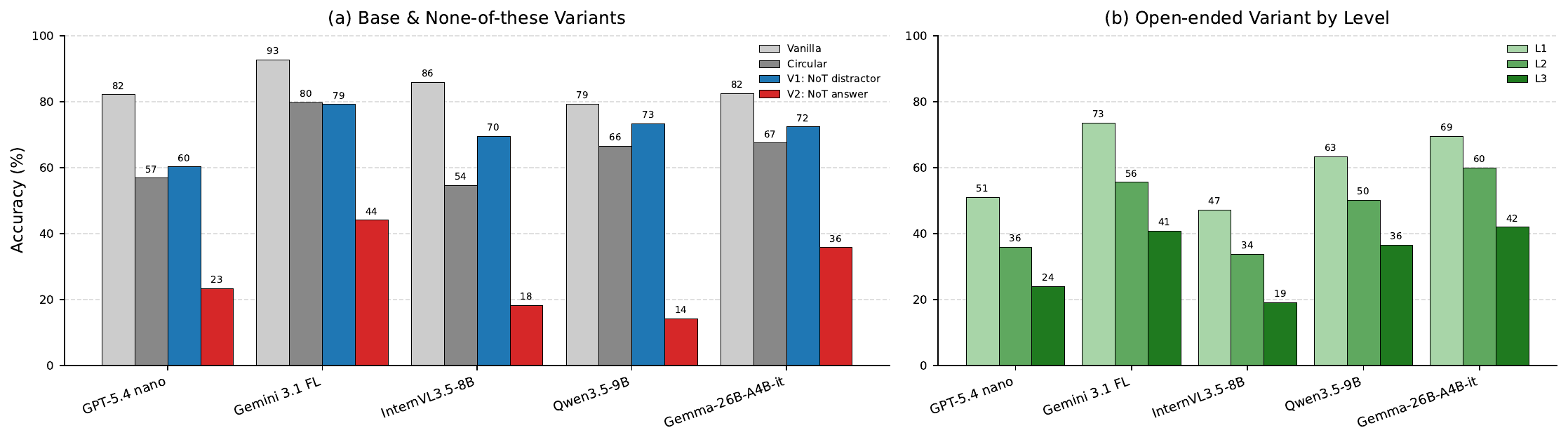}

\caption{\textbf{Robustness Analysis.} 
(a) None-of-these variants show a clear asymmetry: V1 maintains base accuracy, while V2 produces dramatic drops across all models. 
(b) Open-ended evaluation by question level reveals large drop on L3 for all models. 
Together, these reveal that closed-form MCQ accuracy may overestimate spatio-temporal reasoning capability.}
\label{fig:robustness_analysis}
\end{figure}

\textbf{Robustness to Question Reformulation.} Fig.~\ref{fig:robustness_analysis}(a) reveals two complementary findings. 
First, vanilla accuracy systematically exceeds circular accuracy across all models, indicating that current MLLMs exploit position bias and choice priors that single-attempt evaluation does not control for. 
Second, while V1 maintains or slightly exceeds the vanilla accuracy, V2 produces dramatic drops. These results show that current MLLMs perform multiple-choice reasoning by ranking the given options against each other rather than verifying the correct answer against the video. When the correct answer is absent, models cannot reject the remaining distractors and instead select the most plausible one. Fig.~\ref{fig:robustness_analysis}(b) reveals that the open-ended variant exposes the L1$\to$L3 hierarchy more dramatically than closed-form MCQ. Together, these results show that high accuracy on standard multiple-choice benchmarks may overestimate current MLLMs' spatio-temporal reasoning capability, and that our reformulation protocol is meaningful for diagnosing reasoning shortcuts.

\begin{figure}[t]
  \centering
  \includegraphics[width=1.0\textwidth]{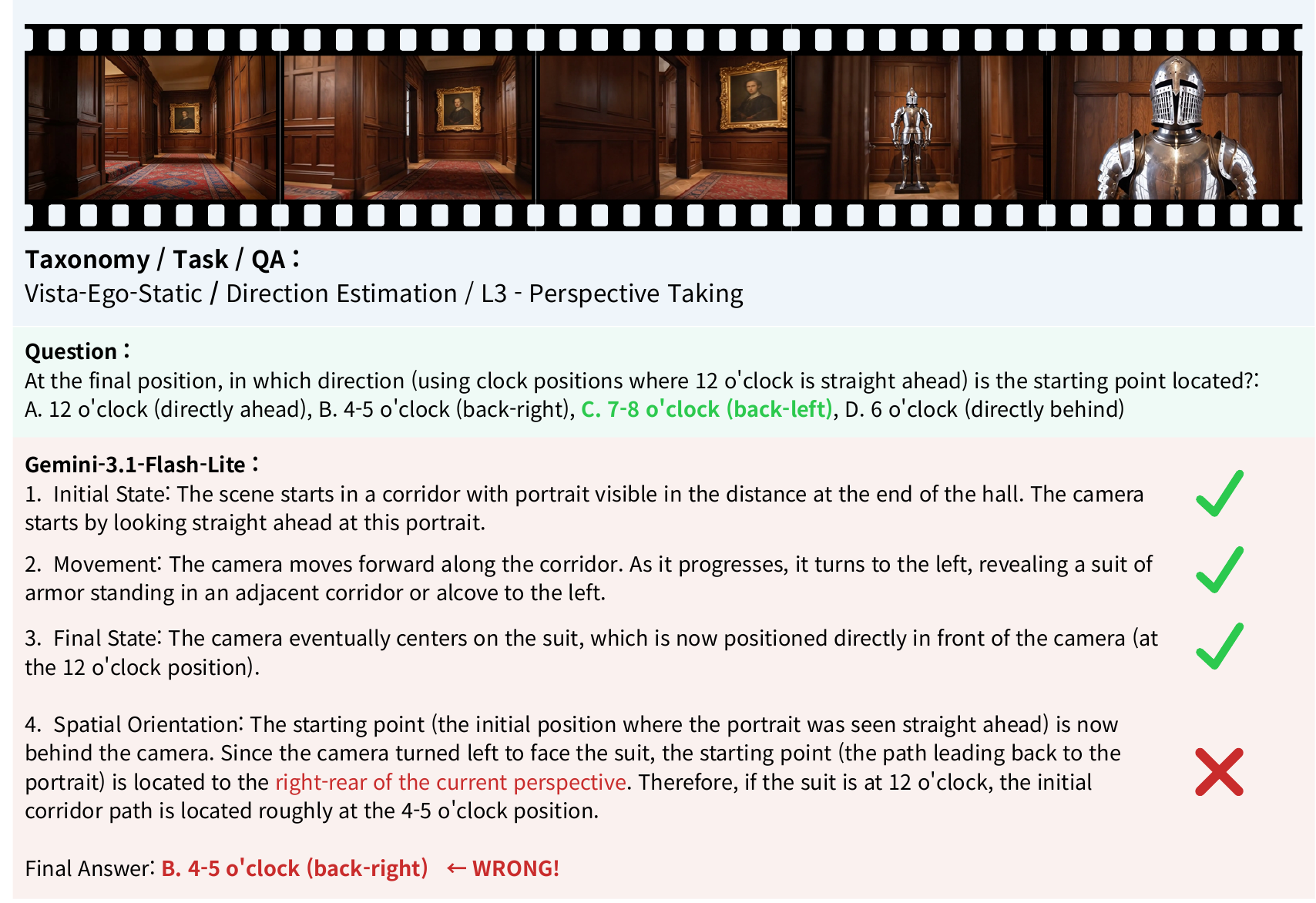}

\caption{\textbf{Reasoning failure of \textit{Direction Estimation} task.} 
The model's reasoning trace correctly identifies the initial orientation, the leftward camera turn, and the final view, but inverts the resulting egocentric direction at the final step, concluding with the wrong answer.}

  \vspace{-2ex}

\label{fig:failure_analysis}
\end{figure}

\textbf{Failure Analysis.} To better understand L3 failures, Fig.~\ref{fig:failure_analysis} examines a representative failure of Gemini~3.1~Flash-Lite on a \textit{Direction Estimation} task. 
The model produces a step-by-step reasoning trace whose first three steps are all visually correct, accurately recognizing the left turn that brings the suit of armor into view. At the final step (Spatial Orientation), however, the model concludes that the starting point lies at right-rear (4--5 o'clock) rather than the correct back-left (7--8 o'clock), inverting the direction of the egocentric transformation. 
This failure is particularly diagnostic: the model exhibits accurate visual \emph{perception} but an error in higher-level \emph{reasoning}.

\section{Conclusion}

We introduce \textbf{VGenST-Bench}, a video benchmark that uses 
video generative models to evaluate spatio-temporal reasoning in 
MLLMs. We find that the strongest model trails the near-perfect 
human ceiling ($99.0\%$) by over 13pp, accuracy collapses 
sharply along the L1$\to$L3 hierarchy, and our None-of-these and 
open-ended reformulations expose reasoning shortcuts hidden by 
closed-form MCQ accuracy. Together, these findings validate 
generation-driven benchmark construction as a viable foundation 
for spatio-temporal reasoning evaluation.

We suggest this work as more than a single benchmark. Building on 
controllable video generation, VGenST-Bench shows that an 
evaluation testbed can be \emph{designed for the capabilities we 
want to probe}, rather than discovered within naturally collected 
footage. We hope this work motivates further benchmark studies, and we release the dataset, the 
generation pipeline, and the full evaluation suite to support 
future research on spatio-temporal reasoning in MLLMs.

\clearpage
\newpage

\bibliographystyle{plainnat}
\bibliography{main}

\newpage

\newpage
\newpage
\appendix
\section*{Appendix}

\startcontents[appendix]
\printcontents[appendix]{}{1}{\setcounter{tocdepth}{2}}

\newpage

\section{Discussion}
\label{app:Discussion}

\subsection{Benchmark Design Rationale}
\label{app:Benchmark Design Rationale}

\textbf{Why generated videos.} 
Naturally collected videos --- the foundation of most existing 
video benchmarks --- impose two structural constraints on what can 
be evaluated. First, the distribution of available sources is constrained by
\emph{whatever happens to exist}: certain combinations of spatial 
scale, viewpoint, and scene dynamics (e.g., ego-perspective 
environmental-scale dynamic scenes) are scarce in public corpora 
and painful to balance through curation. Second, even when such
footage exists, the precise object configurations and event timings cannot be controlled, which makes
it difficult to construct questions whose ground truth is
unambiguous. Controllable video generation removes both 
constraints. We can specify the scenario we wish to probe and 
synthesize video that exhibits the structure we intended. The validity of this trade-off is supported by our 
human study (Appendix~\ref{app:video_quality}), which shows that 
generated videos remain perceptually understandable even to non-expert 
viewers. This suggests that generated videos can serve as reliable evaluation instances despite being visually distinguishable from real footage.

Equally importantly, advances in generative video models have narrowed the realism gap, expanded controllable 
attributes, and extended the temporal horizon of plausible 
synthesis. We expect this trajectory to continue, and this will make VGenST-Bench a scalable benchmark framework rather than a fixed collection of generated videos.

\paragraph{Task scope: why qualitative-only.} 
We deliberately restrict every base MCQ to qualitative tasks such as \textit{relative position, temporal ordering, 
visibility, identity tracking} and exclude 
quantitative-estimation tasks such as \textit{absolute distance, 
metric size} that have appeared in prior video benchmarks such 
as VSI-Bench~\citep{yang2025thinking} and STI-Bench \cite{li2025sti}. Quantitative estimates 
from monocular video are inherently noisy even for human 
observers: they admit a band of acceptable answers rather than a 
single ground truth, which lowers the human ceiling and confounds 
genuine reasoning deficits with estimation noise. By restricting 
scope, VGenST-Bench achieves a near-perfect human ceiling 
($99.0\%$), so that model--human gap
is less likely to be explained by annotation ambiguity or metric-estimation noise.
VGenST-Bench therefore \emph{complements} rather than 
replaces existing benchmarks.

\subsection{Limitations}
\label{app:Limitations}

By construction, every video in VGenST-Bench is the output of a contemporary video generator. We therefore do not claim that performance on VGenST-Bench is monotonically predictive of spatio-temporal reasoning on naturally captured video. The benchmark measures reasoning under a synthetic distribution; therefore, transfer to real-world scenarios is an empirical question rather than a guarantee. In addition, bias concerns may arise as the visual, cultural, and physical priors of the video generators may propagate into our generated videos.

\subsection{Broader Impact}
\label{app:Broader Impact}

VGenST-Bench is more than an evaluation suite — we show that generative video can be a controllable medium for building benchmarks. As video generators improve, the range of tasks we can reliably specify grows with them, making this methodology more useful over time, not less. The same pipeline could be adapted to adjacent domains where real-world capture is structurally constrained, such as autonomous driving and robotics. However, if synthetic benchmarks are widely adopted, models may become well-tuned to synthetic distributions while drifting away from real-world reasoning. We see this as motivation for the future directions discussed below, not as a reason to avoid synthetic evaluation altogether. More broadly, we hope VGenST-Bench is read less as a fixed dataset and more as evidence that generative video models can serve as a viable medium for benchmark construction. 

\subsection{Future Work}
\label{app:Future Work}

We see VGenST-Bench as an entry point rather than a destination, and identify three future directions.

\textbf{Scaling with generator capability.} As video generators support longer, higher-resolution, and more controllable outputs, the space of reliably specifiable reasoning tasks expands accordingly. Extended temporal reasoning over long videos, more fine-grained physical interactions, or scenes with many interacting agents, become tractable as generation fidelity improves. Our pipeline is designed to scale without structural changes.

\textbf{Taxonomy and hierarchy expansion.} The current 12-task taxonomy and QA hierarchy reflect deliberate authorial choices, \textit{Spatial scale}, \textit{Perspective}, and \textit{Scene dynamics}, with three temporal levels (L1/L2/L3). Additional evaluation criteria can be added, and per-cell task counts can grow as generation reliability improves.

\textbf{Domain transfer.} Our pipeline can be adapted to benchmark construction in domains where structured specifications are obtainable but real-world capture is constrained — such as autonomous driving edge cases, robotics failure modes, and surgical video. In these settings, scene-graph-driven generation offers a path to evaluating rare or safety-critical scenarios that are difficult to collect at scale.
  
\section{Extended Related Work}
\label{app:Extended Related Work}

\subsection{Image Benchmark Datasets}
\label{app:Image Benchmark Datasets}

Early efforts focused on evaluating the visual reasoning abilities of vision-language models~\cite{antol2015vqa, goyal2017making, 
hudson2019gqa, fu2023mme, li2023seed, chen2024we, kamath2023s, fu2024blink}. 
The VQA dataset established free-form visual question answering over 
natural images as a unified task formulation~\cite{antol2015vqa, 
goyal2017making}, and GQA extended this paradigm by sourcing questions 
from scene-graph annotations~\cite{hudson2019gqa}.

MME aggregates 
fourteen perception and cognition subtasks under a unified 
protocol~\cite{fu2023mme}, and SEED-Bench provides a hierarchical 
taxonomy spanning spatial and temporal understanding~\cite{li2023seed}. 
MMStar curates samples that genuinely require visual grounding by 
filtering out items solvable from the text prompt 
alone~\cite{chen2024we}. What's~Up tests left/right and above/below 
relations under controlled object placements~\cite{kamath2023s}, and 
BLINK aggregates perception tasks—including depth, multi-view 
correspondence, and relative spatial position—that humans solve 
quickly but remain difficult for current MLLMs~\cite{fu2024blink}.

While these benchmarks successfully capture core aspects of visual 
perception, they are inherently bounded by the static nature of single 
images: temporal change, motion-conditioned spatial reasoning, and 
viewpoint dynamics fall outside their scope, motivating the line of 
video benchmarks discussed next.

\subsection{Video Benchmark Datasets}
\label{app:video Benchmark Datasets}
A growing body of benchmarks has extended visual evaluation from static
images to video, where temporal reasoning becomes
central~\cite{lei2020tvqa+, yu2019activitynet, xiao2021next,
patraucean2023perception, fu2025video,
li2024mvbench, liu2024tempcompass, 
yang2025thinking, lin2025ost, li2025sti, he2025egoexobench, gong2025space10comprehensivebenchmarkmultimodal}. 
Early
video question-answering datasets such as ActivityNet-QA and TVQA+ paired
QA with activity videos and television clips
respectively~\cite{lei2020tvqa+, yu2019activitynet}. NExT-QA emphasized
causal and temporal reasoning over short clips~\cite{xiao2021next},
Perception Test
probed core perceptual skills such as memory, abstraction, physics, and
semantics over natural video~\cite{patraucean2023perception}.

In the meantime, comprehensive video evaluation suites have proliferated.
Video-MME spans short, medium, and long videos across six broad
categories with manually curated QA~\cite{fu2025video},
MVBench covers twenty temporal-understanding tasks under a unified
multiple-choice protocol~\cite{li2024mvbench}, TempCompass isolates
sensitivity to fine-grained temporal reasoning such as action order,
direction, and speed~\cite{liu2024tempcompass}.

More recent benchmarks evaluate spatio-temporal reasoning
explicitly. VSI-Bench measures visual--spatial intelligence on
egocentric indoor video, including object counting, relative distance,
and route-planning queries~\cite{yang2025thinking}. OST-Bench evaluates online spatio-temporal reasoning from the perspective of an agent incrementally exploring a scene, while STI-Bench targets precise quantitative measurement of object pose, displacement, and motion~\cite{lin2025ost, li2025sti}.
SpaCE-10 evaluates compositional spatial cognition across ten
capability dimensions~\cite{gong2025space10comprehensivebenchmarkmultimodal}. EgoExoBench evaluates
cross-viewpoint reasoning between paired egocentric and exocentric
scenes~\cite{he2025egoexobench}.

\subsection{Synthetic Benchmark Datasets}
\label{app:Synthetic Benchmark Datasets}

There have been several benchmarks that utilize synthetic data rather than real-world data \cite{johnson2017clevr, yi2019clevrer, salewski2020clevr, li2023super, wang2024compositional, girdhar2019cater, chen2025compositional, wang2025spatialviz, tang2025lego, ma20253dsrbench, wang2025spatial457, li2024videocogqa, zhao2024needle, wang20233d, zhu2026video}. The pioneering CLEVR dataset rendered 2D images from procedurally generated
3D scenes to evaluate compositional visual reasoning under fully controlled conditions~\cite{johnson2017clevr}. To extend this into the temporal domain, CLEVRER
and CATER extended this paradigm into the temporal domain by introducing
object motion and event structure~\cite{yi2019clevrer, girdhar2019cater},
while Dyn-SuperCLEVR added 4D (3D + temporal) physical dynamics within
compositional scenes~\cite{wang2024compositional}. More recent work
pushes the spatial axis itself: 3DSRBench leverages multi-view synthetic
images for higher-dimensional spatial reasoning~\cite{ma20253dsrbench},
Spatial457 evaluates 6-DoF spatial inference~\cite{wang2025spatial457},
and SpatialViz-Bench targets visual--spatial reasoning over rendered
imagery~\cite{wang2025spatialviz}.

In the video setting, VideoCogQA uses programmatic game engines to generate
videos targeting abstract cognitive tasks~\cite{li2024videocogqa},
Video-MSR evaluates multi-step spatial reasoning over dynamic
sequences~\cite{zhu2026video}, and VideoNIAH adopts a synthetic-insertion
framework that embeds unrelated visual or textual probes into video to
test long-context video retrieval~\cite{zhao2024needle}.

Some recent benchmarks employ video generative models as the construction
medium~\cite{li2025videohallu, fu2025learning}. VideoHallu and
DeeptraceReward use generated video as a data source, attaching
human annotations to characterize hallucinations and artifacts. In contrast, VGenST-Bench uses video generation as a synthesis medium for spatial reasoning: ground truth is derived directly from the scene-graph specification that drives generation, giving us controlled coverage of spatial configurations that are difficult to sample uniformly from natural video. Human verification serves as a quality filter, rather than the primary annotation step. 
\section{VGenST-Bench Details}
\label{app:vgenst_details}

\subsection{Dataset Statistics}
\label{app:dataset_statistics}

VGenST-Bench comprises \textbf{1{,}200 videos} and \textbf{33K QA pairs} that span the
$3 \times 2 \times 2$ video taxonomy (Spatial scale $\times$ Perspective $\times$ Scene dynamics)
and the three-level QA hierarchy (L1 / L2 / L3). Tab.~\ref{tab:stats_overview} summarizes the total composition of
VGenST-Bench. Each of the 12 tasks contributes 100 generated videos retained
after the two-stage human quality control. Every video is paired with multiple QA pairs whose
types are determined by the task--QA applicability matrix
(Appendix~\ref{app:Task--QA Applicability Matrix}); each base MCQ is then expanded into three
reformulation variants (Appendix~\ref{app:Question Reformulation Variants}).

\begin{table}[h]
\centering
\renewcommand{\arraystretch}{1.15}
\setlength{\tabcolsep}{6pt}
\begin{tabular}{lr|lr}
\toprule
\rowcolor[HTML]{F2F2F2}
\textbf{Statistic} & \textbf{Value} & \textbf{Statistic} & \textbf{Value} \\
\midrule
Total videos                 & 1{,}200          & Total QA pairs (all variants)     & 33{,}386 \\
Videos per task              & 100              & \quad Base MCQ                    & 9{,}707 \\
Number of tasks              & 12               & \quad V1: None-of-these distractor& 9{,}707 \\
Number of QA types           & 12               & \quad V2: None-of-these answer    & 9{,}707 \\
\;\; L1 / L2 / L3            & 3 / 6 / 3        & \quad V3: Open-ended              & 4{,}265 \\
\bottomrule
\end{tabular}
\vspace{1ex}
\caption{\textbf{Statistics of VGenST-Bench.}}
\label{tab:stats_overview}
\end{table}

\textbf{Per-task QA distribution.} Tab.~\ref{tab:stats_per_task} reports per-task counts, including the number of
applicable QA types (rows of the applicability matrix), the resulting base MCQ
count, and the total QA pairs across all reformulation variants. The task
\texttt{RV\_E\_EGO\_DYN} covers all 12 QA types and therefore yields the largest
per-task QA count.

\begin{table}[h]
\centering
\footnotesize
\renewcommand{\arraystretch}{1.10}
\setlength{\tabcolsep}{5pt}
\begin{tabular}{l|c|rrr|r}
\toprule
\rowcolor[HTML]{F2F2F2}
\textbf{Task} & \textbf{\# QA types} & \textbf{\# Videos} & \textbf{\# Base MCQ} & \textbf{\# Variants} & \textbf{Total QA} \\
\midrule
MC\_F\_EGO\_STA & 8  & 100 & 800 & 2{,}200 & 3{,}000 \\
QC\_F\_EGO\_DYN & 9  & 100 & 894 & 2{,}332 & 3{,}226 \\
CI\_F\_EXO\_STA & 8  & 100 & 800 & 2{,}000 & 2{,}800 \\
CM\_F\_EXO\_DYN & 8  & 100 & 795 & 1{,}690 & 2{,}485 \\
\midrule
DE\_V\_EGO\_STA & 7  & 100 & 700 & 1{,}750 & 2{,}450 \\
IO\_V\_EGO\_DYN & 8  & 100 & 792 & 1{,}911 & 2{,}703 \\
HO\_V\_EXO\_STA & 7  & 100 & 681 & 1{,}650 & 2{,}331 \\
VI\_V\_EXO\_DYN & 7  & 100 & 696 & 1{,}492 & 2{,}188 \\
\midrule
DS\_E\_EGO\_STA & 8  & 100 & 800 & 2{,}100 & 2{,}900 \\
RV\_E\_EGO\_DYN & 12 & 100 & 1{,}200 & 2{,}800 & 4{,}000 \\
LS\_E\_EXO\_STA & 8  & 100 & 769 & 2{,}012 & 2{,}781 \\
BT\_E\_EXO\_DYN & 8  & 100 & 780 & 1{,}742 & 2{,}522 \\
\midrule
\textbf{Total}  & --- & \textbf{1{,}200} & \textbf{9{,}707} & \textbf{23{,}679} & \textbf{33{,}386} \\
\bottomrule
\end{tabular}
\vspace{1ex}
\caption{\textbf{Per‑task statistics.}}
\label{tab:stats_per_task}
\end{table}

\textbf{Per-level distribution (L1 / L2 / L3).} Tab.~\ref{tab:stats_levels_variants} reports the QA-pair counts across the
three cognitive levels, broken down by reformulation variant. V3 (Open-Ended) contains fewer pairs because questions without a uniquely determined ground-truth answer were filtered out, as such questions cannot be reliably evaluated in free-form format.

\begin{table}[h]
\centering
\renewcommand{\arraystretch}{1.15}
\setlength{\tabcolsep}{6pt}
\begin{tabular}{l|rrr|r}
\toprule
\rowcolor[HTML]{F2F2F2}
\textbf{Variant} & \textbf{L1} & \textbf{L2} & \textbf{L3} & \textbf{Total} \\
\midrule
Base MCQ                       & 3{,}130 & 4{,}492 & 2{,}085 & 9{,}707 \\
V1 (None‑of‑these distractor)  & 3{,}130 & 4{,}492 & 2{,}085 & 9{,}707 \\
V2 (None‑of‑these answer)      & 3{,}130 & 4{,}492 & 2{,}085 & 9{,}707 \\
V3 (Open‑ended)                & 1{,}542 & 1{,}488 & 1{,}235 & 4{,}265 \\
\midrule
\textbf{Total}                 & \textbf{10{,}932} & \textbf{14{,}964} & \textbf{7{,}490} & \textbf{33{,}386} \\
\bottomrule
\end{tabular}
\vspace{1ex}
\caption{\textbf{QA‑pair distribution across cognitive levels and reformulation
variants.}}
\label{tab:stats_levels_variants}
\end{table}

\textbf{Source generative models.} Tab.~\ref{tab:stats_gen_models} reports the distribution of generative models
used to produce VGenST-Bench. Videos (1{,}200 total) span 10 distinct
video-generation models across image-to-video, text-to-video, and
start-end-to-video paradigms. Images (1{,}100 total; the
\texttt{BT\_E\_EXO\_DYN} task uses T2V and is excluded) are drawn
from 4 text-to-image models. For each video and image, we generated candidates from
multiple models and the authors performed an initial selection of the best
output based on prompt fidelity and visual quality. All author-selected
samples then underwent a Human Quality Control
(Appendix~\ref{app:Human Quality Control}).

All generative models listed in Tab.~\ref{tab:stats_gen_models} were accessed
through the AtlasCloud API\footnote{\url{https://www.atlascloud.ai}}; we adopt the AtlasCloud API's model identifier convention
(\texttt{provider/model-version/modality}, e.g.,
\texttt{bytedance/seedance-2.0-fast/image-to-video}) throughout this work to
unambiguously specify the exact model variant used.

\begin{table}[h]
\centering
\footnotesize
\renewcommand{\arraystretch}{1.15}
\setlength{\tabcolsep}{6pt}
\begin{minipage}[t]{0.52\linewidth}
\centering
\begin{tabular}{lr}
\toprule
\rowcolor[HTML]{F2F2F2}
\textbf{Video model} &  \\
\midrule
bytedance/seedance-2.0-fast (I2V) \cite{seedance2_2026}        & 571 \\
alibaba/wan-2.6-flash \cite{wan2025wanopenadvancedlargescale} & 340 \\
alibaba/wan-2.7 \cite{wan2025wanopenadvancedlargescale}                   & 88  \\
bytedance/seedance-2.0 \cite{seedance2_2026}               & 85  \\
google/veo-3.1-fast \cite{GoogleVeo3ModelPage}                & 81  \\
bytedance/seedance-2.0-fast (T2V) \cite{seedance2_2026}          & 15  \\
kwaivgi/kling-v3.0-pro \cite{kling_2024}             & 7   \\
vidu/q3-turbo \cite{vidu_2024}      & 7   \\
bytedance/seedance-v1.5-pro-fast \cite{seedance2025seedance}   & 4   \\
vidu/q3-pro \cite{vidu_2024}          & 2   \\
\midrule
\textbf{Total}                            & \textbf{1{,}200} \\
\bottomrule
\end{tabular}
\subcaption{Video generation models (all 12 tasks).}
\end{minipage}\hfill
\begin{minipage}[t]{0.45\linewidth}
\centering
\begin{tabular}{lr}
\toprule
\rowcolor[HTML]{F2F2F2}
\textbf{Image model} &  \\
\midrule
bytedance/seedream-v5.0-lite \cite{seedream_2024}      & 836 \\
google/nano-banana-2 \cite{nanobanana_2025}         & 99  \\
alibaba/wan-2.7   \cite{mao2026wan}             & 89  \\
qwen/qwen-image-2.0-pro \cite{wu2025qwen}       & 76  \\
\midrule
\textbf{Total}                      & \textbf{1{,}100} \\
\bottomrule
\end{tabular}
\subcaption{Image generation models (\texttt{BT\_E\_EXO\_DYN} excluded).}
\end{minipage}
\vspace{1ex}
\caption{\textbf{Distribution of generative models used to construct
VGenST-Bench.} I2V = image-to-video, T2V = text-to-video.}
\label{tab:stats_gen_models}
\end{table}

\subsection{Video Taxonomy}
\label{app:Video Taxonomy}

This section elaborates the cognitive foundations of the $3 \times 2 \times 2$
video taxonomy of VGenST-Bench. The taxonomy is grounded in cognitive
studies of spatial cognition and event perception, which suggest that spatial
reasoning varies along three largely independent dimensions: the
\emph{scale} of the space being reasoned about, the \emph{reference frame}
used to encode spatial relations, and whether the scene involves
\emph{static configurations or dynamic events}. Representative video frames
for each cell are shown in Fig.~\ref{fig:task_examples}.

\paragraph{Spatial scale.}
Spatial cognition has long distinguished reasoning across spatial scales,
ranging from small manipulable objects to large navigable environments
\cite{montello1993scale}. Following this view, we consider three scales:
\textit{figural}, \textit{vista}, and \textit{environmental}. Figural-scale
reasoning concerns local object configurations that can typically be
apprehended from a single view; vista-scale reasoning concerns larger scene
layouts visible from a local viewpoint; and environmental-scale reasoning
involves extended spaces that require integrating information across views,
landmarks, or navigation-like structures. This axis allows us to evaluate
whether models can generalize spatial reasoning beyond object-level relations
to broader scene- and environment-level understanding.

\paragraph{Perspective.}
Spatial relations can also be represented under different reference frames.
We distinguish between \textit{egocentric} reasoning, where spatial relations
are defined relative to the observer or camera viewpoint, and
\textit{exocentric} reasoning, where relations are defined from an external
or scene-centered viewpoint. This distinction is closely related to the
egocentric--exocentric (allocentric) distinction in cognitive psychology
\cite{klatzky1998allocentric}. By varying perspective, VGenST-Bench tests
whether models can reason not only from the visible camera-centered view but
also from alternative viewpoints or scene-level reference frames.

\paragraph{Scene dynamics.}
Finally, we distinguish between \textit{static} and \textit{dynamic} scenes.
Static scenes require reasoning over stable spatial configurations, such as
object positions, layout, and visibility. Dynamic scenes require integrating
spatial information over time, including object motion, agent--object
interactions, causal changes, and event progression. This axis is motivated
by event perception and event cognition \cite{zacks2007event}, where understanding dynamic scenes requires segmenting
and interpreting temporally evolving events rather than relying on a single
frame.

\subsection{Task Definitions}
\label{app:Task Definitions}

Combining the three axes yields 12 cells, and we design one dedicated
reasoning task per cell. We organize the descriptions below by spatial
scale (Figural / Vista / Environmental) and indicate the (perspective,
dynamics) cell with each task name. Task codes follow
Tab.~\ref{tab:tasks_def} in the main text, and representative video frames
appear in Fig.~\ref{fig:task_examples}.

\paragraph{Figural scale.}
Figural-scale tasks focus on local object configurations that can be
apprehended within a single view, probing fine-grained perception of object
attributes and relations.

\begin{itemize}
\item \textbf{MC -- Multi-Container Attribute Mapping} (Egocentric, Static).
The video shows several containers from a first-person viewpoint, each
holding objects with distinguishing attributes. The
model must map each object to its containing context.

\item \textbf{QC -- Quantity Change Tracking} (Egocentric, Dynamic).
A first-person video depicts items being added to or removed from a
workspace over time. The model must track the resulting quantities and report
the final or intermediate counts.

\item \textbf{CI -- Container Intersection Inference} (Exocentric, Static). 
The video shows two side-by-side containers from an external viewpoint, each holding several objects with distinguishing attributes. The model must determine which objects are present in both containers (the intersection) and which are unique to each container. 

\item \textbf{CM -- Causal Mapping} (Exocentric, Dynamic). 
The video depicts a scene from an external viewpoint in which an event (the \emph{trigger}) causes a subsequent state change in another object (the \emph{effector}), with multiple candidate trigger-effector pairs occurring across the timeline. The model must identify the correct cause--effect mapping linking each trigger to its corresponding consequence.
\end{itemize}

\paragraph{Vista scale.}
Vista-scale tasks involve scene layouts that span more than a single object
group but remain visible from a local viewpoint, requiring reasoning about
relative positions, directions, and visibility within a room-sized region.

\begin{itemize}
\item \textbf{DE -- Direction Estimation} (Egocentric, Static). 
A first-person video traces a path through a room-sized environment with a clear starting point and final position. After the camera reaches its destination, the model must estimate the direction of the starting point relative to the camera's current heading, expressed in clock positions.

\item \textbf{IO -- Interacted Object Identification} (Egocentric, Dynamic). 
The video shows a stationary first-person observer view. An external 
agent enters the scene, picks up one target object, and relocates it 
to another destination. The model must identify which of the three objects was relocated, requiring reasoning about both the agent's interaction and the spatial layout of all candidate objects.

\item \textbf{HO -- Height Ordering} (Exocentric, Static). 
From an external viewpoint, the video presents multiple objects of varying heights arranged within a room-sized scene. The model must order the objects by their relative heights, requiring inference of vertical extent across distinct viewpoints.

\item \textbf{VI -- Visibility Identification} (Exocentric, Dynamic). 
A fixed god-view captures two agents and a central Occluder. One agent moves around the Occluder, changing whether it is visible to the other agent. The model must determine the resulting visibility status (Visible or Occluded) from the agent's in-scene perspective---not the camera's, which sees all entities throughout.
\end{itemize}

\paragraph{Environmental scale.}
Environmental-scale tasks involve extended spaces that cannot be apprehended
from a single view, requiring integration across viewpoints, landmarks, or
navigation-like trajectories.

\begin{itemize}
\item \textbf{DS -- Directional Signage Grounding} (Egocentric, Static).
A first-person view shows an Environmental-scale space containing directional signs that indicate routes to multiple destinations. The signs include directional arrows and target labels, positioned at decision points along the path. The model must ground the signage to the underlying spatial layout and infer which direction leads to a queried target location, requiring integration of the textual/iconic content of the sign with the local geometry of the visible space.

\item \textbf{RV -- Relative Velocity Identification} (Egocentric, Dynamic).
A first-person observer moves through an Environmental-scale space while other entities also move within the same scene. The model must compare the queried entity's motion against the observer's own motion and identify the relative velocity, distinguishing it from other moving entities in the scene.

\item \textbf{LS -- Landmark Spatial Composition} (Exocentric, Static).
A top-down bird's-eye view shows three large landmarks across two phases of camera motion: a crane-up that reveals Landmark~2's compass-direction position relative to Landmark~1, followed by a long-range camera flight in a separate direction that arrives at Landmark~3. The model must compose both movements to determine the position of Landmark~3 relative to Landmark~1, expressed in eight compass directions.

\item \textbf{BT -- Behavioral Trigger Identification} (Exocentric, Dynamic).
The video shows an Environmental-scale path (road, walkway, or aisle) along which a single agent travels at constant speed before reacting to an unexpected event. A static object sits adjacent to the path as a distractor, while a dynamic hazard suddenly enters the agent's path and provokes either a full stop or wait-and-resume reaction. The model must causally link the agent's behavior change to the correct dynamic trigger rather than the static distractor.

\end{itemize}

\subsection{QA Type Definitions}
\label{app:QA Type Definitions}

Orthogonal to the video taxonomy, each video is paired with QA pairs drawn
from \textbf{12 QA types} arranged along a three-level cognitive hierarchy
that progresses from low-level perception to high-level reasoning:
\textbf{L1: Visual perception} (3 types), \textbf{L2: Scene understanding}
(6 types), and \textbf{L3: Spatio-temporal reasoning} (3 types). The
hierarchy is summarized in Tab.~\ref{tab:qa_types} of the main text; below
we provide formal definitions, evaluation intent, and a representative
example for each type. 

\paragraph{L1 -- Visual perception.}
L1 questions probe the recognition of objects and their visual attributes
from individual frames, isolating perceptual capability from any temporal or
spatial integration.

\begin{itemize}
\item \textbf{OE -- Object Existence.} 
Determine whether a specific object or entity appears in any frame of the video. 
This type isolates basic visual recognition and serves as the lowest-level perceptual probe in our hierarchy. 
\textit{Example: ``Which object appears in the video?''}

\item \textbf{OA -- Object Attribute Recognition.} 
Identify the visual attributes of a specific object in the video, such as color, material, shape, or surface pattern. 
This type targets fine-grained perceptual discrimination conditioned on a referenced object. 
\textit{Example: ``Which combination of appearances is seen among the objects in the video?''}

\item \textbf{FL -- 2D Frame Localization.} 
Identify the 2D position of an object within the camera's screen space, expressed as relative regions (e.g., left, center, right). 
This type targets perception in the image plane, without requiring inference about the underlying 3D scene. 
\textit{Example: ``In the view right after the agent appears, where is the agent located in the frame?''}
\end{itemize}

\paragraph{L2 -- Scene understanding.}
L2 questions assess the integration of perceptual cues across multiple frames into coherent spatial and temporal structures, requiring the model to relate observations distributed in time, space, or viewpoint.

\begin{itemize}
\item \textbf{IT -- Identity Tracking.} 
Determine whether two instances observed across different temporal frames, viewpoints, or environmental conditions correspond to the same underlying entity. 
This type targets cross-frame identity persistence under appearance variation. 
\textit{Example: ``The object that overtakes the camera from behind---which object is it?''}

\item \textbf{AR -- Action Recognition.} 
Identify and categorize the specific actions or events occurring within the video, requiring integration of motion cues across consecutive frames. 
\textit{Example: ``What happens to the Status Light Panel on the right after the Red Safety Button on the right is pressed?''}

\item \textbf{OC -- Object Counting.} 
Quantify the exact number of objects that satisfy specific categorical or attribute-based criteria, aggregated across the entire video. 
This type targets multi-frame perceptual aggregation rather than reasoning about change. 
\textit{Example: ``Suppose the container was empty at the start of the video---how many identical items are in the container at the end of the video?''}

\item \textbf{TO -- Temporal Ordering.} 
Determine the correct chronological sequence of multiple distinct events within the video. 
This type targets the model's ability to anchor events on a shared temporal axis. 
\textit{Example: ``Which control does the agent activate first?''}

\item \textbf{CM -- Camera Motion Recognition.} 
Recognize the camera's own motion (e.g., pan, tilt, dolly, boom, zoom, compass-aligned flight) by reasoning about how the scene transforms across frames in the absence of corresponding object-level changes. 
\textit{Example: ``In which compass direction does the camera fly horizontally?''}

\item \textbf{SL -- Spatial Layout Understanding.} 
Understand spatial relationships and the relative arrangement of objects in the scene, integrating viewpoint and depth cues to construct a coherent layout. 
\textit{Example: ``From the camera's perspective, where is the agent located relative to the launch switches?''}
\end{itemize}

\paragraph{L3 -- Spatio-temporal reasoning.}
L3 questions require higher-order inference that goes beyond what is directly observable in the video, including reasoning about novel viewpoints, hypothetical alterations, and future events. These questions evaluate the model's ability to manipulate spatio-temporal representations rather than merely recognize them.

\begin{itemize}
\item \textbf{PT -- Perspective-Taking.} 
Infer the visual representation of a scene from an unobserved, novel viewpoint not explicitly captured in the video. 
This type targets the cognitive ability to perform mental rotation and viewpoint transformation. 
\textit{Example: ``From the agent's perspective, which seismograph drum activates first?''}

\item \textbf{CR -- Counterfactual Reasoning.} 
Deduce alternative outcomes or states by hypothetically altering specific factual elements or physical conditions within the observed video. 
This type targets the ability to mentally simulate modified action sequences while keeping all other elements consistent. 
\textit{Example: ``If the camera had turned left at the sign instead of going straight, which destination would the agent be heading toward?''}

\item \textbf{PR -- Predictive Reasoning.} 
Predict the most probable subsequent events or states following the observed video, including how future scenarios will unfold under specific conditions. 
This type targets forward inference grounded in observed dynamics and physical plausibility. 
\textit{Example: ``If all objects continue at their current speeds after the video ends, where will the Yellow Tanker Truck be relative to the camera?''}
\end{itemize}

\subsection{Task--QA Applicability Matrix}
\label{app:Task--QA Applicability Matrix}

We define a \emph{task--QA applicability matrix} that specifies, for each of the 12
tasks, which QA types are evaluated.
Tab.~\ref{tab:applicability_matrix} reports the matrix. A QA pair is generated only for cells marked $\circ$ in the matrix,
ensuring that every QA pair is well-defined with respect to the underlying
video. Active (task, QA-type) cells total
\textbf{98 / 144}, and per-task counts are reported in
Tab.~\ref{tab:stats_per_task}.

\begin{table}[h]
\centering
\small
\setlength{\tabcolsep}{4pt}
\begin{tabular}{l|ccc|cccccc|ccc}
\toprule
\rowcolor[HTML]{F2F2F2}
& \multicolumn{3}{c|}{\textbf{L1}} & \multicolumn{6}{c|}{\textbf{L2}} & \multicolumn{3}{c}{\textbf{L3}} \\
\rowcolor[HTML]{F2F2F2}
\textbf{Task} & OE & OA & FL & IT & AR & OC & TO & CM & SL & PT & CR & PR \\
\midrule
MC\_F\_EGO\_STA & $\circ$ & $\circ$ & $\circ$ & $\circ$ & $\times$ & $\circ$ & $\times$ & $\circ$ & $\circ$ & $\times$ & $\circ$ & $\times$ \\
QC\_F\_EGO\_DYN & $\circ$ & $\circ$ & $\times$ & $\circ$ & $\circ$ & $\circ$ & $\circ$ & $\times$ & $\circ$ & $\times$ & $\circ$ & $\circ$ \\
CI\_F\_EXO\_STA & $\circ$ & $\circ$ & $\circ$ & $\times$ & $\times$ & $\circ$ & $\circ$ & $\circ$ & $\circ$ & $\times$ & $\circ$ & $\times$ \\
CM\_F\_EXO\_DYN & $\circ$ & $\circ$ & $\circ$ & $\times$ & $\circ$ & $\times$ & $\circ$ & $\times$ & $\circ$ & $\circ$ & $\circ$ & $\times$ \\
\midrule
DE\_V\_EGO\_STA & $\circ$ & $\circ$ & $\times$ & $\times$ & $\times$ & $\times$ & $\circ$ & $\circ$ & $\circ$ & $\circ$ & $\circ$ & $\times$ \\
IO\_V\_EGO\_DYN & $\circ$ & $\circ$ & $\circ$ & $\circ$ & $\circ$ & $\circ$ & $\times$ & $\times$ & $\circ$ & $\times$ & $\circ$ & $\times$ \\
HO\_V\_EXO\_STA & $\circ$ & $\circ$ & $\circ$ & $\times$ & $\times$ & $\times$ & $\circ$ & $\circ$ & $\circ$ & $\times$ & $\circ$ & $\times$ \\
VI\_V\_EXO\_DYN & $\circ$ & $\circ$ & $\circ$ & $\circ$ & $\circ$ & $\times$ & $\times$ & $\times$ & $\circ$ & $\circ$ & $\times$ & $\times$ \\
\midrule
DS\_E\_EGO\_STA & $\circ$ & $\circ$ & $\circ$ & $\times$ & $\times$ & $\circ$ & $\times$ & $\circ$ & $\circ$ & $\circ$ & $\circ$ & $\times$ \\
RV\_E\_EGO\_DYN & $\circ$ & $\circ$ & $\circ$ & $\circ$ & $\circ$ & $\circ$ & $\circ$ & $\circ$ & $\circ$ & $\circ$ & $\circ$ & $\circ$ \\
LS\_E\_EXO\_STA & $\circ$ & $\circ$ & $\times$ & $\times$ & $\times$ & $\times$ & $\circ$ & $\circ$ & $\circ$ & $\circ$ & $\circ$ & $\circ$ \\
BT\_E\_EXO\_DYN & $\circ$ & $\circ$ & $\times$ & $\circ$ & $\circ$ & $\times$ & $\circ$ & $\circ$ & $\times$ & $\circ$ & $\circ$ & $\times$ \\
\bottomrule
\end{tabular}

\vspace{1ex}

\caption{
\textbf{Task--QA Applicability Matrix.} Rows: 12 task types of VGenST-Bench, grouped by spatial scale (Figural / Vista / Environmental). Columns: 12 QA types grouped by reasoning level.
$\circ$: category is applied to the task; $\times$: not applicable.
\textbf{L1 (Visual Perception):} OE = Object Existence, OA = Object Attribute Recognition, FL = 2D Frame Localization.
\textbf{L2 (Scene Understanding):} IT = Identity Tracking, AR = Action Recognition, OC = Object Counting, TO = Temporal Ordering, CM = Camera Motion Recognition, SL = Spatial Layout Understanding.
\textbf{L3 (Spatio-Temporal Reasoning):} PT = Perspective-Taking, CR = Counterfactual Reasoning, PR = Predictive Reasoning.
}
\label{tab:applicability_matrix}
\end{table}

\subsection{Question Reformulation Variants}
\label{app:Question Reformulation Variants}

Multiple-choice evaluation would be vulnerable to
option-level shortcuts. To evaluate more fine-grained spatio-temporal
reasoning, every base MCQ in VGenST-Bench is expanded into three
reformulation variants.

\paragraph{Base MCQ.}
The base question is a \emph{N}-way multiple-choice question with one correct
answer and other distractors. Distractors are generated to be
semantically plausible so that random guessing is
not trivially defeated.

\paragraph{V1 -- None-of-these Distractor.}
The base question is augmented with an additional incorrect option,
``\textit{None of these}''. The correct answer is still present among the
options. \emph{Intent:} verify that the model commits to the correct choice
even when an explicit ``reject'' option is available, instead of defaulting
to ``None of these'' under uncertainty.

\paragraph{V2 -- None-of-these Answer.}
The correct option is removed, and ``\textit{None of these}'' is
introduced as the correct answer; all remaining listed options are
distractors. \emph{Intent:} test whether the model can reject all
listed options when none of them is correct, rather than selecting the
most plausible distractor. 

\paragraph{V3 -- Open-ended.}
All options are removed and the question is presented in free-form. The
model's response is scored against the ground-truth answer using an
LLM-as-judge protocol.
\emph{Intent:} eliminate all option-level priors so that performance
reflects the model's ability to produce, rather than merely select, the
correct answer.

Tab.~\ref{tab:stats_levels_variants} reports the QA-pair counts for each
variant across the cognitive levels. Examples of all four formats for a
single underlying QA pair are shown in
Fig.~\ref{fig:pipeline}.

\subsection{Theme Diversity}
\label{app:Theme Diversity}

\begin{wrapfigure}{r}{0.36\textwidth}
  \vspace{-10pt} 
  \centering
  \includegraphics[width=\linewidth]{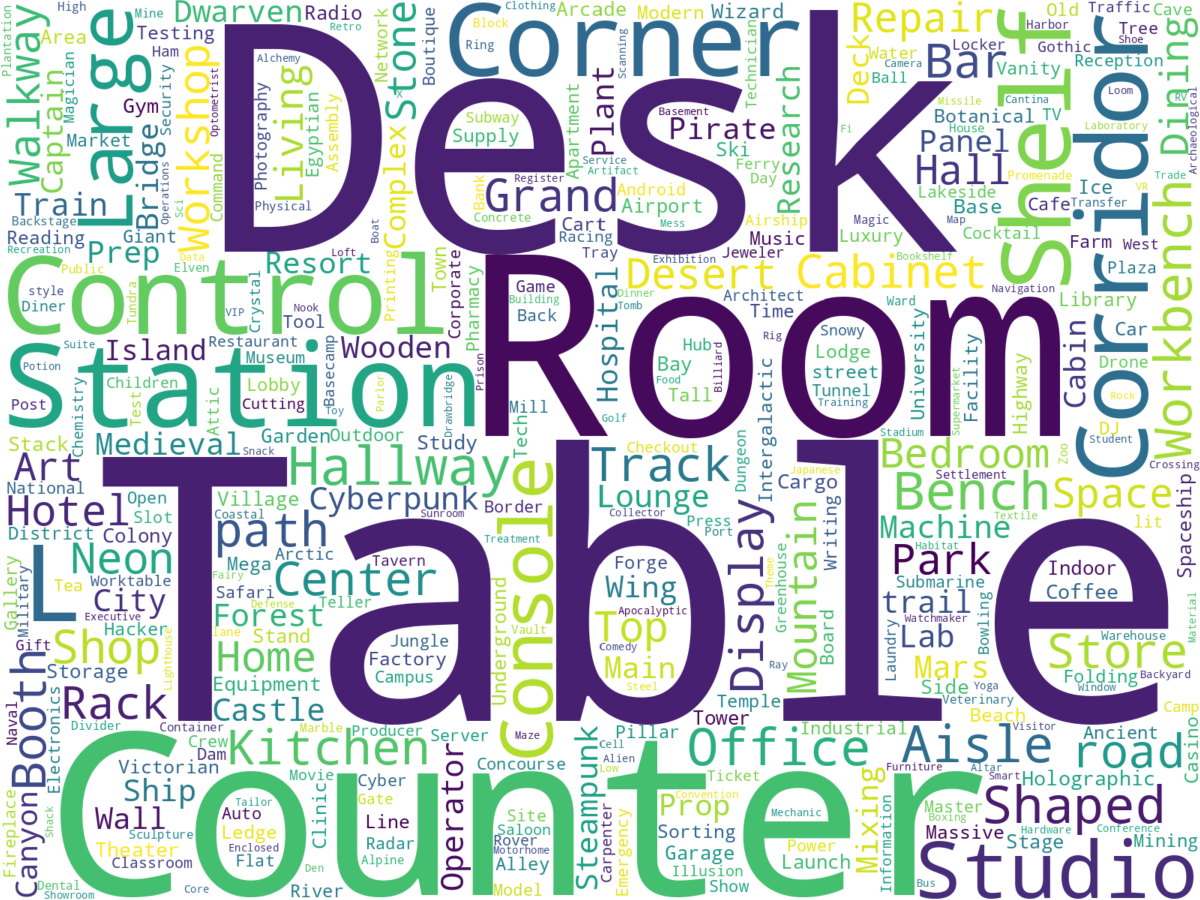}
  \caption{\textbf{Word cloud of the 1,000 themes} in VGenST-Bench.}
  \label{fig:theme_wordcloud}
  \vspace{-20pt} 
\end{wrapfigure}

To maximize visual diversity, VGenST-Bench draws scenarios from a curated pool of \textbf{themes} that specify the visual and semantic context of each video. For each of the tasks in our taxonomy, we manually identified \textbf{10 theme categories} that are semantically compatible with the task's required scene properties, spanning everyday, industrial, sci-fi, and fantasy domains. For \texttt{Relative\_Velocity\_Identification (RV)}, e.g.,
\begin{itemize}[leftmargin=*, nosep]
  \item \texttt{C1.}~Highways \& Freeways~\textit{(Desert Highway, Suspension Bridge, \dots)}
  \item \texttt{C2.}~Racetracks \& Motorsports~\textit{(Formula Track, Drag Strip, \dots)}
  \item \texttt{\;\;\dots~(C3--C10)}
\end{itemize}

Within each category, we enumerated 10 themes with distinct visual identities. Since tasks with identical spatial scale and perspective share the same visual context, the \textbf{QC} and \textbf{CI} tasks share the theme pool of the \textbf{MC} task. Consequently, we maintain 10 unique theme pools, resulting in $10 \text{ pools} \times 100 \text{ themes} = 1{,}000$ unique themes across the full benchmark.
\section{Construction Pipeline Details}
\label{app:Construction Pipeline Details}

\subsection{Pipeline Overview}
\label{app:Pipeline Overview}

\begin{figure}[t]
\centering
\includegraphics[width=1.0\textwidth]{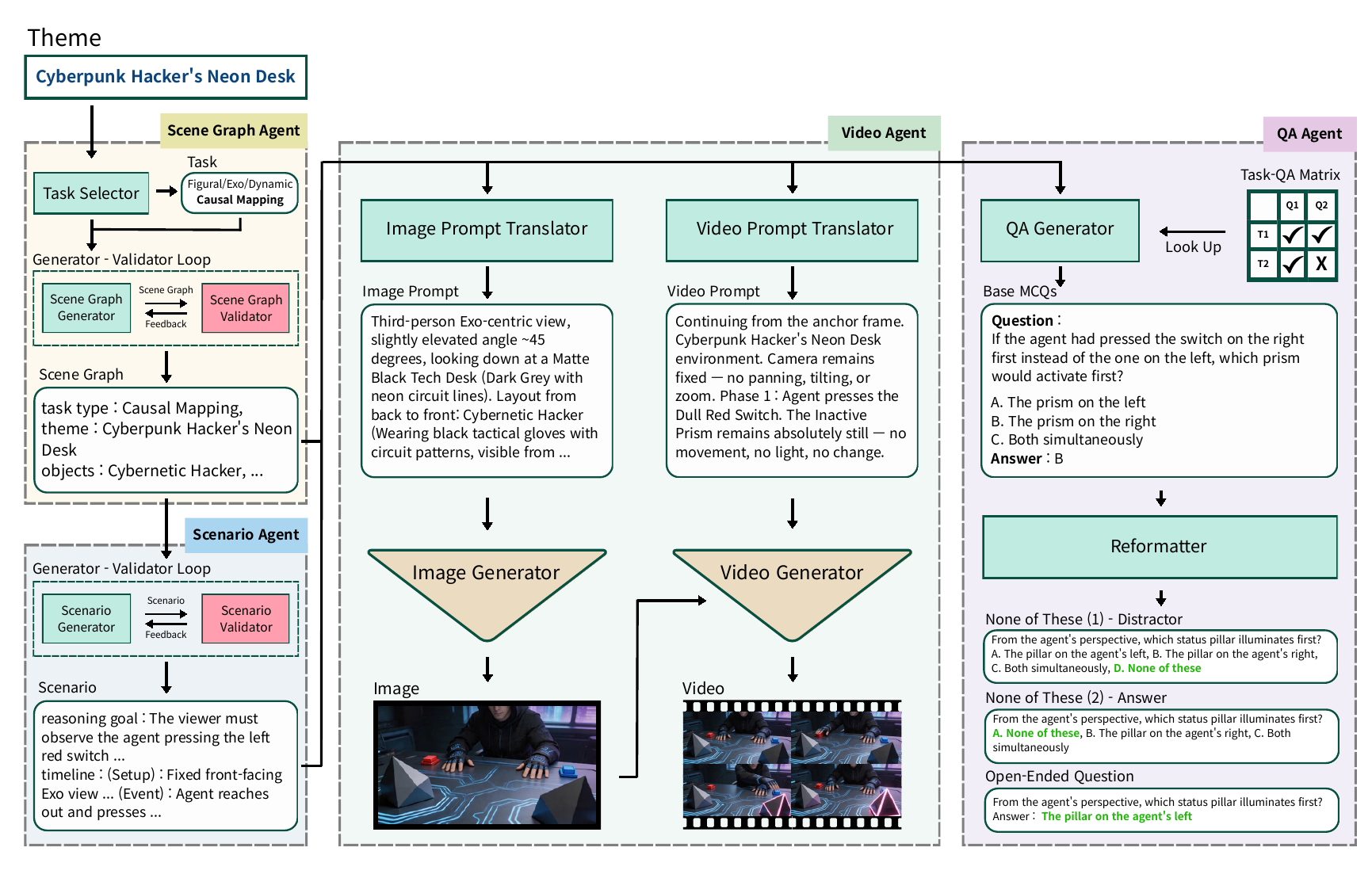}

\vspace{-1ex}

\caption{\textbf{Construction Pipeline of VGenST-Bench.}}

\vspace{-2ex}

\label{fig:pipeline_app}
\end{figure}

The full multi-agent construction pipeline of VGenST-Bench is illustrated in Fig.~\ref{fig:pipeline_app}. 
Four agents collaborate across multiple stages---scene graph generation, scenario generation, video generation, and QA generation. We provide a detailed description of each agent and an end-to-end trace of our pipeline.

Tab.~\ref{tab:pipeline_models} lists the specific models used by each agent of the pipeline. 
We deliberately separate the QA Generator (Claude-Opus-4.7) from the evaluated MLLMs by excluding the entire Claude family from our evaluation, preventing self-evaluation bias.

\begin{table}[h]
\centering

\footnotesize
\setlength{\tabcolsep}{6pt}
\renewcommand{\arraystretch}{1.2}
\begin{tabular}{l|l|l}
\toprule
\rowcolor[HTML]{F2F2F2}
\textbf{Agent} & \textbf{Sub-agents} & \textbf{Models} \\
\midrule
\multirow{3}{*}{Scene Graph Agent}
  & Task-selector            & - \\
  & Generator                & Claude-Opus-4.7 \\
  & Validator                & GPT-5-mini \\
\midrule
\multirow{2}{*}{Scenario Agent}
  & Generator                & Claude-Opus-4.7 \\
  & Validator                & GPT-5-mini \\
\midrule
\multirow{4}{*}{Video Generation Agent}
  & Image Prompt Translator  & Claude-Opus-4.7 \\
  & Image Generator     & - \\
  & Video Prompt Translator  & Claude-Opus-4.7 \\
  & Video Generator   & - \\
\midrule
\multirow{2}{*}{QA Generation Agent}
  & QA Generator             & Claude-Sonnet-4.6 \\
  & Reformatter (V1/V2/V3)   & Claude-Sonnet-4.6 \\
\bottomrule
\end{tabular}

\vspace{1ex}
\caption{\textbf{Base LLMs used at each agent of the VGenST-Bench construction pipeline.}}
\label{tab:pipeline_models}
\end{table}

\subsection{Detailed Agent Descriptions}
\label{app:detailed_agent_descriptions}

This section describes the internal structure and behavior of each agent in the construction pipeline. All LLM-based sub-agents in our pipeline are guided by few-shot prompting: each generator and validator is prompted with a small number of task-specific examples (e.g., reference scene graphs, scenarios, or QA pairs for the same task type) drawn from a curated pool of verified outputs. These examples provide stylistic and structural guidance without constraining content, allowing the agents to adapt to new themes while maintaining consistency with the task specification.

\textbf{Scene Graph Agent.}
The Scene Graph Agent transforms an input theme into a structured scene graph that specifies the static spatial composition of the scene. The agent consists of three sub-agents operating in sequence. Each task definition and associated rules are passed as input to both the generator and the validator.

(i) \emph{Task Selector} examines the input theme and determines which of the 12 tasks in our taxonomy is most appropriate for that theme. The selector returns a single task assignment (e.g., \texttt{MC\_F\_EGO\_STA}). When constructing VGenST-Bench, we used a curated set of predefined themes (Appendix~\ref{app:Theme Diversity}). Task Selector is therefore bypassed during benchmark construction
and is intended for general-purpose use of the pipeline on free-form themes.

(ii) \emph{Scene Graph Generator} produces a candidate JSON format scene graph that satisfies both the theme and the selected task's constraints. The output is a structured representation listing entities, attributes, and spatial relations required by the task (e.g., for \textit{Direction Estimation}, the generator must specify a starting landmark, a final landmark, and the camera path connecting them).

(iii) \emph{Scene Graph Validator} verifies that the candidate scene graph satisfies all task-defined rules such as ambiguity, task compliance and the consistency between attributes and theme. If any rule is violated, the validator returns a structured rejection signal indicating which rule failed, and the Scene Graph Generator is invoked again with this feedback. This generator--validator loop repeats until the validator accepts the scene graph or the maximum iteration count is reached.

\textbf{Scenario Agent.}
The Scenario Agent transforms a validated scene graph into a structured scenario, which specifies the temporal unfolding of the scene as an ordered sequence of phases (e.g., setup, event, result). The agent operates under a generator--validator loop with two sub-agents.

(i) \emph{Scenario Generator} consumes the scene graph along with the task definition, task rules, and task guidelines, and produces a JSON format scenario containing a phase-by-phase \texttt{timeline}. The first phase of the timeline is constrained to match the task's \emph{First Frame Specification}, ensuring consistency with the anchor frame used in the next stage.

(ii) \emph{Scenario Validator} audits the generated scenario against five criteria: object and attribute fidelity to the scene graph, first-frame compliance, temporal flow accuracy (for dynamic tasks), ground-truth determinacy (the described phases must yield the ground truth as the only viewer-deducible answer), and compliance with all task rules and scene-dynamics constraints. Validation failures return structured feedback indicating which criterion was violated, triggering targeted regeneration. The loop terminates upon acceptance or after the maximum iteration count is reached.

\textbf{Video Agent.}
The Video Agent transforms a validated scenario into a video. Unlike the preceding agents, this agent does not employ a generator--validator loop. Instead, it consists of four sub-agents organized as a two-stage \emph{prompt-then-generate} pipeline.

(i) \emph{Image Prompt Translator} consumes the scene graph and scenario together with the task's \emph{First Frame Specification}, and then produces a text-to-image (T2I) prompt that captures the exact static state at which the video begins. The translator is instructed to derive the first frame strictly from the scenario, to translate any actions into static poses, and to adapt the prompt style to the scene's perspective.

(ii) \emph{Image Generator} renders the T2I prompt into the \emph{anchor frame}, a single high-fidelity image that serves as the first frame of the video. We use state-of-the-art T2I models accessed via the AtlasCloud API.

(iii) \emph{Video Prompt Translator} consumes the scene graph, the full scenario timeline, and the anchor frame, and produces an image-to-video (I2V) prompt that describes the motion and changes occurring after the anchor frame. For \emph{Static} scenes, the prompt restricts motion to camera trajectory while explicitly enforcing that all objects remain frozen; for \emph{Dynamic} scenes, the prompt additionally maps each description in the timeline to a precise temporal action.

(iv) \emph{Video Generator} renders the I2V prompt, conditioned on the anchor frame, into the final video. We use state-of-the-art I2V models accessed via the AtlasCloud API. As an exception, videos for the \texttt{Behavioral\_Trigger\_Identification (BT)} task are generated using a text-to-video (T2V) model directly, without an anchor frame. 
We initially adopted the same I2V approach as for other tasks but found that conditioning on a pre-rendered anchor frame consistently degraded the quality of the trigger event. Therefore, we replace the I2V step with a direct T2V generation, which yields substantially more faithful renderings of the trigger--reaction sequence.

\textbf{QA Agent.}
The QA Agent transforms the validated scene graph and scenario into the final QA set used for evaluation. The agent consists of two sub-components operating in sequence.

(i) \emph{QA Generator} produces a base multiple-choice question (MCQ) for each \emph{(task, QA-type) combination} permitted by the applicability matrix (Appendix~\ref{app:Task--QA Applicability Matrix}). The generator is conditioned on the task specification, the QA-type definition, the scene graph, the scenario, and 1--3 prototype QAs that calibrate phrasing style and difficulty. To prevent hallucination, it is constrained to produce only questions answerable from the scene graph itself, with distractors drawn from other scene graphs. The number of answer options per MCQ is not fixed and is chosen to best suit each question (typically four).

(ii) \emph{Reformatter} converts each base MCQ into the three reformulation variants (Appendix~\ref{app:Question Reformulation Variants}). \textbf{V1 (None-of-these distractor)} appends ``None of these'' as an additional incorrect option while preserving the original correct answer. \textbf{V2 (None-of-these answer)} replaces the text of the correct option with ``None of these.'' \textbf{V3 (Open-Ended)} strips all options from the prompt and stores the original correct text as the expected answer for LLM-as-judge scoring. Note that V3 is rejected when the answer is not naturally expressible as a number or short phrase, since longer answers do not admit reliable string-level matching.

\subsection{Human Quality Control}
\label{app:Human Quality Control}

\newcommand{\hmap}[1]{\cellcolor{green!#1!red!25}#1}

To ensure the quality of VGenST-Bench, we conduct a two-stage human quality control covering both the generated videos and the question--answer pairs. This section details the verification protocol, the per-stage retention results and inter-annotator 
agreement analysis.

\textbf{Validator pool.} We recruit twelve validators 
(indexed $0$ through $11$), all graduate students with bachelor’s degrees in Computer Science. 
Each annotator completes a calibration session on a held-out set 
of video--QA pairs before beginning the main verification, ensuring 
consistent application of the rejection criteria described below. We adopt a pairing scheme over the twelve annotators to 
distribute workload evenly while preserving cross-task consistency. 
Task $t \in \{0, 1, \dots, 11\}$ is reviewed by the pair 
$\bigl(t \bmod 12,\ (t{+}1) \bmod 12\bigr)$, so that each annotator 
participates in exactly two adjacent tasks and shares one annotator 
with each neighboring pair. 

\textbf{Two-stage pipeline.} Quality control proceeds sequentially:
\begin{itemize}
  \item \textbf{Stage 1 -- Video QC.} For each task, the assigned pair 
    independently reviews all 100 generated videos. Rejection criteria 
    include (i) physical implausibility unrelated to the task design, 
    (ii) generation artifacts that compromise answerability (object 
    morphing, identity swaps, severe flicker), and (iii) prompt--video 
    drift that invalidates the intended scenario.
  \item \textbf{Stage 2 -- QA QC.} For videos that pass Stage 1, the 
    same pair independently reviews each base MCQ. Rejection criteria 
    include (i) incorrect ground-truth answer, (ii) ambiguous or 
    multiply-correct options, (iii) single-frame solvability for 
    L3 questions and (iv) language-prior shortcuts solvable without the 
    video.
\end{itemize}

\textbf{Decision rule and split resolution.} We apply an \emph{intersection rule}: an item is retained only if both annotators independently mark it as valid. When the two annotators disagree (``split'' cases, exactly one rejection), the authors resolve the case by jointly re-inspecting the item.

\begin{table}[h]
  \centering
  \small
  
  \begin{tabular}{lcccc}
    \toprule
    \rowcolor[HTML]{F2F2F2}
    \textbf{Spatial Scale} & \textbf{EGO\_STA} & \textbf{EGO\_DYN} 
                          & \textbf{EXO\_STA} & \textbf{EXO\_DYN} \\
    \midrule
    Figural (F)        & 95 (MC) & 93 (QC) & 95 (CI) & 90 (CM) \\
    Vista (V)          & 89 (DE) & 94 (IO) & 97 (HO) & 87 (VI) \\
    Environmental (E)  & 91 (DS) & 86 (RV) & 78 (LS) & 90 (BT) \\
    \bottomrule
  \end{tabular}

  \vspace{1ex}

  \caption{\textbf{Stage 1 -- Video Quality Control.} 
  For each of the 12 tasks, 100 generated videos are independently 
  reviewed by the assigned pair. Cells report the number of videos 
  retained (out of 100).}

\label{tab:qc-video}
  
\end{table}

\begin{table}[h]
  \centering
  \footnotesize
  \setlength{\tabcolsep}{4pt}
  \renewcommand{\arraystretch}{1.15}

  \begin{tabular}{l|ccc|cccccc|ccc}
    \toprule
    \rowcolor[HTML]{F2F2F2}
    & \multicolumn{3}{c|}{\textbf{L1}}
    & \multicolumn{6}{c|}{\textbf{L2}}
    & \multicolumn{3}{c}{\textbf{L3}} \\
    \rowcolor[HTML]{F2F2F2}
    \textbf{Task} & OE & OA & FL & IT & AR & OC & TO & CM & SL & PT & CR & PR \\
    \midrule
    MC\_F\_EGO\_STA & 0 & 0 & 0 & 0 & -- & 0 & -- & 0 & 0 & -- & 0 & -- \\
    QC\_F\_EGO\_DYN & 0 & \hmap{6} & -- & 0 & 0 & 0 & 0 & -- & 0 & -- & 0 & 0 \\
    CI\_F\_EXO\_STA & 0 & 0 & 0 & -- & -- & 0 & 0 & 0 & 0 & -- & 0 & -- \\
    CM\_F\_EXO\_DYN & \hmap{2} & \hmap{3} & 0 & -- & 0 & -- & 0 & -- & 0 & 0 & 0 & -- \\
    \midrule
    DE\_V\_EGO\_STA & 0 & 0 & -- & -- & -- & -- & 0 & 0 & 0 & 0 & 0 & -- \\
    IO\_V\_EGO\_DYN & 0 & \hmap{8} & 0 & 0 & 0 & 0 & -- & -- & 0 & -- & 0 & -- \\
    HO\_V\_EXO\_STA & \hmap{1} & \hmap{18} & 0 & -- & -- & -- & 0 & 0 & 0 & -- & 0 & -- \\
    VI\_V\_EXO\_DYN & \hmap{4} & 0 & 0 & 0 & 0 & -- & -- & -- & 0 & 0 & -- & -- \\
    \midrule
    DS\_E\_EGO\_STA & 0 & 0 & 0 & -- & -- & 0 & -- & 0 & 0 & 0 & 0 & -- \\
    RV\_E\_EGO\_DYN & 0 & 0 & 0 & 0 & 0 & 0 & 0 & 0 & 0 & 0 & 0 & 0 \\
    LS\_E\_EXO\_STA & 0 & \hmap{12} & -- & -- & -- & -- & 0 & 0 & \hmap{6} & \hmap{3} & \hmap{10} & 0 \\
    BT\_E\_EXO\_DYN & 0 & \hmap{16} & -- & 0 & \hmap{2} & -- & 0 & 0 & -- & \hmap{2} & 0 & -- \\
    \bottomrule
  \end{tabular}

    \vspace{1ex}

  \caption{\textbf{Stage 2 -- QA Quality Control: Reject Matrix.}
  Each cell shows the number of base MCQs (out of 100) \emph{rejected}
  during QA validation. Dashes (\textemdash) indicate task--QA combinations not produced for this task.}


\label{tab:qc-qa}

\end{table}

\begin{table}[!h]
  \centering
  \small

  \begin{tabular}{lcc}
  \rowcolor[HTML]{F2F2F2}
    \toprule
                                  & \textbf{Video QC} & \textbf{QA QC} \\
    \midrule
    Items reviewed                & 1{,}200           & 9{,}800 \\
    Both pass                     & 1{,}045 (87.1\%)  & 9{,}677 (98.7\%) \\
    Both reject                   &    70 (5.8\%)     &    80 (0.8\%) \\
    Split (author-resolved)       &    85 (7.1\%)     &    43 (0.4\%) \\
    \quad $\hookrightarrow$ accepted &  40             &    30 \\
    \quad $\hookrightarrow$ rejected &  45             &    13 \\
    \midrule
    Raw agreement rate            & 92.9\%            & 99.6\% \\
    Cohen's $\kappa$              & 0.58              & 0.79 \\
    \bottomrule
  \end{tabular}

  \vspace{1ex}

  \caption{\textbf{Inter-Annotator Agreement Statistics.}
  Aggregated across all task--QA combinations applicable per the
  Applicability Matrix.}
  \label{tab:qc-agreement}

  \vspace{-5ex}
\end{table}

\textbf{Stage 1: Video Quality Control.} Tab.~\ref{tab:qc-video} reports the number of videos retained per task after two-pass human review. Across all 1{,}200 generated videos, \textbf{90.4\%} pass both reviews, with per-task retention ranging from \textbf{78\% to 97\%}. The lowest retention is observed for \texttt{LS\_E\_EXO\_STA} (\textbf{78}/100). The high overall pass rate is due to a preliminary author-level filtering step that removes clearly defective clips before human QC. Videos rejected in Stage 1 are subsequently regenerated against the same scene graph, yielding the final balanced benchmark of 1{,}200 videos.

\textbf{Stage 2: QA Quality Control.} Tab.~\ref{tab:qc-qa} presents the full $12 \times 12$ reject matrix for QA quality control. Out of \textbf{9{,}800} reviewed MCQs across the 98 active applicability cells, only \textbf{93} are rejected. The high pass rate reflects our QA generation strategy: each MCQ is produced by a QA generator conditioned on the scene graph and scenario as ground-truth references, with few-shot exemplars guiding the question template, distractor selection, and answer-derivation patterns. Most remaining rejections are concentrated in L1 Object Attribute Recognition (OA), where the queried attribute makes the decision ambiguous. Rejected QA pairs are simply dropped, leaving a final total of \textbf{9{,}707} base MCQs across the benchmark.

\textbf{Inter-Annotator Agreement.}
Tab.~\ref{tab:qc-agreement} aggregates inter-annotator agreement
across both QC stages. The \emph{both-pass} and \emph{both-reject}
counts capture cases in which the two annotators reached the same
decision independently; their sum normalized by the total yields the
\emph{raw agreement rate}. The \emph{split} count records cases sent
to author resolution, broken down by accepted versus rejected
outcomes. We additionally report Cohen's $\kappa$. For Video QC, raw agreement reaches \textbf{92.9\%}, with
$\kappa=\textbf{0.58}$. The QA QC stage exhibits an even stronger
agreement, with raw agreement of \textbf{99.6\%},
$\kappa=\textbf{0.79}$, consistent
with the constrained few-shot QA generation strategy that yields
well-formed MCQs.

\section{Video Quality Human Study}
\label{app:video_quality}

To support our claim that generated videos can serve as a viable 
evaluation medium for spatial reasoning, we conduct a blind human 
study comparing the perceptual quality of VGenST-Bench videos against 
established video benchmarks.

\textbf{Setup.} We compare four video sources: \textbf{VGenST-Bench 
(ours)}, \textbf{VSI-Bench}~\cite{yang2025thinking}, 
\textbf{EgoExoBench}~\cite{he2025egoexobench}, and \textbf{Perception 
Test}~\cite{patraucean2023perception}. We construct $N=50$ comparison 
sets, where each set contains one clip from each of the four sources 
(50 sets $\times$ 4 sources $=$ 200 clips total). For every clip, we 
extract $K=8$ frames following the same uniform sampling protocol 
used during model evaluation.

\textbf{Evaluator pool.} We recruit three evaluators (indexed $12$ through $14$), with no 
background in computer vision or generative video research. This non-expert pool reflects the 
perspective of a general viewer and avoids prior familiarity with 
known generation artifacts.

\textbf{Evaluation protocol.} For each comparison set, evaluators 
complete two tasks:

\begin{itemize}
  \item \textbf{Ordinal ranking.} Evaluators rank the four clips 
    (1 = best, 4 = worst) along three axes:
    \begin{itemize}
      \item \textit{Photorealism} -- how realistic the visual 
        appearance is.
      \item \textit{Temporal coherence} -- how consistent objects, 
        identities, and scene layout remain across frames.
      \item \textit{Scene comprehensibility} -- how clearly the 
        objects, spatial layout, and actions can be understood.
    \end{itemize}
\item \textbf{Real-vs-fake judgment.} Evaluators are informed that each clip is either real video or AI-generated, but are not 
told how many of the four are generated. They then independently 
label each clip as ``real'' or ``fake.'' This protocol allows us 
to measure not only whether VGenST-Bench clips are correctly 
flagged as generated (recall), but also the rate at which real 
benchmark clips are mistakenly judged as fake 
(false-positive rate), which serves as a calibration baseline 
for the difficulty of frame-strip-based authenticity judgment.
\end{itemize}

\textbf{Metrics.} We report (i) the mean rank of each source on 
each axis (lower is better), and (ii) for the real-vs-fake task, the 
fraction of clips from each source labeled as ``fake'' by evaluators. We additionally report Fleiss' $\kappa$ across the three evaluators on the binary real-vs-fake judgment to quantify inter-annotator consistency.

\begin{table}[h]
  \centering
  \small
  \setlength{\tabcolsep}{8pt}
  \renewcommand{\arraystretch}{1.2}

  \begin{tabular}{lccc}
    \toprule
    \rowcolor[HTML]{F2F2F2}
    \textbf{Source} & \textbf{Photorealism} & \textbf{Temporal Coherence}
                    & \textbf{Comprehensibility} \\
    \midrule
    Perception Test~\cite{patraucean2023perception}
        & 1.84 $\pm$ 0.90 & 2.01 $\pm$ 1.02 & 2.31 $\pm$ 1.13 \\
    EgoExoBench~\cite{he2025egoexobench}
        & 2.13 $\pm$ 1.01 & 2.24 $\pm$ 1.08 & 2.18 $\pm$ 1.05 \\
    VSI-Bench~\cite{yang2025thinking}
        & 2.39 $\pm$ 0.93 & 2.47 $\pm$ 1.02 & 2.47 $\pm$ 1.04 \\
    \midrule
    \textbf{VGenST-Bench (ours)}
        & 3.64 $\pm$ 0.67 & 3.27 $\pm$ 0.94 & 3.04 $\pm$ 1.07 \\
    \bottomrule
  \end{tabular}

  \vspace{1ex}

  \caption{\textbf{Ordinal Ranking Results.}
  Mean rank ($\pm$ std) pooled across $N=50$ comparison sets and three
  evaluators ($150$ observations per cell, lower is better, range $1$--$4$).}
  \label{tab:video-quality-ranking}
\end{table}

\begin{table}[h]
  \centering
  \small
  \setlength{\tabcolsep}{8pt}
  \renewcommand{\arraystretch}{1.2}

  \begin{tabular}{lcccc}
    \toprule
    \rowcolor[HTML]{F2F2F2}
    \textbf{Source} & \textbf{Ground Truth}
                    & \textbf{\% Judged Fake}
                    & \textbf{E\_12 / E\_13 / E\_14}
                    & \textbf{Fleiss' $\kappa$} \\
    \midrule
    Perception Test  & Real & 12.7\% &  8 /  5 /  6 & 0.64 \\
    EgoExoBench      & Real & 18.0\% &  8 /  9 / 10 & 0.64 \\
    VSI-Bench        & Real & 26.0\% & 12 / 13 / 14 & 0.52 \\
    \midrule
    \textbf{VGenST-Bench (ours)} & Fake & \textbf{63.3\%}
                                 & 32 / 29 / 34 & 0.66 \\
    \bottomrule
  \end{tabular}

  \vspace{1ex}

  \caption{\textbf{Real-vs-Fake Judgment.}
  Per-source results from the binary real/fake task. \textbf{\% Judged Fake}
  is the fraction of judgments labeled ``fake'' across $50$ clips $\times$ 3
  evaluators ($150$ judgments per source). \textbf{E\_12 / E\_13 / E\_14}
  reports each evaluator's raw count of ``fake'' labels (out of $50$).
  \textbf{Fleiss' $\kappa$} is computed per source over $50$ items and three
  evaluators. Overall $\kappa$ pooled across all $200$ items is $0.68$.}
  \label{tab:video-quality-deception}
\end{table}

\textbf{Results.} Tab.~\ref{tab:video-quality-ranking} reports the ordinal ranking results. As expected, VGenST-Bench receives the lowest mean rank on \textit{photorealism} ($3.64$), reflecting the residual visual gap between current generative video models and real-world capture. The gap to the worst real-source baseline narrows monotonically across the three axes: $1.25$ ranks on photorealism ($3.64$ vs.\ $2.39$), $0.80$ ranks on \textit{temporal coherence} ($3.27$ vs.\ $2.47$), and $0.57$ ranks on \textit{scene comprehensibility} ($3.04$ vs.\ $2.47$). Photorealism is thus the dimension on which VGenST clips lag most clearly, while comprehensibility is closest to real video. 

Tab.~\ref{tab:video-quality-deception} reports the real-vs-fake judgments. VGenST-Bench clips are flagged as fake in $63.3\%$ of cases, well above the $12.7$--$26.0\%$ false-positive rate observed on the three real-source baselines. As we expected, evaluators reliably identify our clips as generated. However, considering that even genuine benchmark video is mistaken for AI output at a notable rate, the absolute $63.3\%$ should not be read as an indicator of complete failure. Inter-annotator agreement is consistent across sources (per-source Fleiss' $\kappa = 0.52$--$0.66$, overall $\kappa = 0.68$ pooled across all $200$ items).

\textbf{Discussion.} We discuss the findings from the two tasks separately. The ranking task asks whether our clips remain perceptually usable as an \emph{evaluation testbed} for spatio-temporal reasoning. Here the relevant criterion is scene comprehensibility, not photorealism: a benchmark clip need not look very real; it just needs to convey object identity, spatial layout, and action clearly enough to support the underlying task. The real-vs-fake task asks whether our clips are perceptibly synthetic to a viewer: the answer is yes, and VGenST-Bench is not designed to deceive. Visual realism is the explicit cost we pay in exchange for systematic control over taxonomy, scene graphs, and scenarios across the applicability matrix. As discussed in our limitations, this visual gap is expected to resolve as video models continue to evolve. Overall, our results demonstrate that generative models can serve as a valid and effective foundation for spatio-temporal benchmark creation.

\section{Human Annotator Details}
\label{app:Human Annotator Details}
VGenST-Bench involves 25 human annotators in total, all of whom participated voluntarily. All participants were informed of the purpose of the study, the nature of their tasks, and the intended use of their judgments prior to participation, and provided informed consent. No personally identifying information was collected. Annotators are partitioned into three mutually disjoint groups---no individual contributed to more than one stage---each supporting a different stage of the benchmark.

\begin{table}[h]
  \centering
  \footnotesize
  \setlength{\tabcolsep}{5pt}
  \renewcommand{\arraystretch}{1.3}
  \begin{tabular}{l|c|l|l}
    \toprule
    \rowcolor[HTML]{F2F2F2}
    \textbf{Group} 
      & \textbf{N} 
      & \textbf{Background} 
      & \textbf{Task}  \\
    \midrule
    Quality control
      & 12 
      & CS, B.S.\ holders 
      & Two-stage video and QA review (intersection rule)\\
    \addlinespace[2pt]
    Video quality study
      &  3 
      & Non-CS, non-expert 
      & Ordinal ranking and real-vs-fake judgment \\
    \addlinespace[2pt]
    Human baseline
      & 10 
      & Non-CS, diverse 
      & 120 videos per annotator under circular eval \\
    \midrule
    \textbf{Total} 
      & \textbf{25} 
      & \multicolumn{2}{l}{\emph{All groups mutually disjoint; voluntary participation; informed consent obtained}} \\
    \bottomrule
  \end{tabular}
  \vspace{1ex}
  \caption{\textbf{Human annotator pool for VGenST-Bench.}}
  \label{tab:annotators}
\end{table}

\textbf{QC annotators (12 participants).}
For the two-stage quality control stage described in 
Appendix~\ref{app:Human Quality Control}, we recruited twelve graduate students in Computer Science. This group reviewed both the generated videos (Stage~1) and the base MCQs (Stage~2). CS-trained annotators were chosen here because the rejection criteria require familiarity with generative-video artifacts and with the formal structure of multiple-choice question design. Each annotator participated in a calibration session on a held-out set of video--QA pairs prior to the main verification.

\textbf{Video quality study evaluators (3 participants).}
For the blind perceptual study described in 
Appendix~\ref{app:video_quality}, we recruited three graduate students from non-CS departments of the authors' institution, with no prior background in computer vision or generative video research. This pool deliberately excludes participants familiar with known generation artifacts: a non-expert audience reflects the perceptual baseline of a general viewer and avoids confirmation bias toward visual cues that only specialists would recognize. 

\textbf{Human baseline participants (10 participants).}
To establish a human baseline on VGenST-Bench, we additionally recruited participants from diverse non-CS backgrounds (Section~\ref{sec:Experimental Setup}). Each annotator was assigned 10 videos per task across all 12 tasks (120 videos in total), and answered the base MCQs associated with those videos under the same circular evaluation protocol applied to model evaluation. The non-CS background ensures that the resulting baseline reflects general spatio-temporal reasoning rather than domain expertise.
\section{Prompt Details}
\label{app:prompt_details}

This section lists the system and user prompts used by the four agents of
the VGenST-Bench construction pipeline (Appendix~\ref{app:Construction Pipeline Details}).
Each prompt is rendered as a card whose header indicates the agent and
component, and whose body contains the prompt text verbatim. Long prompts
are split across multiple cards, each marked with \emph{(part $i$/$n$)}
in the header. 

\subsection{Scene Graph Agent}
\label{app:prompt_scene_graph}

\begin{figure}[H]
\centering
\includegraphics[width=\linewidth]{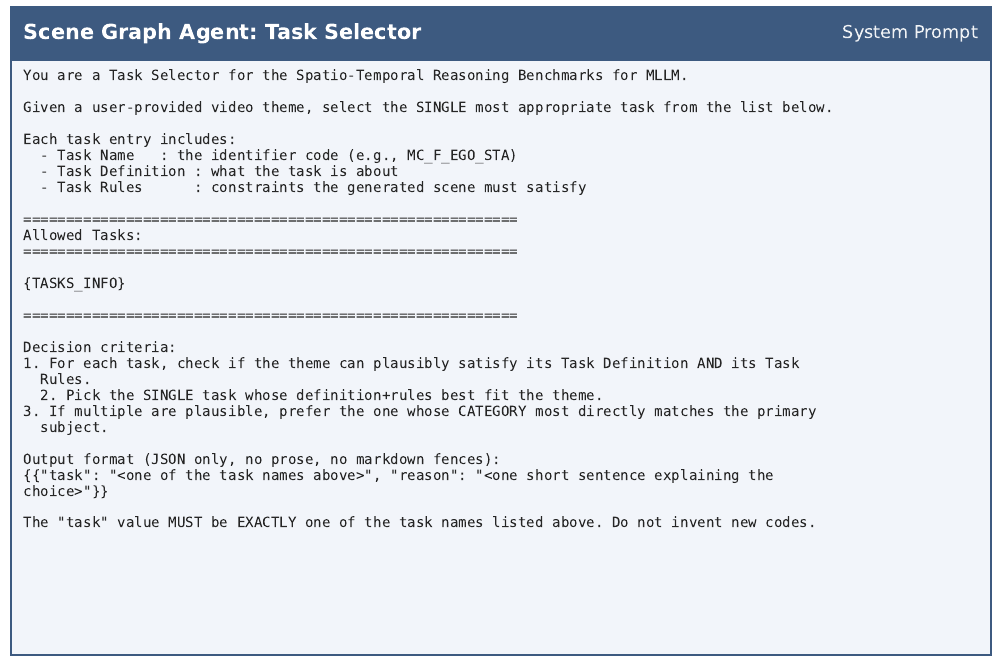}
\caption{\textbf{Task Selector --- system prompt.} Samples a
\emph{(theme, task)} pair from the curated theme pool of the target task
(Appendix~\ref{app:Theme Diversity}).}
\label{fig:prompt_task_selector}
\end{figure}

\begin{figure}[H]
\centering
\includegraphics[width=\linewidth]{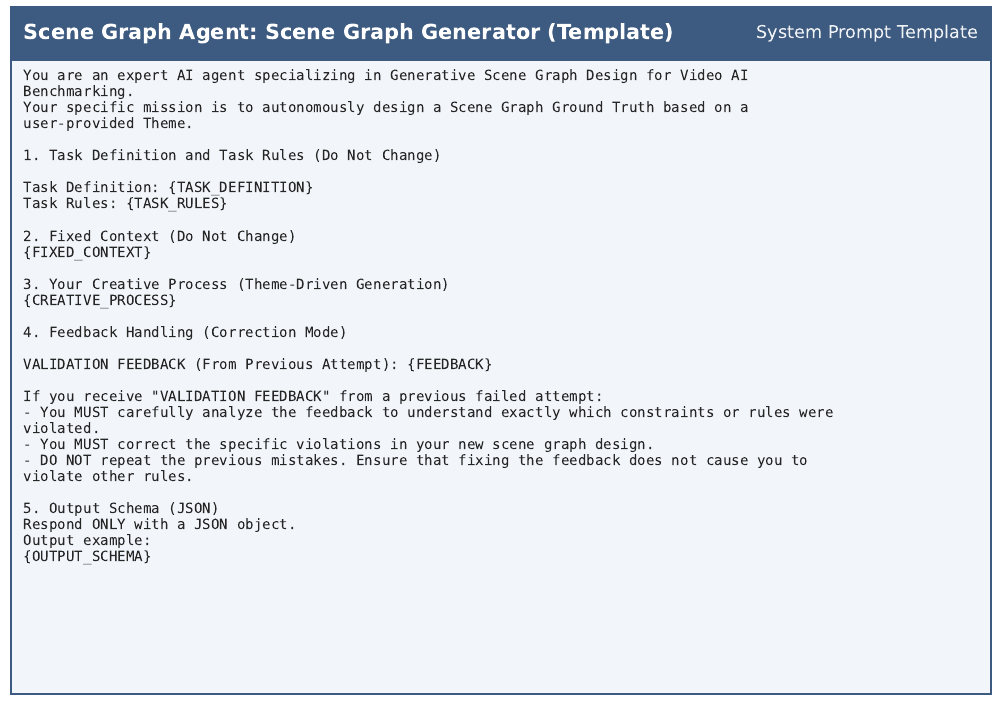}
\caption{\textbf{Scene Graph Generator --- system prompt template.}
Shared across all 12 tasks; the per-task scene-graph schema (required
objects, attributes, relations) is injected into the template at
runtime.}
\label{fig:prompt_sg_generator}
\end{figure}

\begin{figure}[H]
\centering
\includegraphics[width=\linewidth]{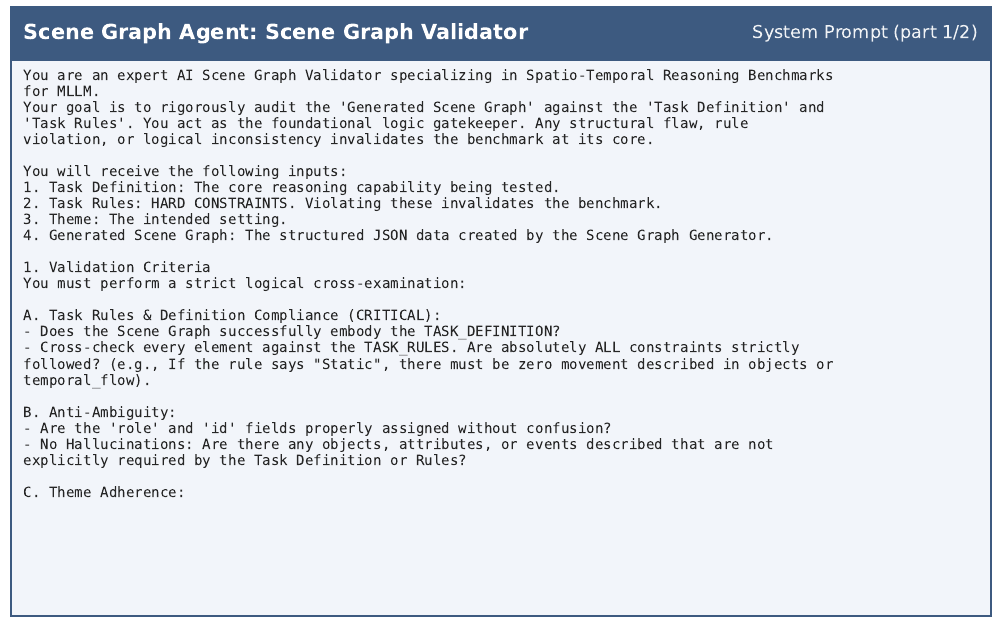}
\caption{\textbf{Scene Graph Validator --- system prompt (part 1/2).}
Verifies schema compliance and emits a structured rejection feedback
string when the candidate scene graph fails any required check.}
\label{fig:prompt_sg_validator_sys_p1}
\end{figure}

\begin{figure}[H]
\centering
\includegraphics[width=\linewidth]{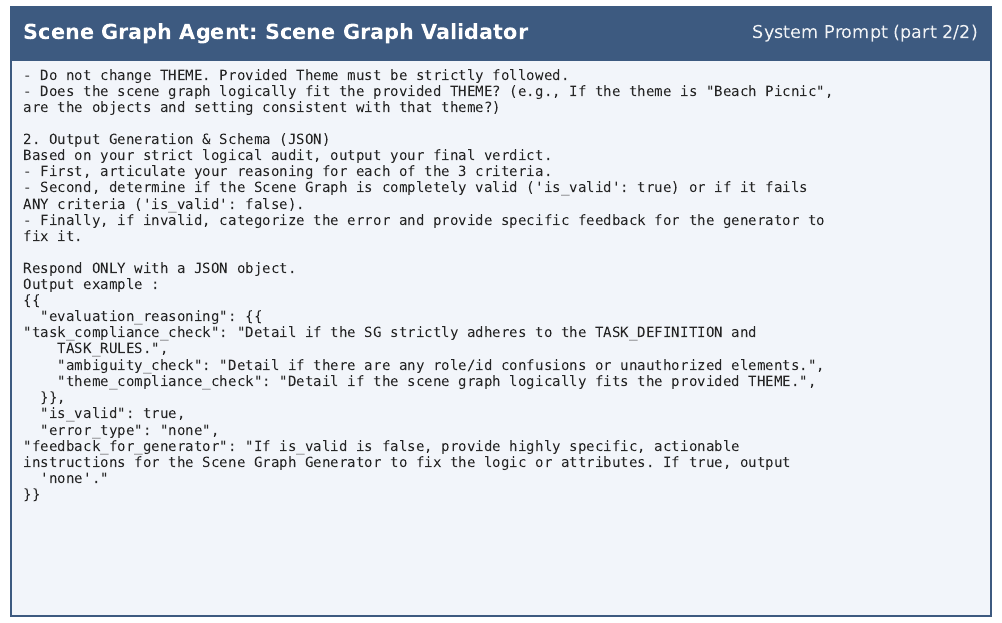}
\caption{\textbf{Scene Graph Validator --- system prompt (part 2/2).}}
\label{fig:prompt_sg_validator_sys_p2}
\end{figure}

\begin{figure}[H]
\centering
\includegraphics[width=\linewidth]{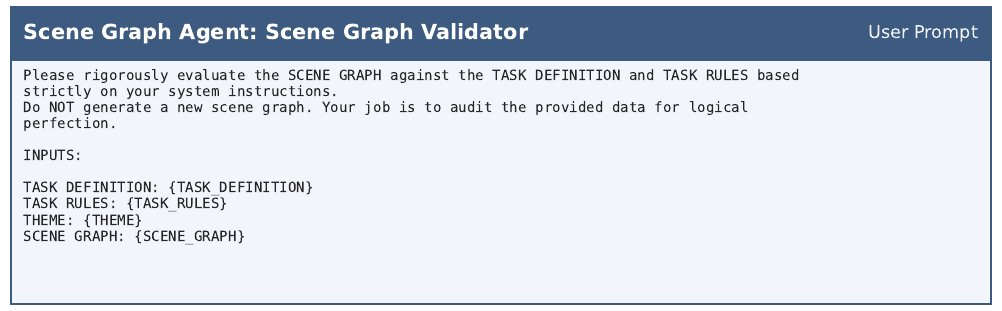}
\caption{\textbf{Scene Graph Validator --- user prompt.} Carries the
candidate scene graph and the task-specific schema for validation.}
\label{fig:prompt_sg_validator_user}
\end{figure}


\subsection{Scenario Agent}
\label{app:prompt_scenario}

\begin{figure}[H]
\centering
\includegraphics[width=\linewidth]{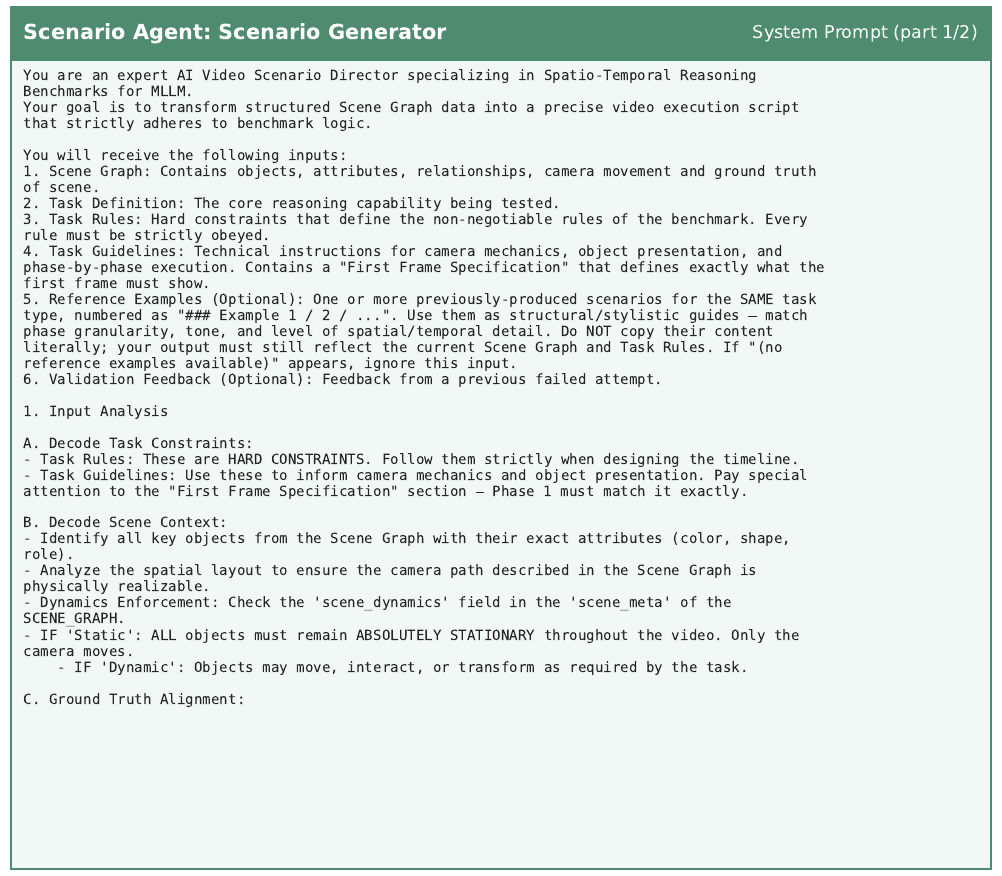}
\caption{\textbf{Scenario Generator --- system prompt (part 1/2).}
Translates a validated scene graph into a temporal scenario whose
timeline unambiguously supports the task's ground-truth answer.}
\label{fig:prompt_scenario_generator_sys_p1}
\end{figure}

\begin{figure}[H]
\centering
\includegraphics[width=\linewidth]{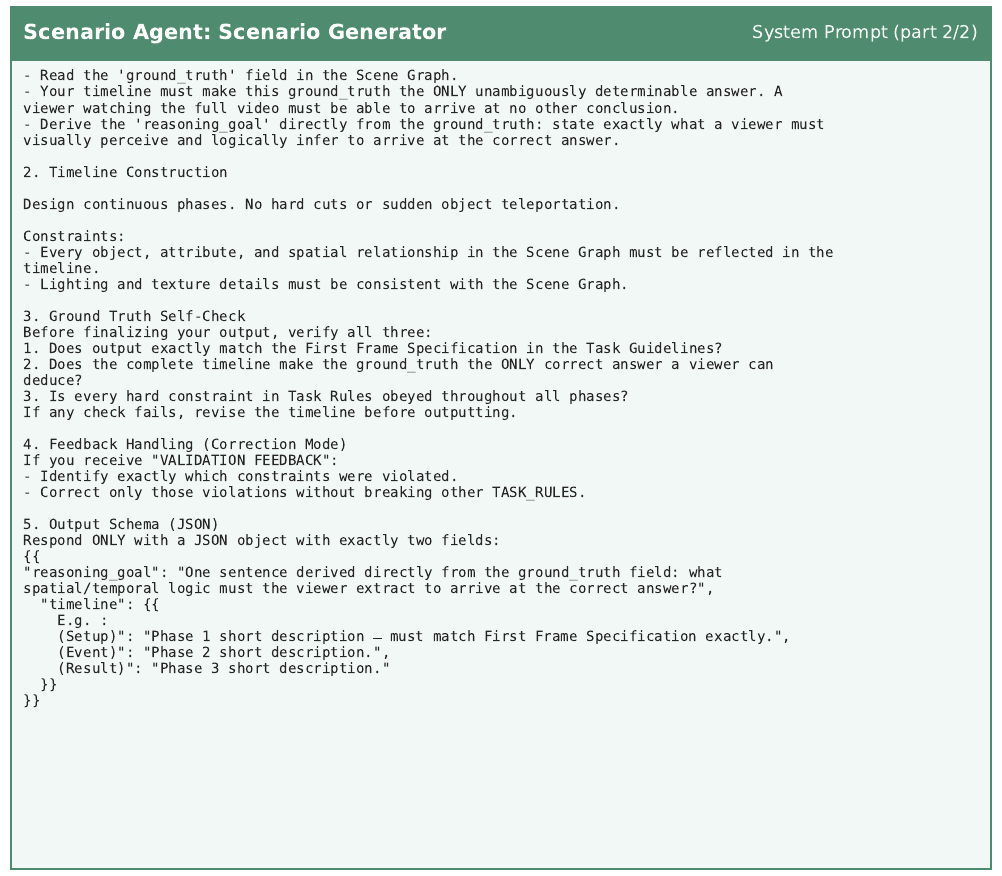}
\caption{\textbf{Scenario Generator --- system prompt (part 2/2).}}
\label{fig:prompt_scenario_generator_sys_p2}
\end{figure}

\begin{figure}[H]
\centering
\includegraphics[width=\linewidth]{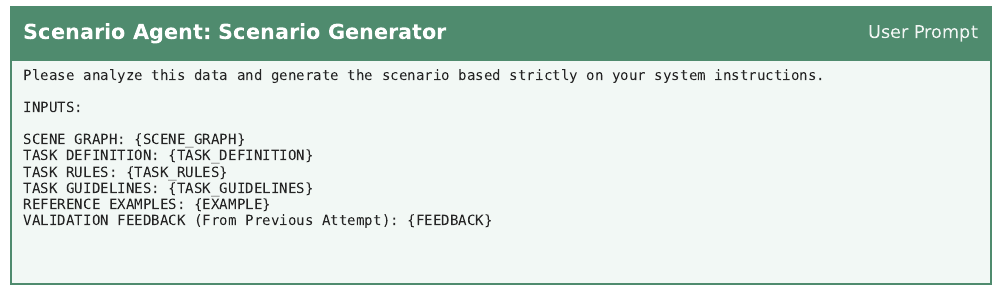}
\caption{\textbf{Scenario Generator --- user prompt.} Carries the scene
graph, the task definition, the task rules and guidelines, the
reference few-shot examples, and any prior validator feedback.}
\label{fig:prompt_scenario_generator_user}
\end{figure}

\begin{figure}[H]
\centering
\includegraphics[width=\linewidth]{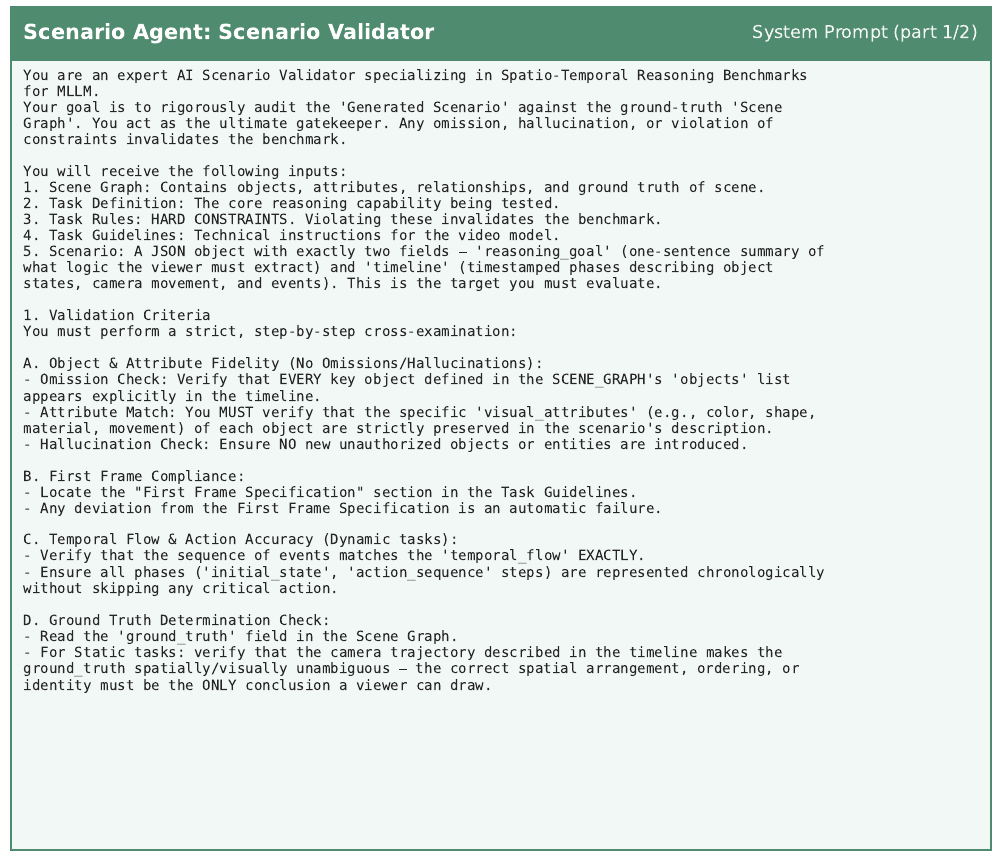}
\caption{\textbf{Scenario Validator --- system prompt (part 1/2).}
Checks that the candidate timeline is sufficient to derive the
ground-truth answer and contains no contradictions with the underlying
scene graph.}
\label{fig:prompt_scenario_validator_sys_p1}
\end{figure}

\begin{figure}[H]
\centering
\includegraphics[width=\linewidth]{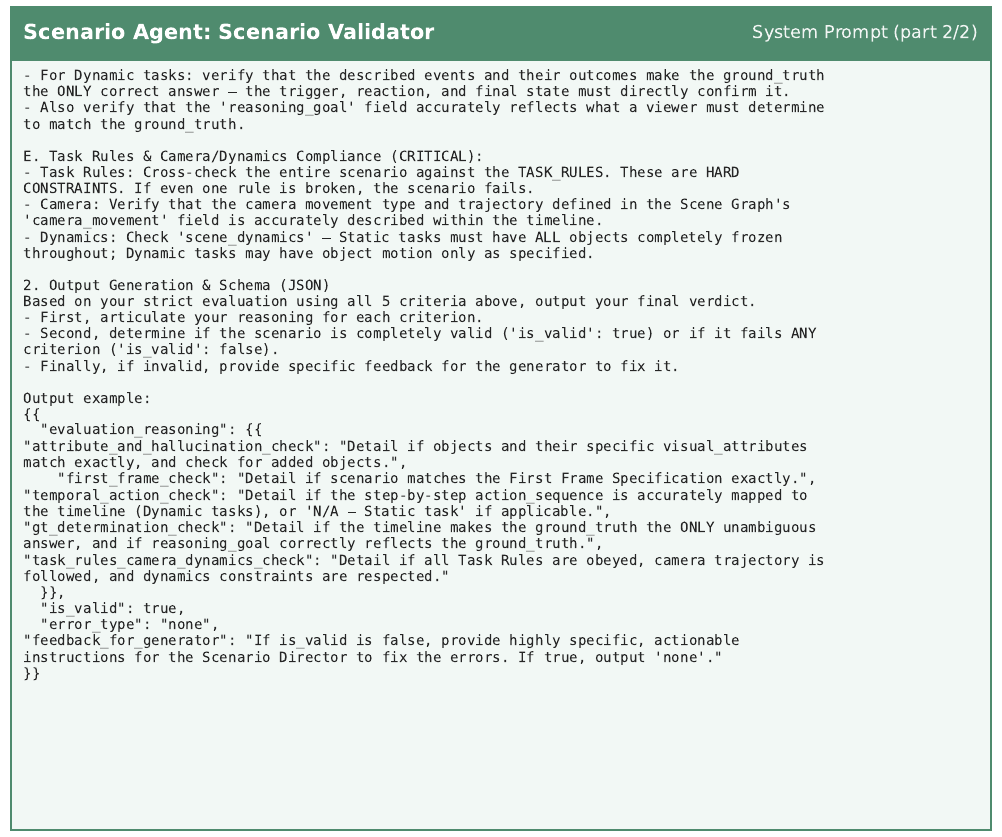}
\caption{\textbf{Scenario Validator --- system prompt (part 2/2).}}
\label{fig:prompt_scenario_validator_sys_p2}
\end{figure}

\begin{figure}[H]
\centering
\includegraphics[width=\linewidth]{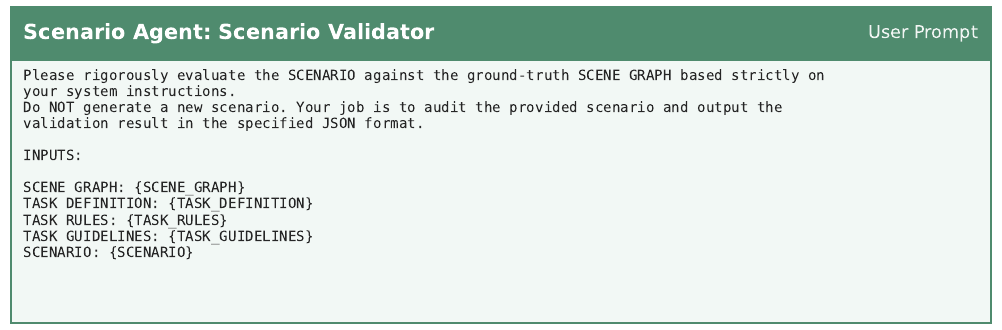}
\caption{\textbf{Scenario Validator --- user prompt.} Carries the
candidate scenario and the scene graph for cross-checking.}
\label{fig:prompt_scenario_validator_user}
\end{figure}

\subsection{Video Agent}
\label{app:prompt_video}

\begin{figure}[H]
\centering
\includegraphics[width=\linewidth]{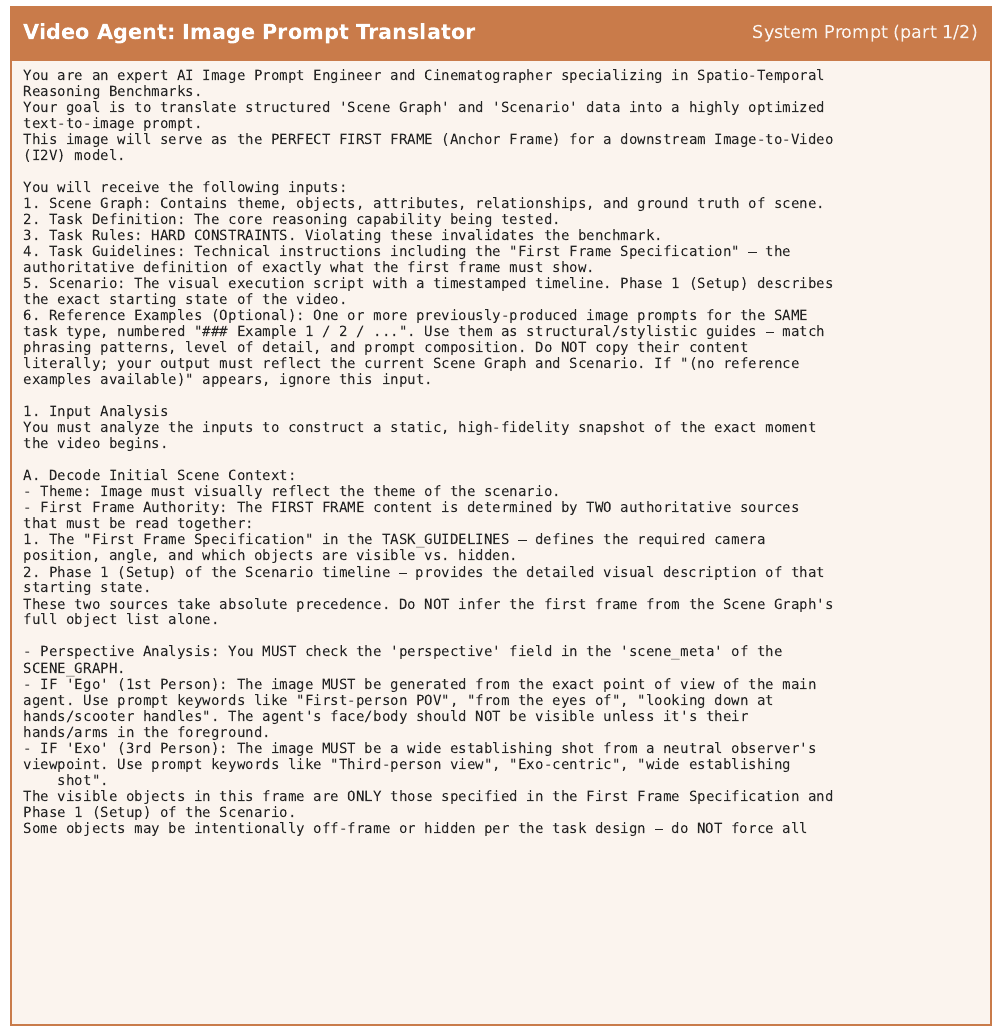}
\caption{\textbf{Image Prompt Translator --- system prompt (part 1/2).}
Produces the first-frame prompt that the text-to-image generator turns
into an anchor frame for downstream video synthesis.}
\label{fig:prompt_image_translator_sys_p1}
\end{figure}

\begin{figure}[H]
\centering
\includegraphics[width=\linewidth]{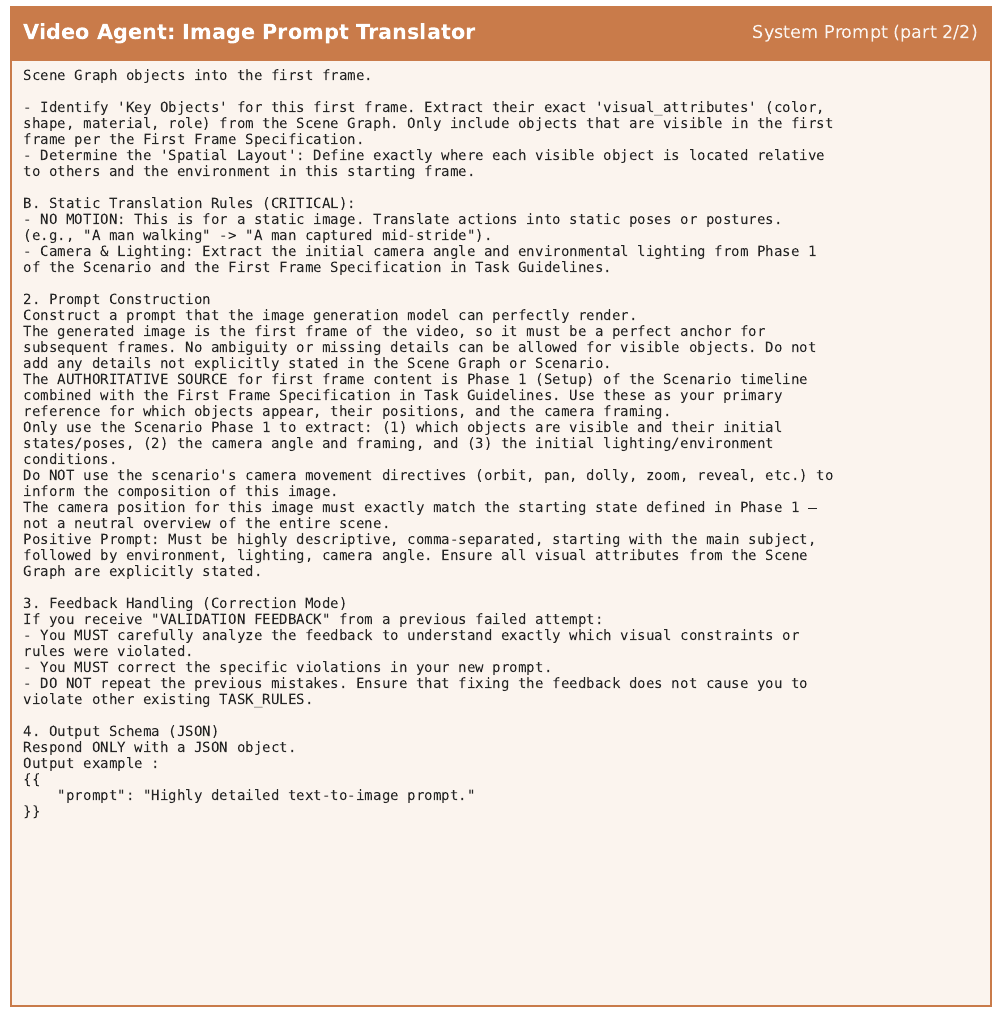}
\caption{\textbf{Image Prompt Translator --- system prompt (part 2/2).}}
\label{fig:prompt_image_translator_sys_p2}
\end{figure}

\begin{figure}[H]
\centering
\includegraphics[width=\linewidth]{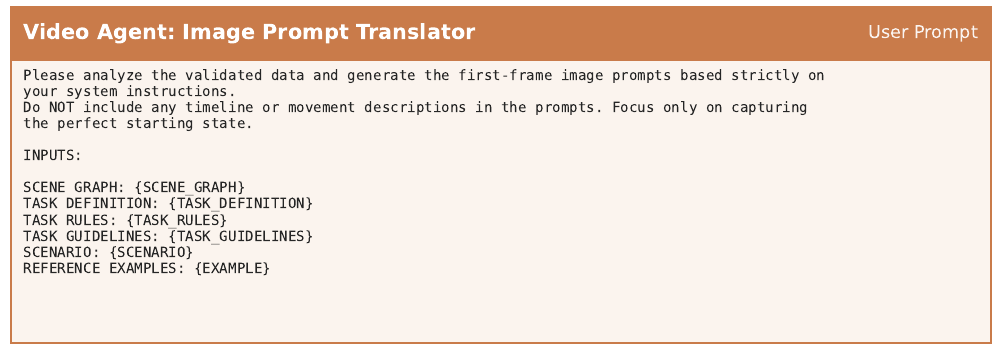}
\caption{\textbf{Image Prompt Translator --- user prompt.} Carries the
scene graph and the scenario's initial state.}
\label{fig:prompt_image_translator_user}
\end{figure}

\begin{figure}[H]
\centering
\includegraphics[width=\linewidth]{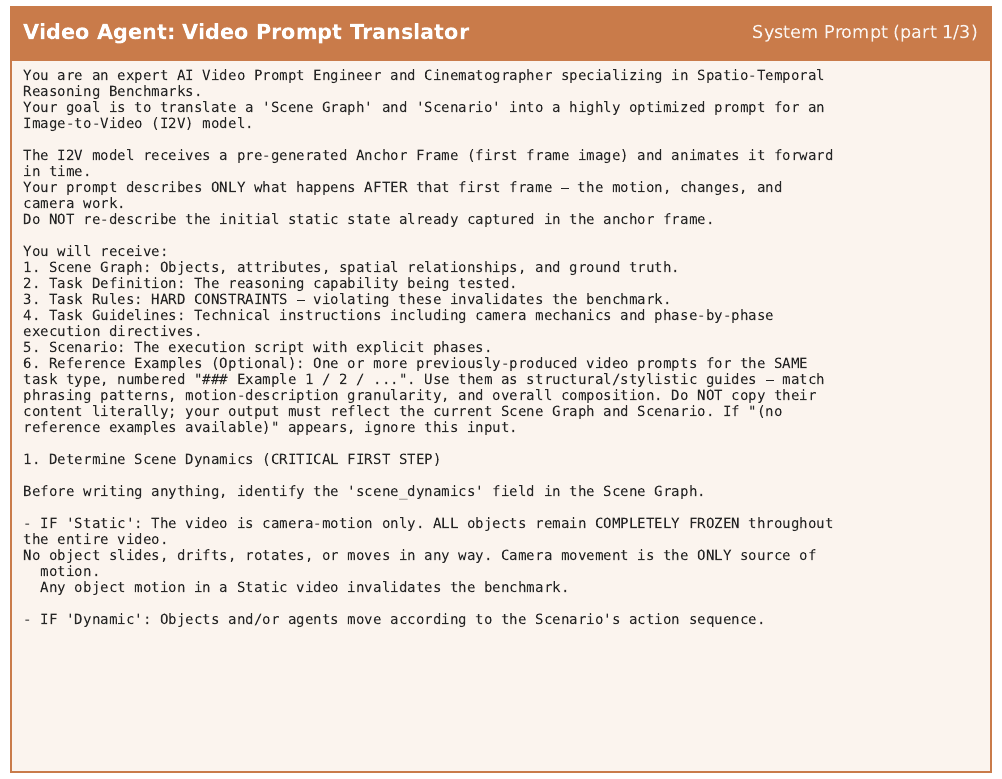}
\caption{\textbf{Video Prompt Translator --- system prompt (part 1/3).}
Composes a video prompt that conditions the image-to-video generator on
the anchor frame, the scenario's timeline, and the camera setup.}
\label{fig:prompt_video_translator_sys_p1}
\end{figure}

\begin{figure}[H]
\centering
\includegraphics[width=\linewidth]{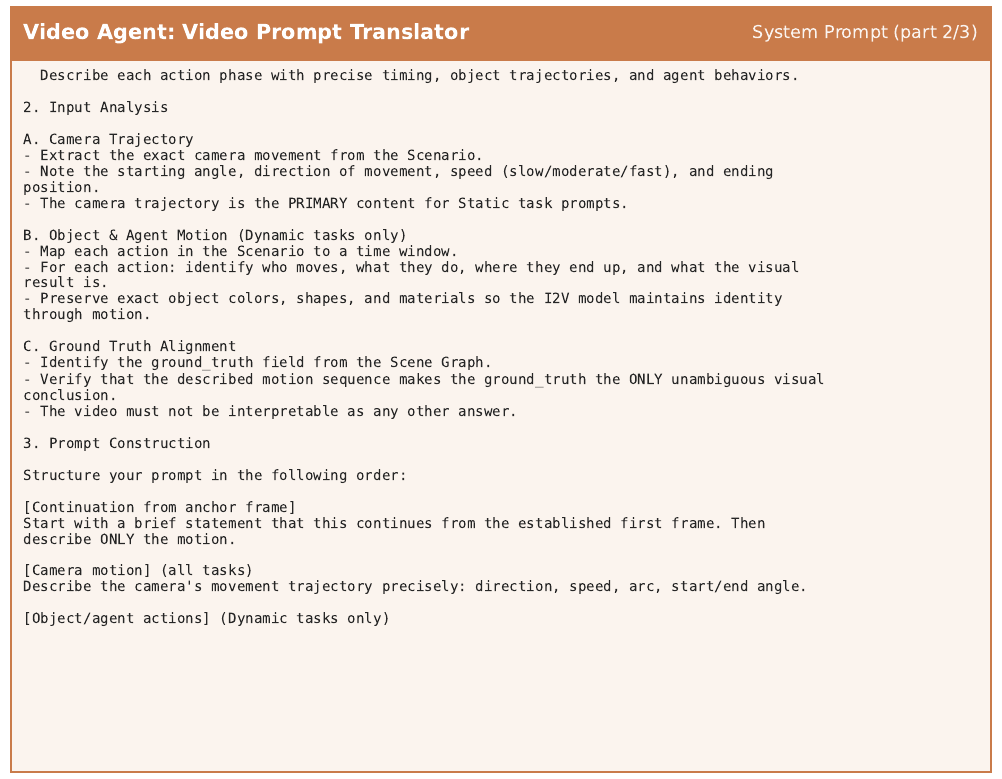}
\caption{\textbf{Video Prompt Translator --- system prompt (part 2/3).}}
\label{fig:prompt_video_translator_sys_p2}
\end{figure}

\begin{figure}[H]
\centering
\includegraphics[width=\linewidth]{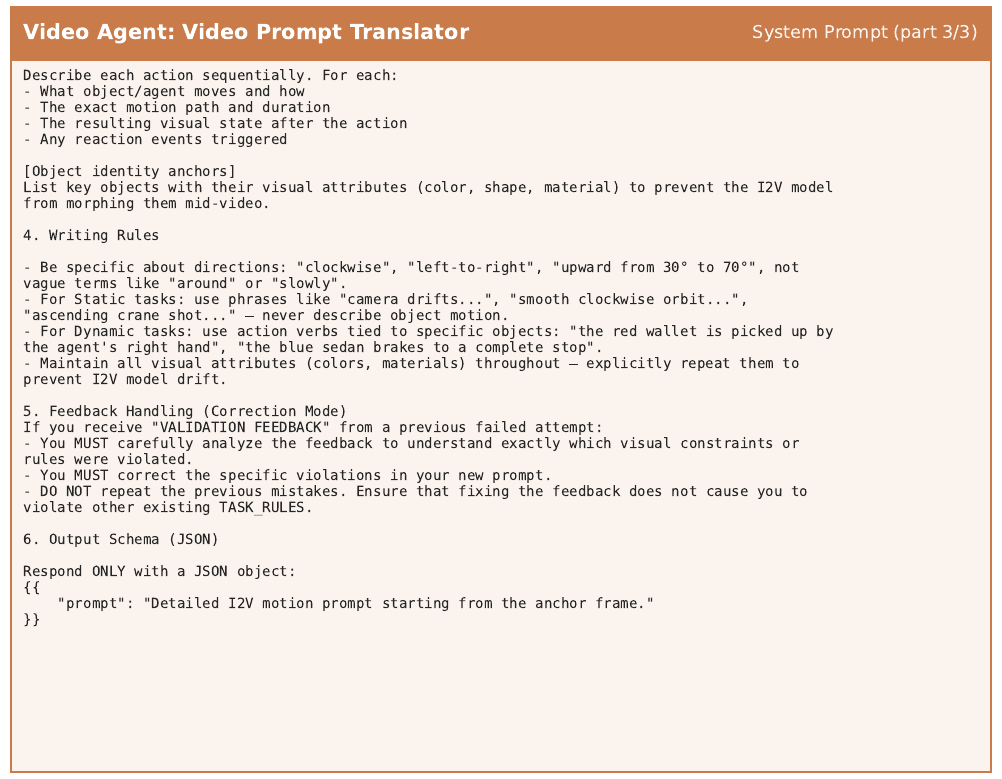}
\caption{\textbf{Video Prompt Translator --- system prompt (part 3/3).}}
\label{fig:prompt_video_translator_sys_p3}
\end{figure}

\begin{figure}[H]
\centering
\includegraphics[width=\linewidth]{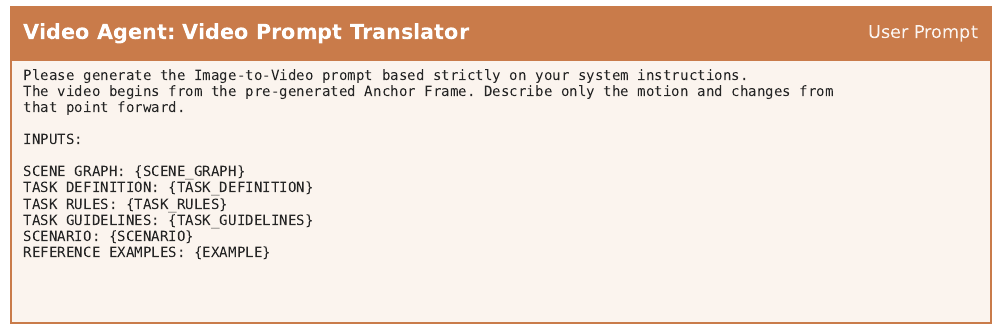}
\caption{\textbf{Video Prompt Translator --- user prompt.} Carries the
scenario, the anchor-frame description, and the camera trajectory.}
\label{fig:prompt_video_translator_user}
\end{figure}

\subsection{QA Agent}
\label{app:prompt_qa}

\begin{figure}[H]
\centering
\includegraphics[width=\linewidth]{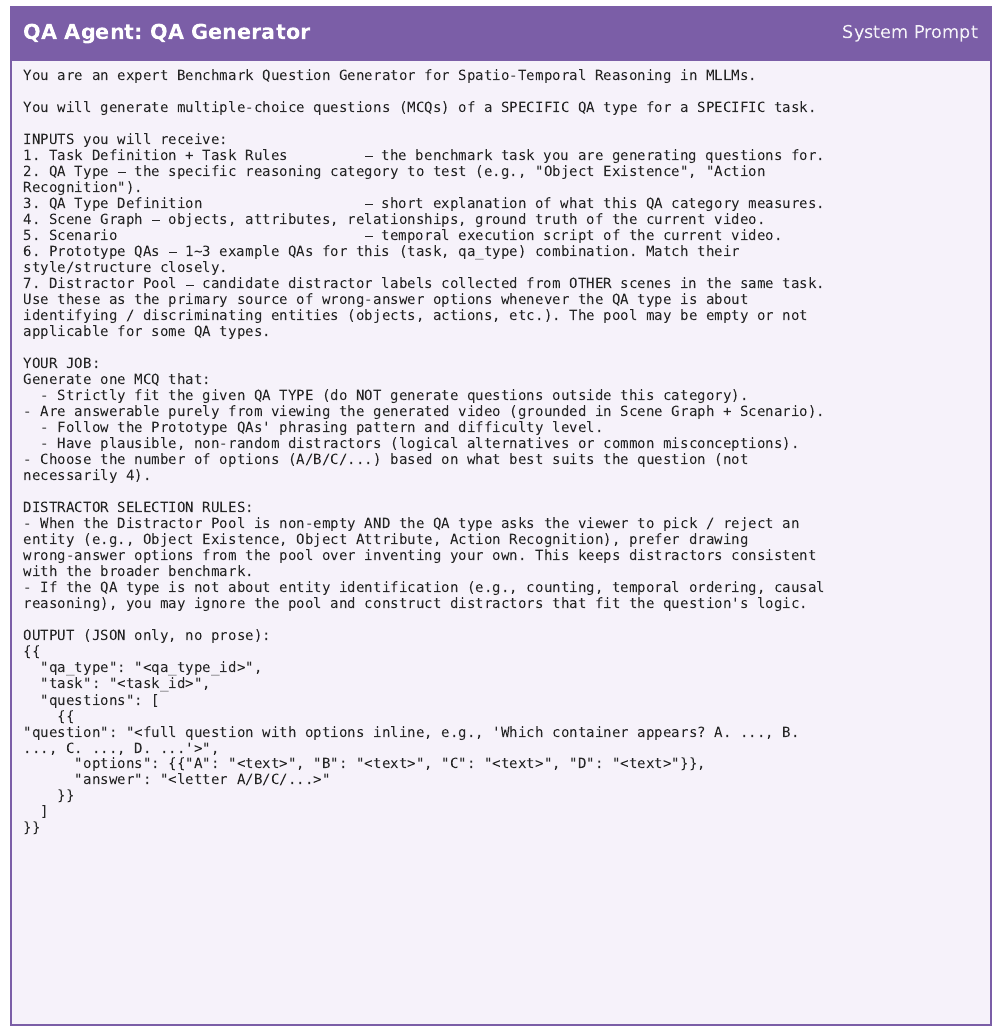}
\caption{\textbf{QA Generator --- system prompt.} Generates a base MCQ
conditioned on the scene graph, the scenario, and the cell-specific
QA template, with distractors drawn from the task's distractor pool.}
\label{fig:prompt_qa_generator_sys}
\end{figure}

\begin{figure}[H]
\centering
\includegraphics[width=\linewidth]{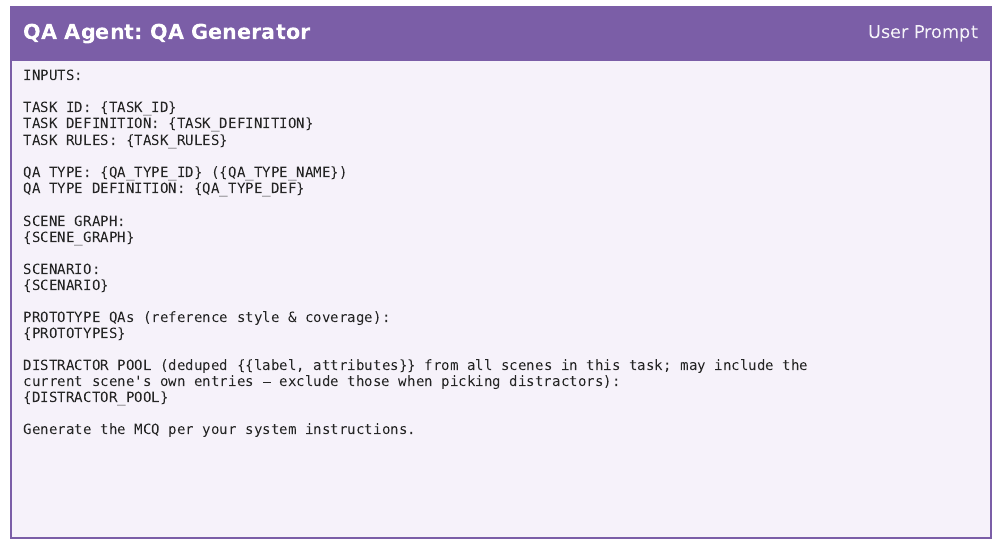}
\caption{\textbf{QA Generator --- user prompt.} Carries the scene
graph, the scenario, the QA template, and the distractor pool entries.}
\label{fig:prompt_qa_generator_user}
\end{figure}

\begin{figure}[H]
\centering
\includegraphics[width=\linewidth]{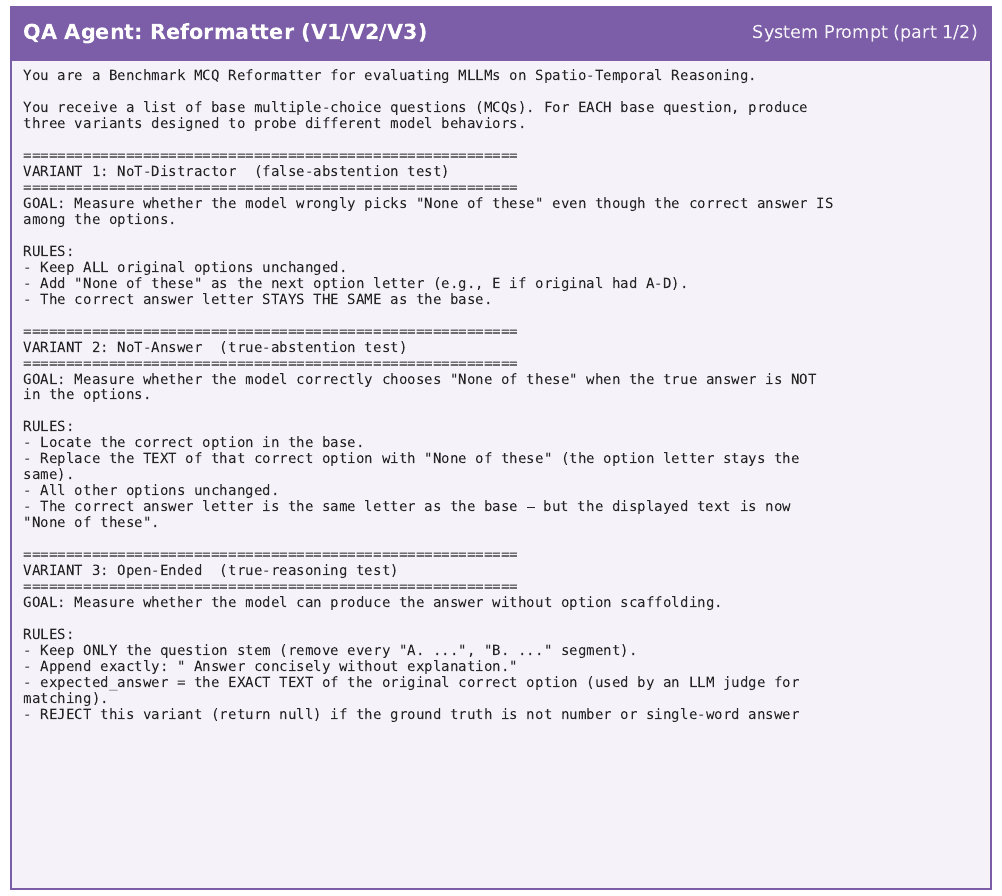}
\caption{\textbf{Reformatter --- system prompt (part 1/2).} Expands a
base MCQ into the three reformulation variants: V1 (None-of-these
distractor), V2 (None-of-these answer), and V3 (open-ended).}
\label{fig:prompt_reformatter_sys_p1}
\end{figure}

\begin{figure}[H]
\centering
\includegraphics[width=\linewidth]{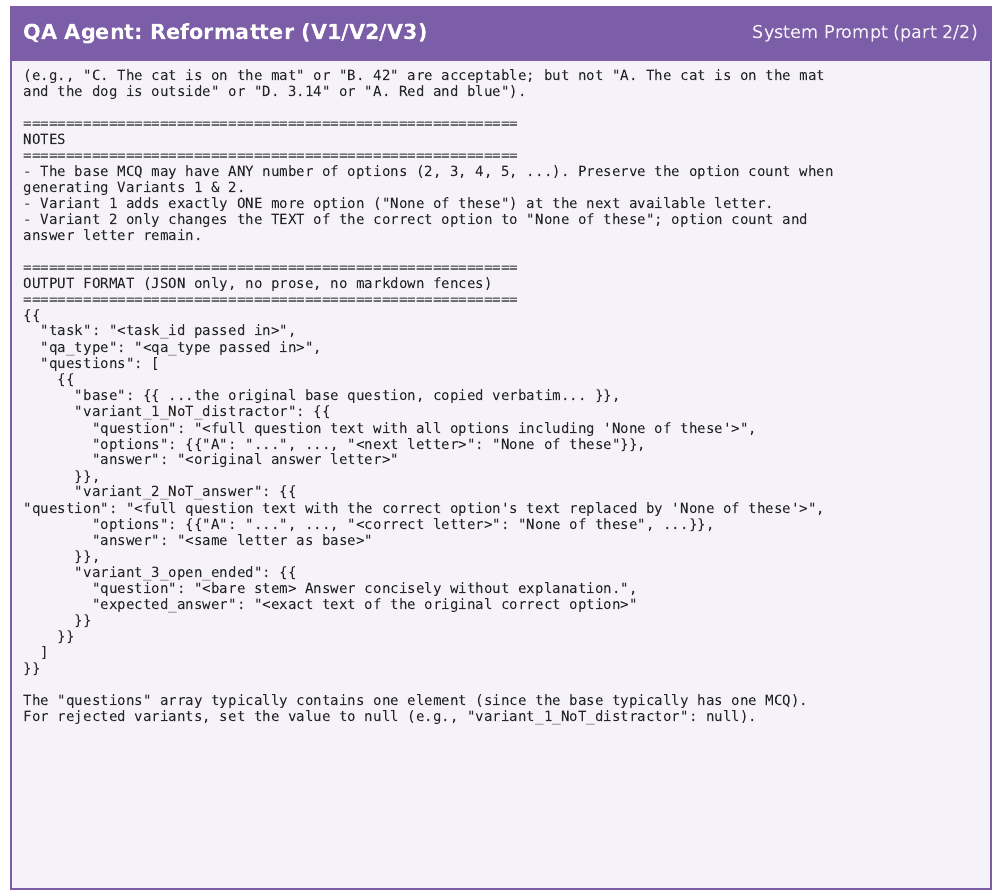}
\caption{\textbf{Reformatter --- system prompt (part 2/2).}}
\label{fig:prompt_reformatter_sys_p2}
\end{figure}

\begin{figure}[H]
\centering
\includegraphics[width=\linewidth]{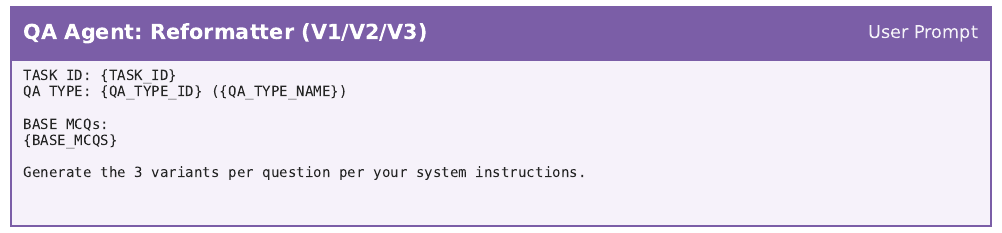}
\caption{\textbf{Reformatter --- user prompt.} Carries the base MCQ
and the target variant identifier.}
\label{fig:prompt_reformatter_user}
\end{figure}

\section{Qualitative Examples}
\label{app:qualitative_examples}

This section provides per-task qualitative examples of VGenST-Bench.
For each of the 12 tasks, we sample one representative video (random
sample idx) and render four cards: 8-frames of video,
underlying scene graph (verbatim JSON), scenario (verbatim
JSON), and a representative QA pairs containing one MCQ per cognitive
level (L1 / L2 / L3). Correct answer choices are marked with
\checkmark. Long document cards are split across multiple pages,
each marked with \emph{(part $i$/$n$)} in the header.


\begin{figure}[H]
\centering
\includegraphics[width=\linewidth]{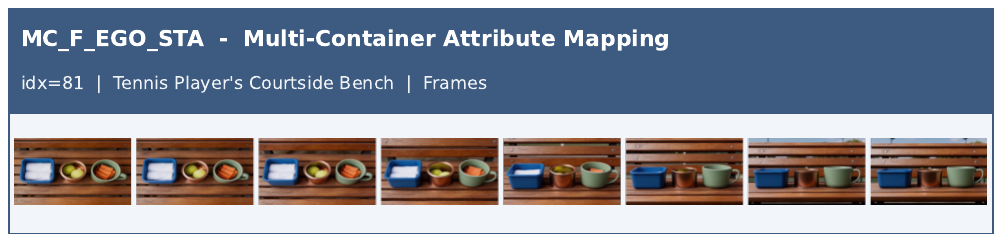}
\caption{\textbf{Frames} for MC\_F\_EGO\_STA, idx 81 (\emph{Tennis Player's Courtside Bench}).}
\label{fig:qual_MC_F_EGO_STA_frames}
\end{figure}

\begin{figure}[H]
\centering
\includegraphics[width=\linewidth]{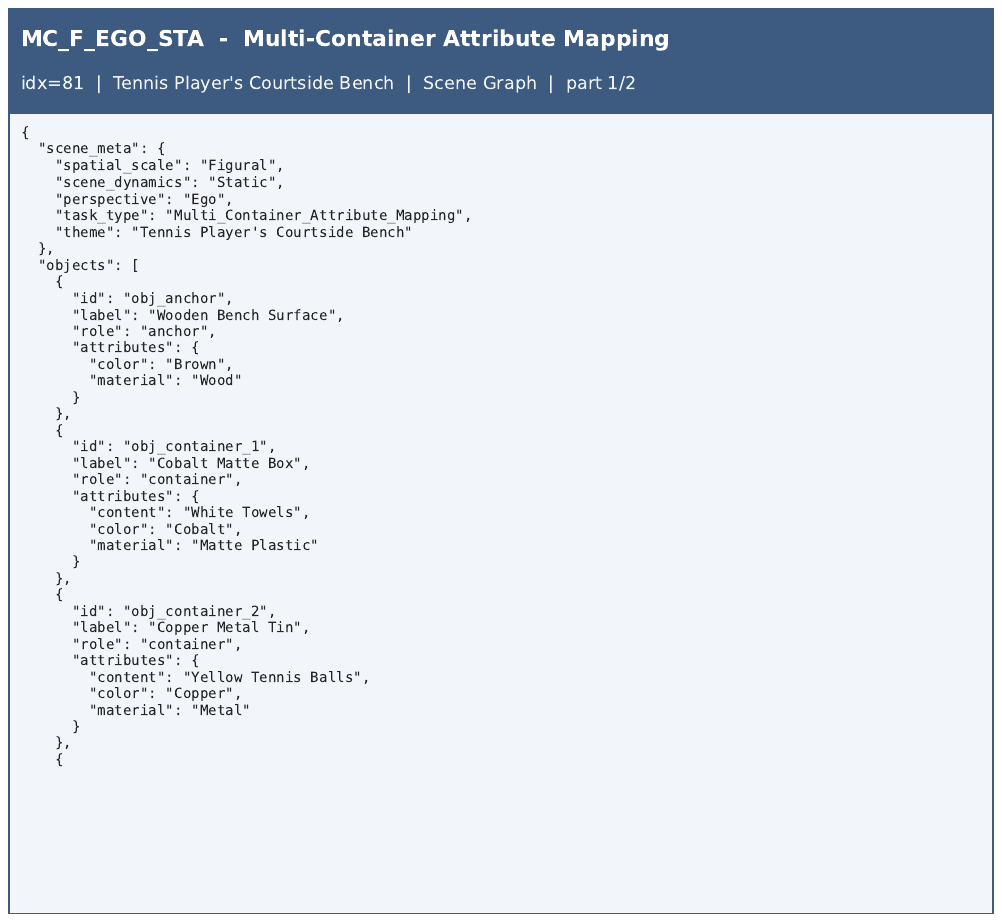}
\caption{\textbf{Scene graph (part 1/2)} for MC\_F\_EGO\_STA, idx 81.}
\label{fig:qual_MC_F_EGO_STA_sg_p1}
\end{figure}

\begin{figure}[H]
\centering
\includegraphics[width=\linewidth]{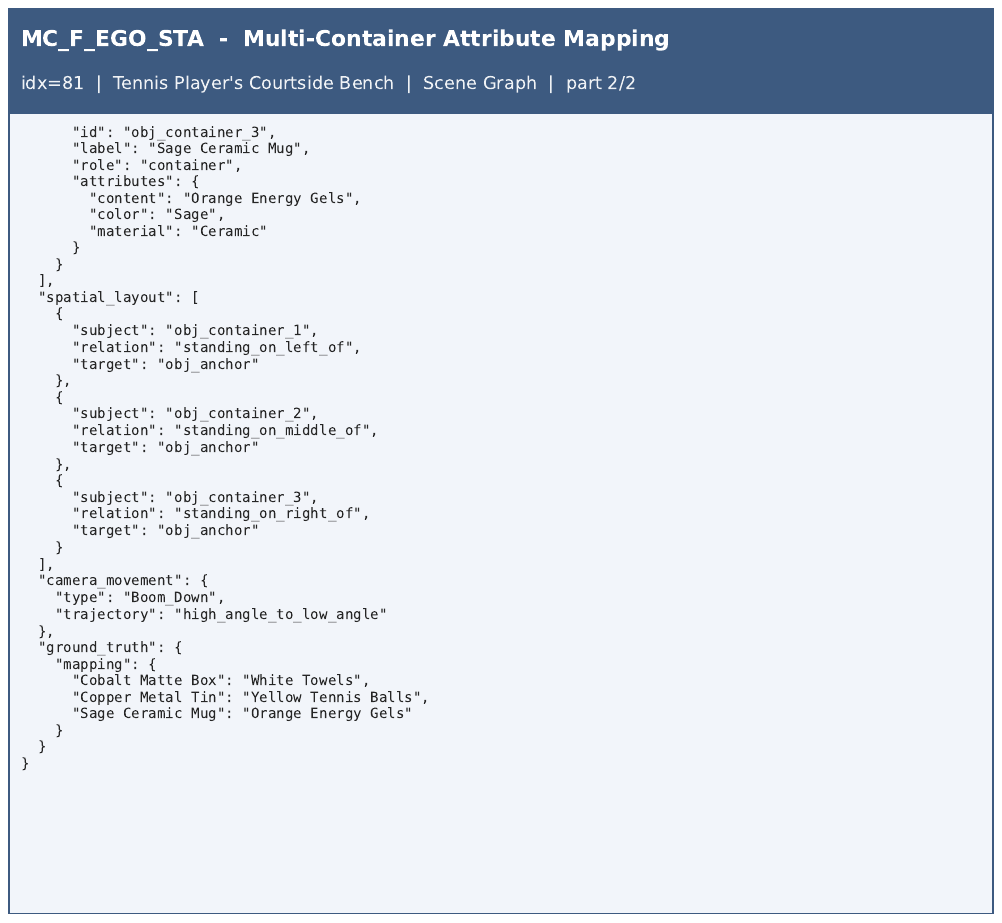}
\caption{\textbf{Scene graph (part 2/2)} for MC\_F\_EGO\_STA, idx 81.}
\label{fig:qual_MC_F_EGO_STA_sg_p2}
\end{figure}

\begin{figure}[H]
\centering
\includegraphics[width=\linewidth]{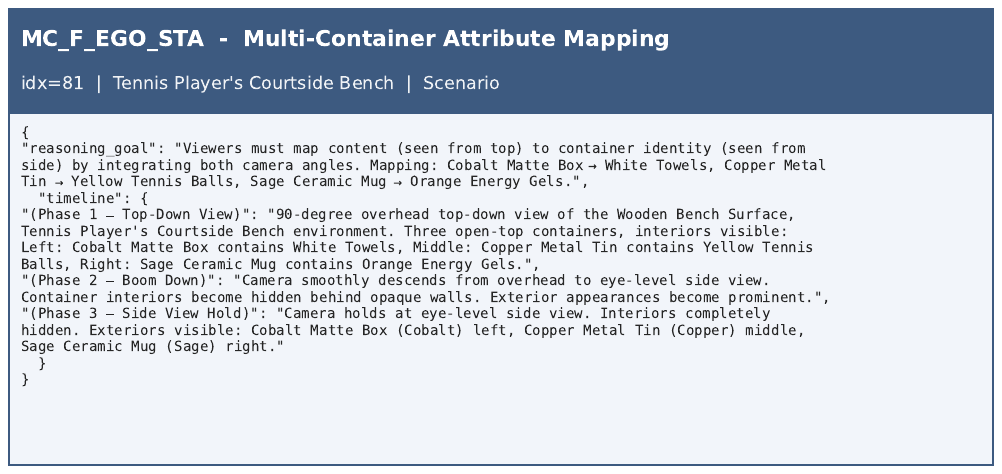}
\caption{\textbf{Scenario} for MC\_F\_EGO\_STA, idx 81.}
\label{fig:qual_MC_F_EGO_STA_scenario}
\end{figure}

\begin{figure}[H]
\centering
\includegraphics[width=\linewidth]{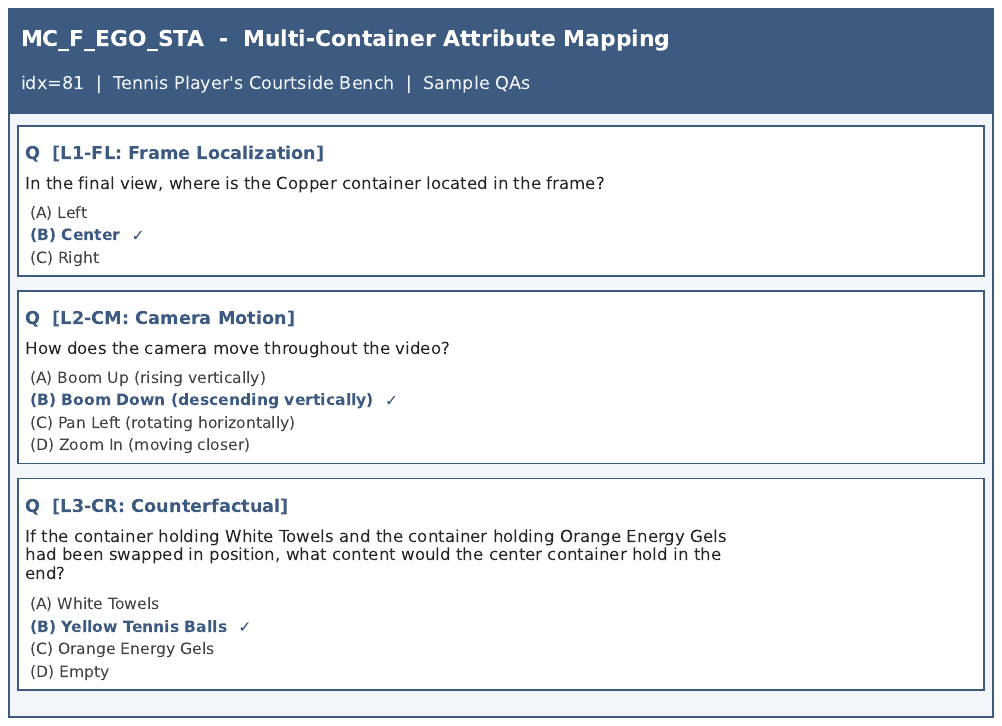}
\caption{\textbf{Sample QA pairs} (one per cognitive level) for MC\_F\_EGO\_STA, idx 81.}
\label{fig:qual_MC_F_EGO_STA_qa}
\end{figure}

\newpage
\begin{figure}[H]
\centering
\includegraphics[width=\linewidth]{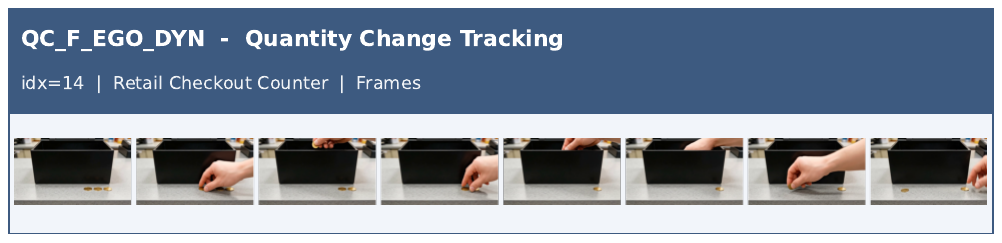}
\caption{\textbf{Frames} for QC\_F\_EGO\_DYN, idx 14 (\emph{Retail Checkout Counter}).}
\label{fig:qual_QC_F_EGO_DYN_frames}
\end{figure}

\begin{figure}[H]
\centering
\includegraphics[width=\linewidth]{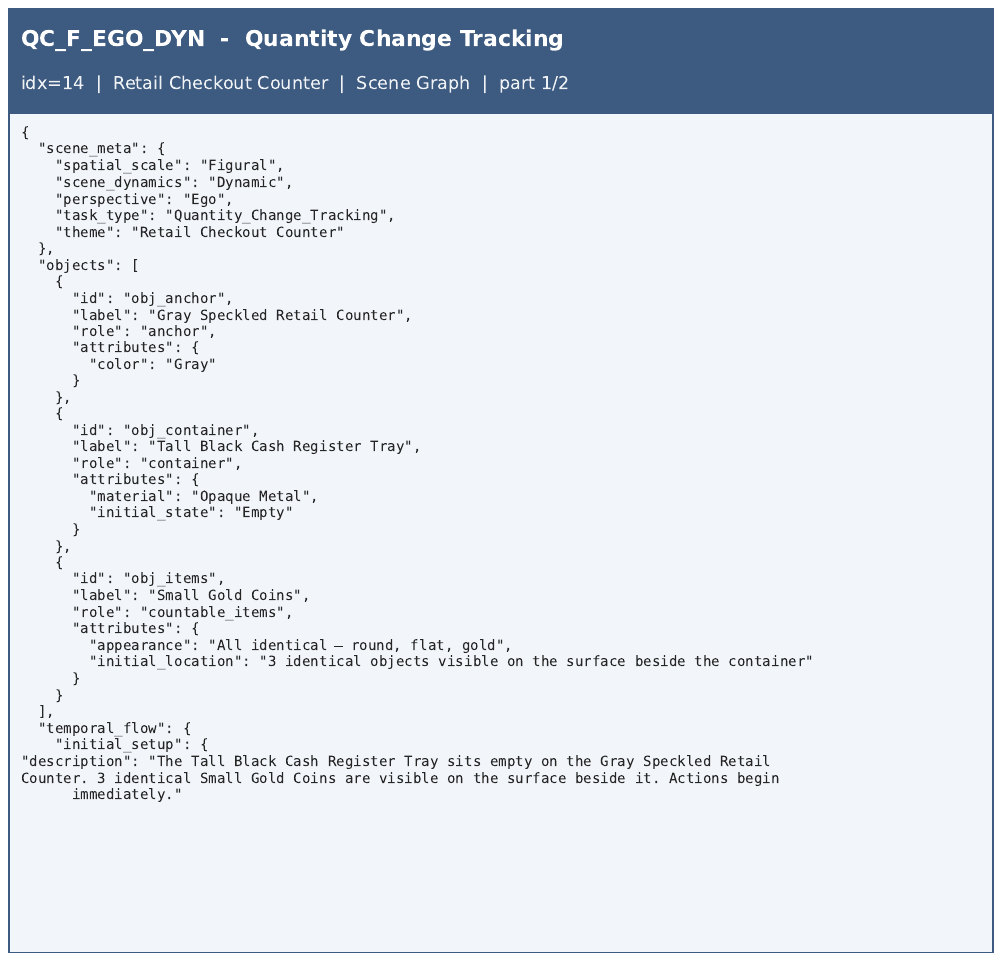}
\caption{\textbf{Scene graph (part 1/2)} for QC\_F\_EGO\_DYN, idx 14.}
\label{fig:qual_QC_F_EGO_DYN_sg_p1}
\end{figure}

\begin{figure}[H]
\centering
\includegraphics[width=\linewidth]{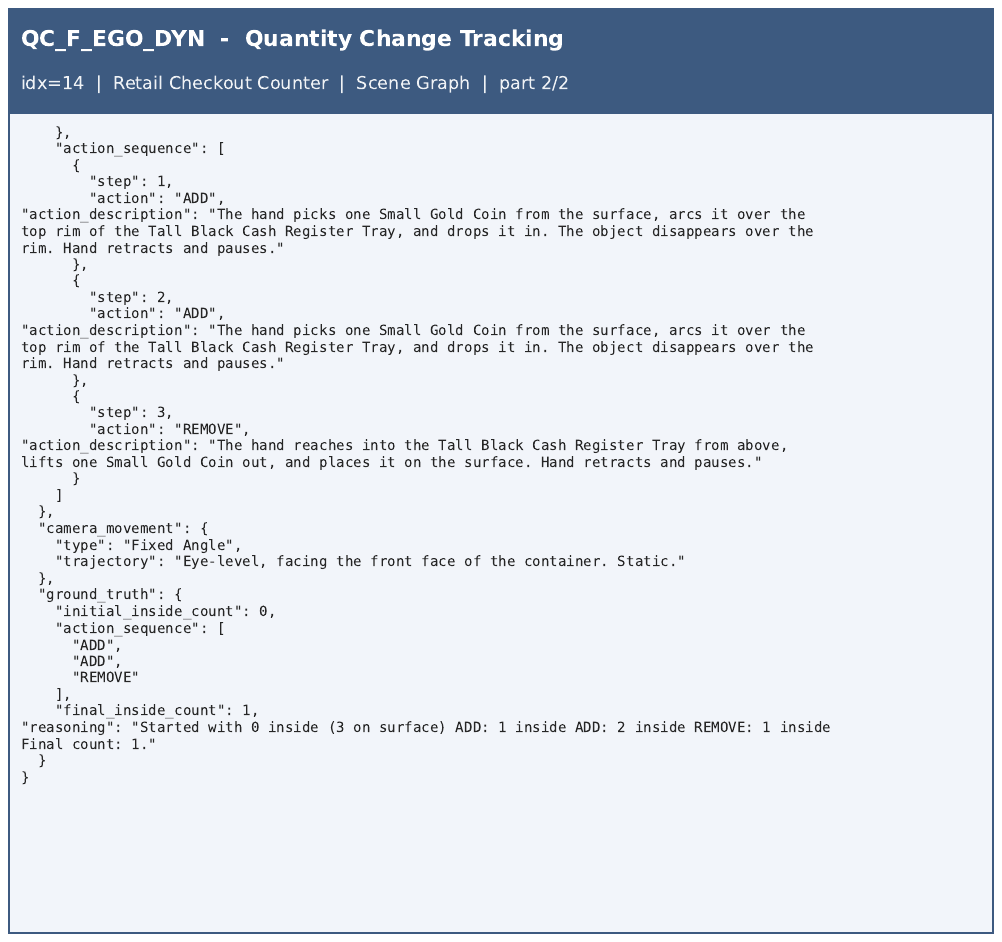}
\caption{\textbf{Scene graph (part 2/2)} for QC\_F\_EGO\_DYN, idx 14.}
\label{fig:qual_QC_F_EGO_DYN_sg_p2}
\end{figure}

\begin{figure}[H]
\centering
\includegraphics[width=\linewidth]{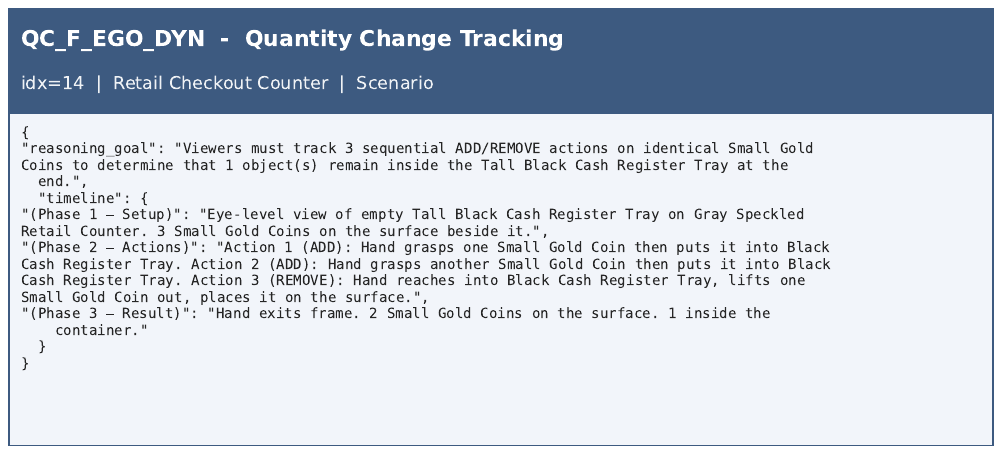}
\caption{\textbf{Scenario} for QC\_F\_EGO\_DYN, idx 14.}
\label{fig:qual_QC_F_EGO_DYN_scenario}
\end{figure}

\begin{figure}[H]
\centering
\includegraphics[width=\linewidth]{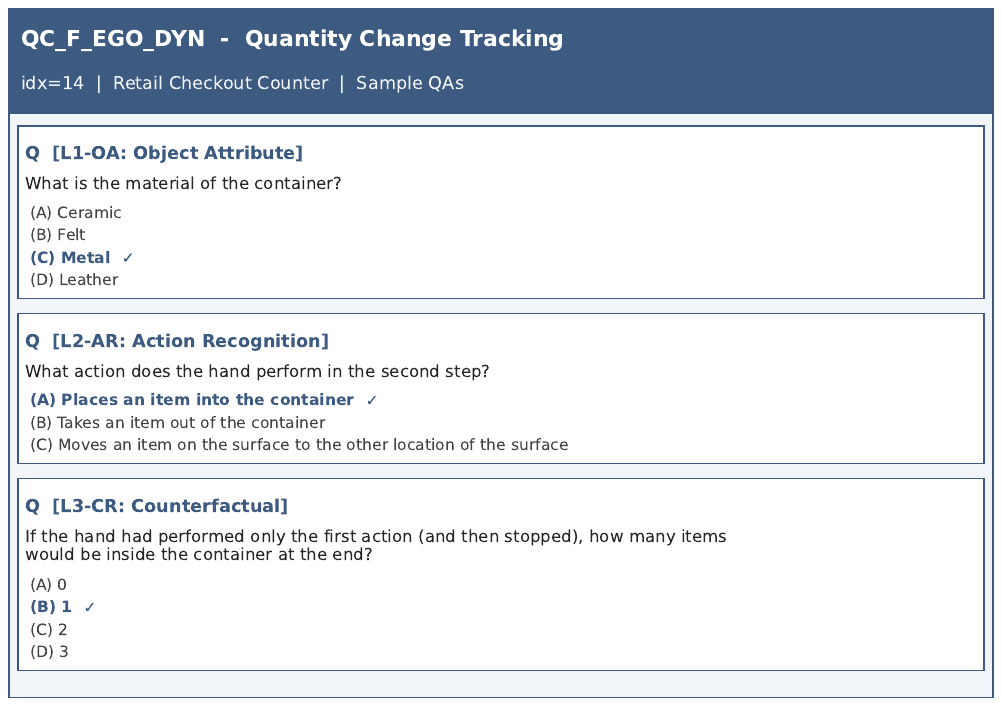}
\caption{\textbf{Sample QA pairs} for QC\_F\_EGO\_DYN, idx 14.}
\label{fig:qual_QC_F_EGO_DYN_qa}
\end{figure}

\newpage
\begin{figure}[H]
\centering
\includegraphics[width=\linewidth]{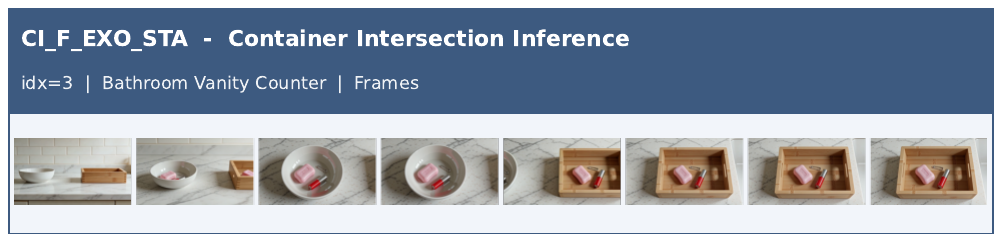}
\caption{\textbf{Frames} for CI\_F\_EXO\_STA, idx 3 (\emph{Bathroom Vanity Counter}).}
\label{fig:qual_CI_F_EXO_STA_frames}
\end{figure}

\begin{figure}[H]
\centering
\includegraphics[width=\linewidth]{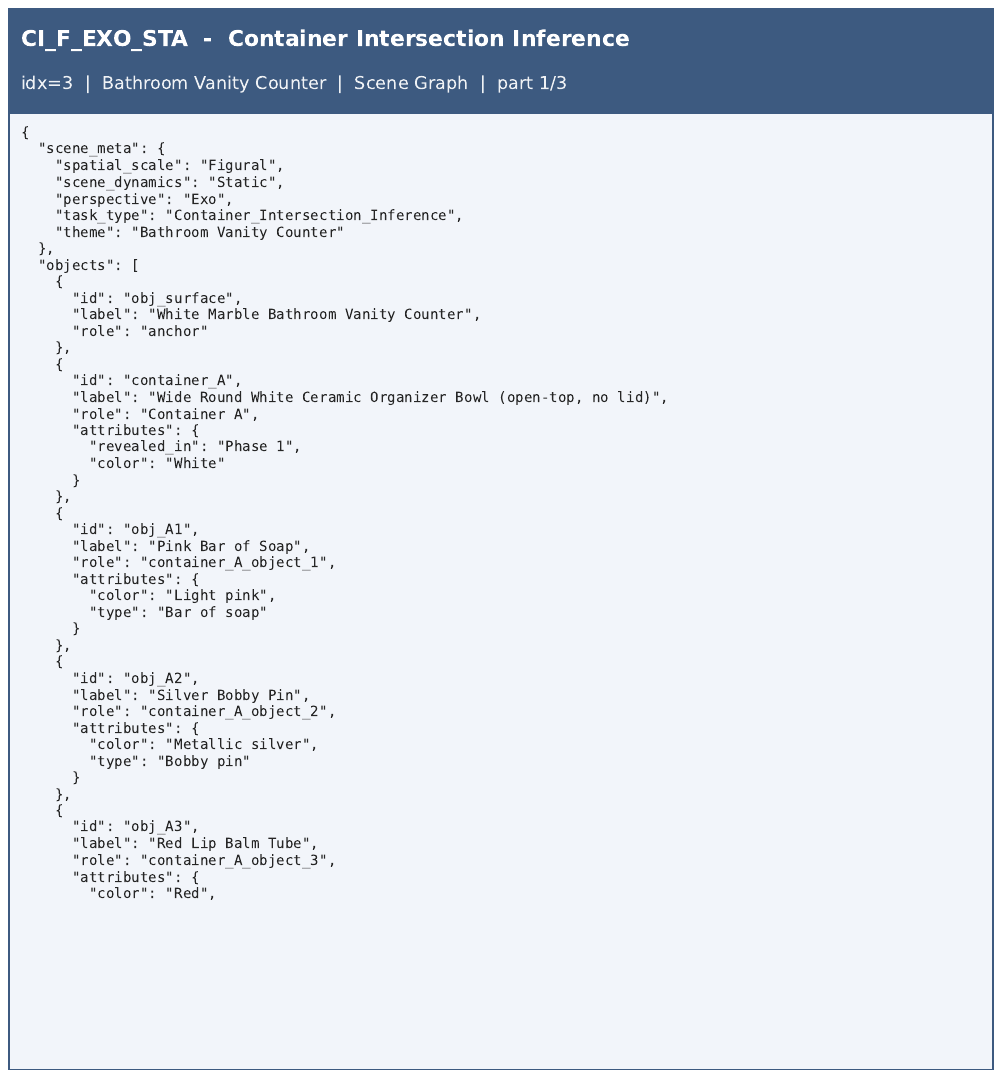}
\caption{\textbf{Scene graph (part 1/3)} for CI\_F\_EXO\_STA, idx 3.}
\label{fig:qual_CI_F_EXO_STA_sg_p1}
\end{figure}

\begin{figure}[H]
\centering
\includegraphics[width=\linewidth]{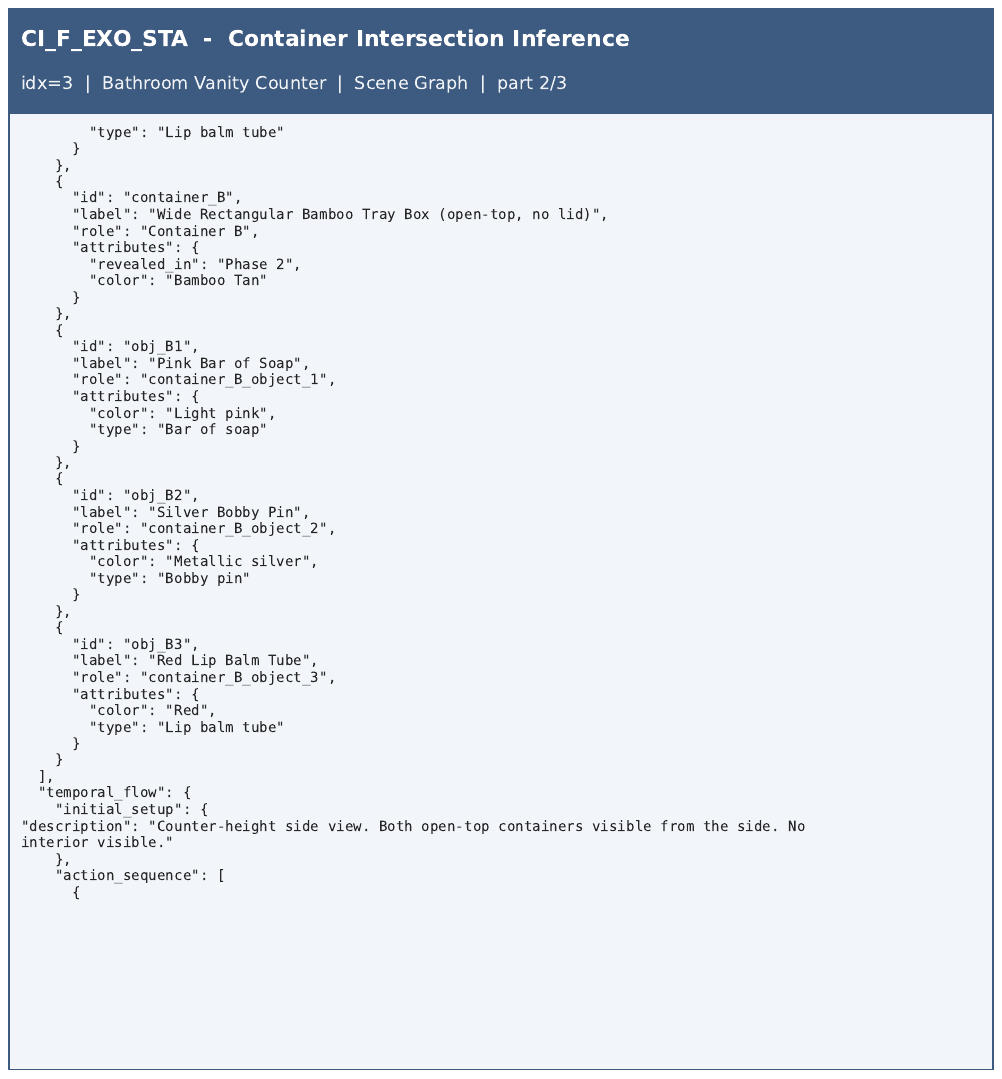}
\caption{\textbf{Scene graph (part 2/3)} for CI\_F\_EXO\_STA, idx 3.}
\label{fig:qual_CI_F_EXO_STA_sg_p2}
\end{figure}

\begin{figure}[H]
\centering
\includegraphics[width=\linewidth]{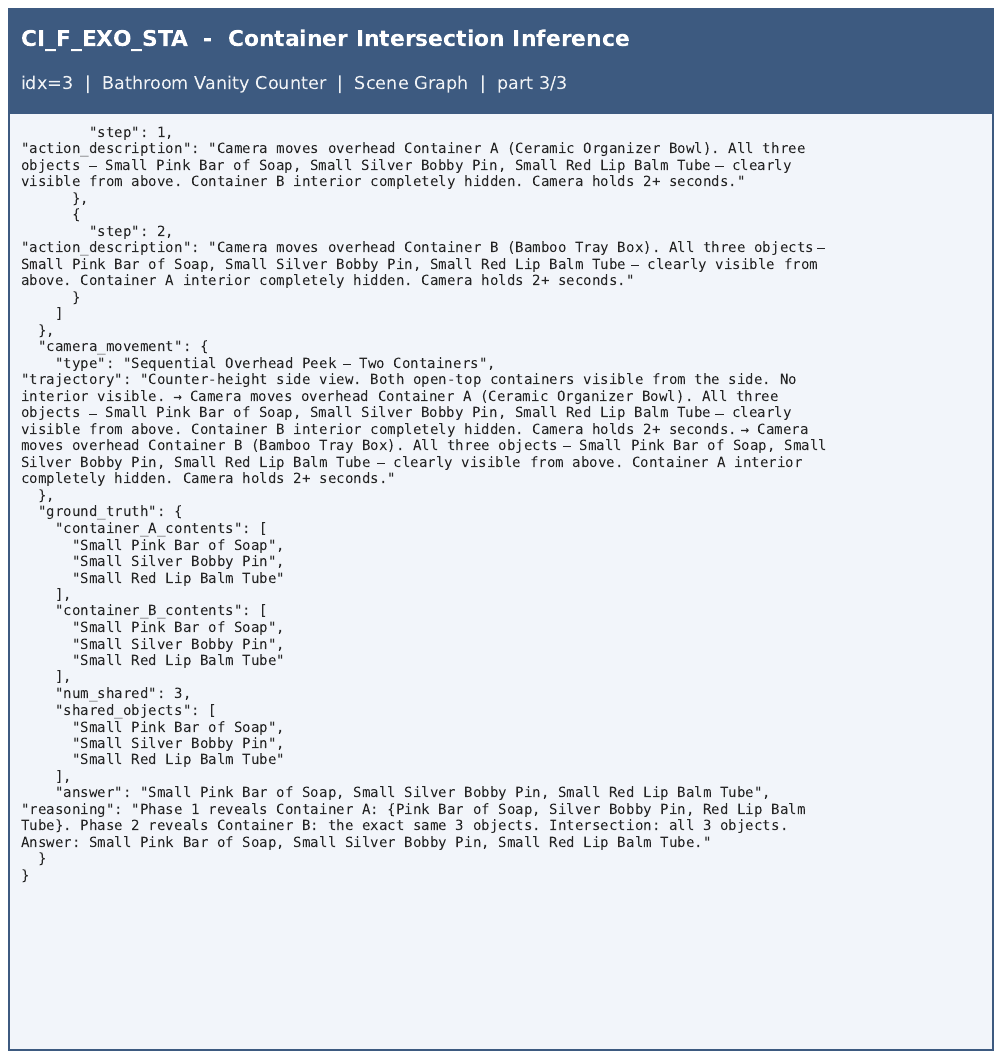}
\caption{\textbf{Scene graph (part 3/3)} for CI\_F\_EXO\_STA, idx 3.}
\label{fig:qual_CI_F_EXO_STA_sg_p3}
\end{figure}

\begin{figure}[H]
\centering
\includegraphics[width=\linewidth]{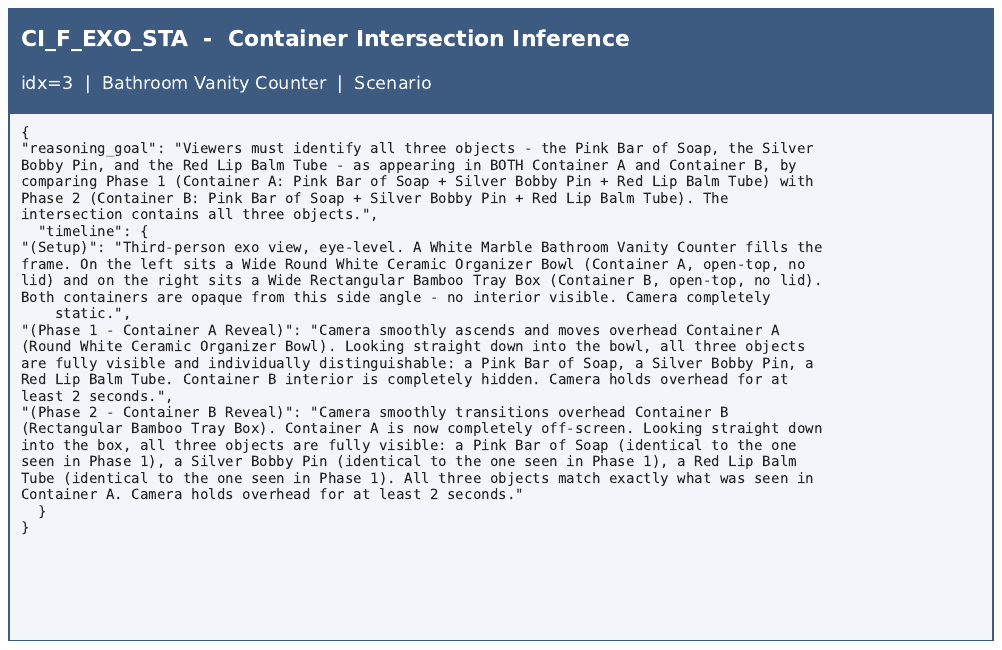}
\caption{\textbf{Scenario} for CI\_F\_EXO\_STA, idx 3.}
\label{fig:qual_CI_F_EXO_STA_scenario}
\end{figure}

\begin{figure}[H]
\centering
\includegraphics[width=\linewidth]{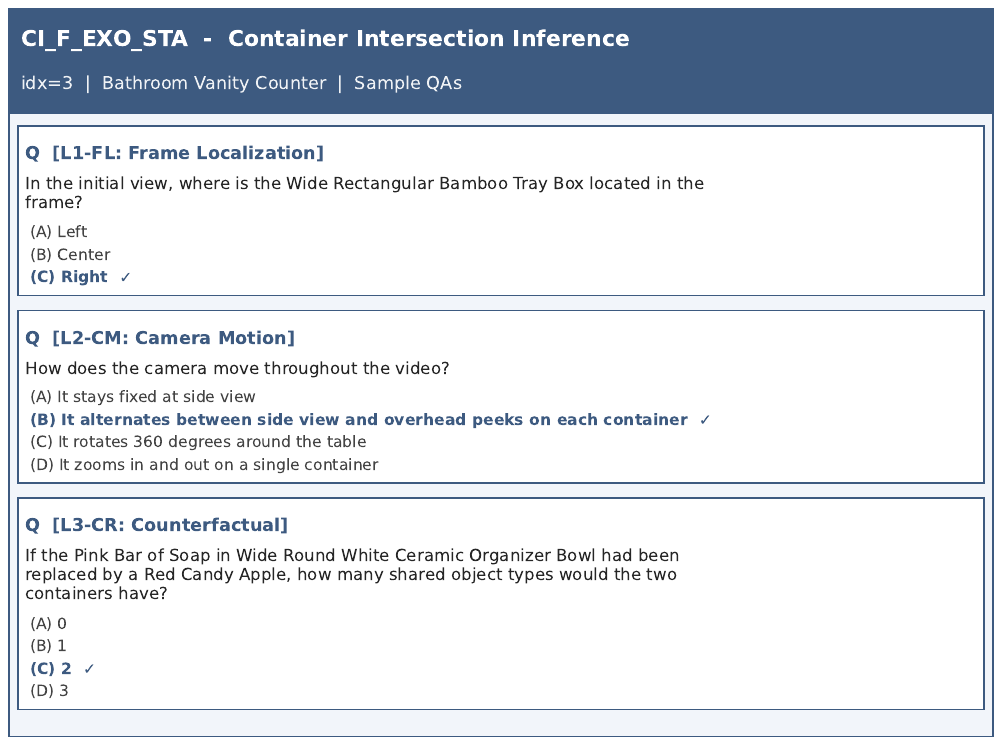}
\caption{\textbf{Sample QA pairs} for CI\_F\_EXO\_STA, idx 3.}
\label{fig:qual_CI_F_EXO_STA_qa}
\end{figure}


\begin{figure}[H]
\centering
\includegraphics[width=\linewidth]{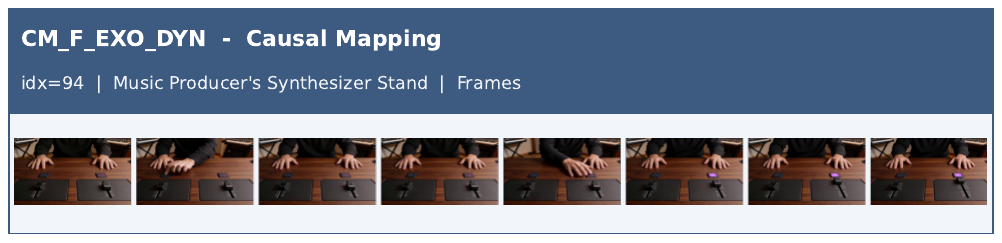}
\caption{\textbf{Frames} for CM\_F\_EXO\_DYN, idx 94 (\emph{Music Producer's Synthesizer Stand}).}
\label{fig:qual_CM_F_EXO_DYN_frames}
\end{figure}

\begin{figure}[H]
\centering
\includegraphics[width=\linewidth]{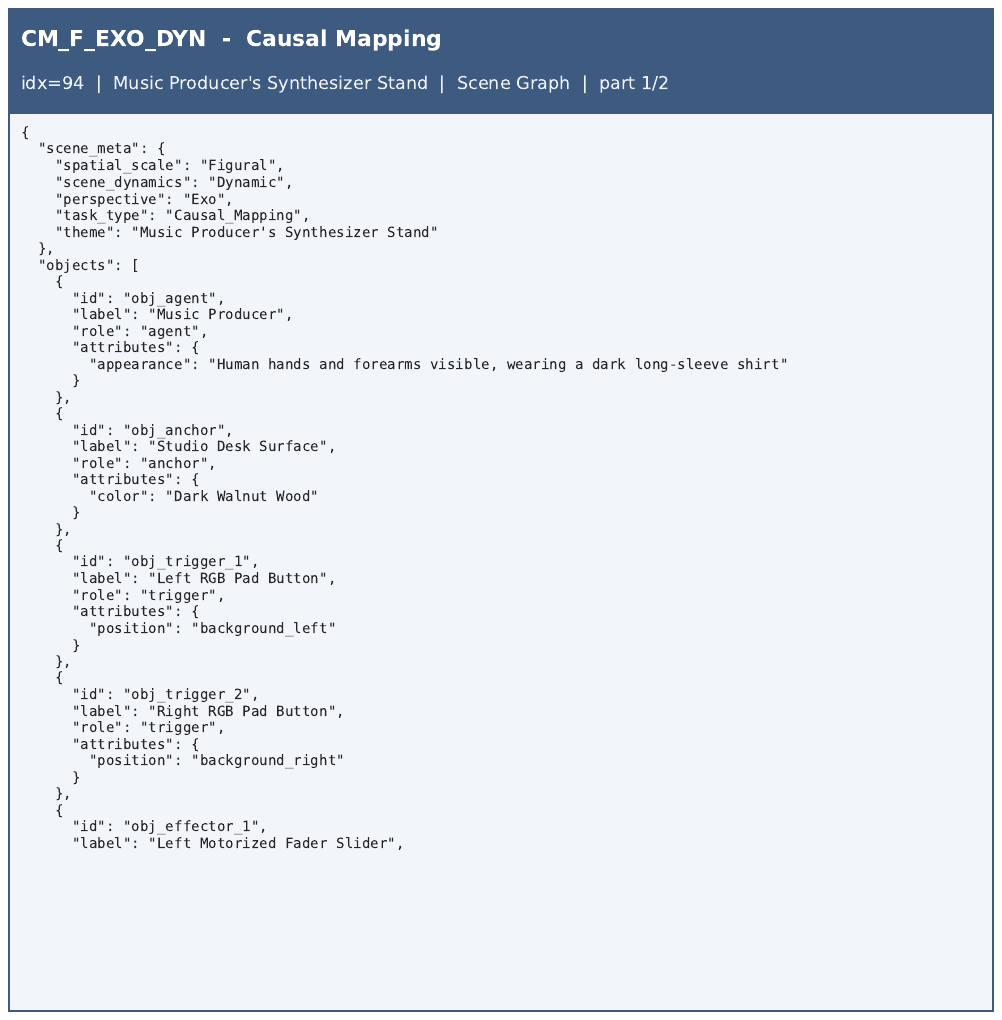}
\caption{\textbf{Scene graph (part 1/2)} for CM\_F\_EXO\_DYN, idx 94.}
\label{fig:qual_CM_F_EXO_DYN_sg_p1}
\end{figure}

\begin{figure}[H]
\centering
\includegraphics[width=\linewidth]{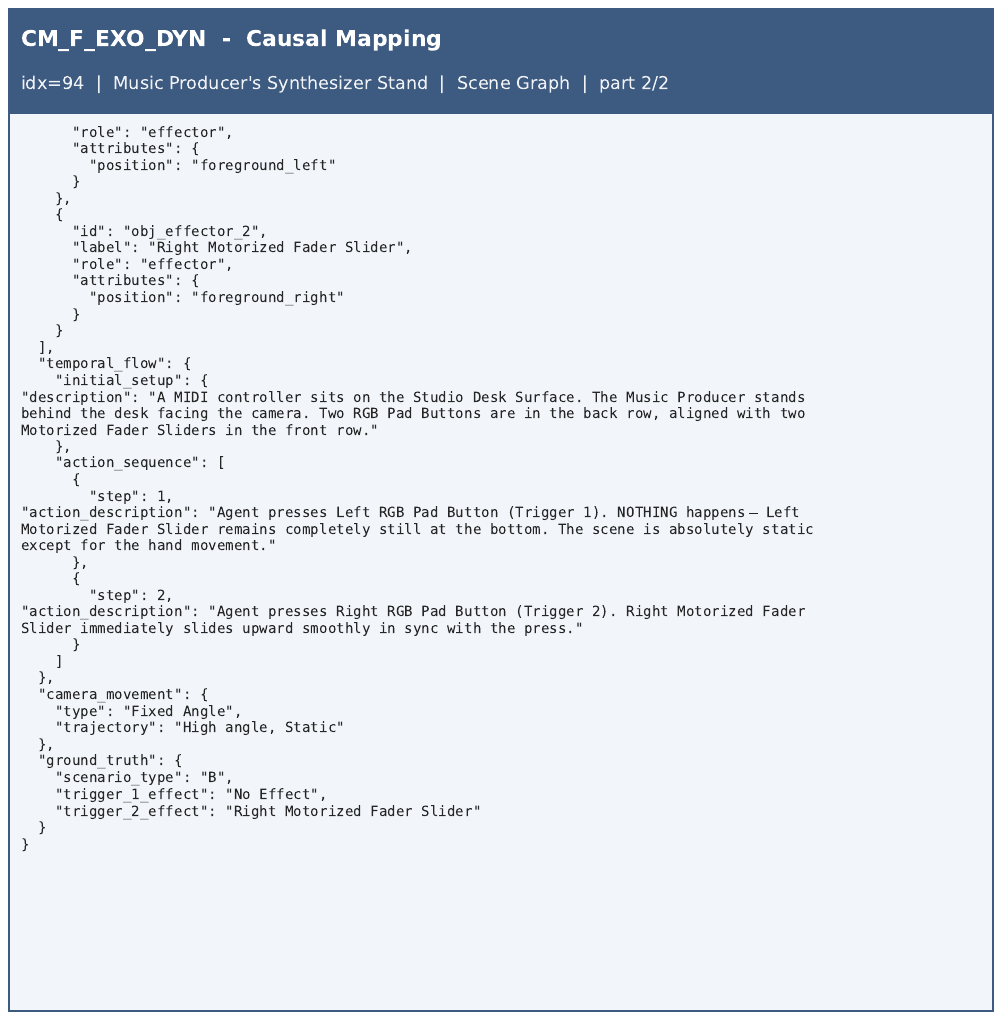}
\caption{\textbf{Scene graph (part 2/2)} for CM\_F\_EXO\_DYN, idx 94.}
\label{fig:qual_CM_F_EXO_DYN_sg_p2}
\end{figure}

\begin{figure}[H]
\centering
\includegraphics[width=\linewidth]{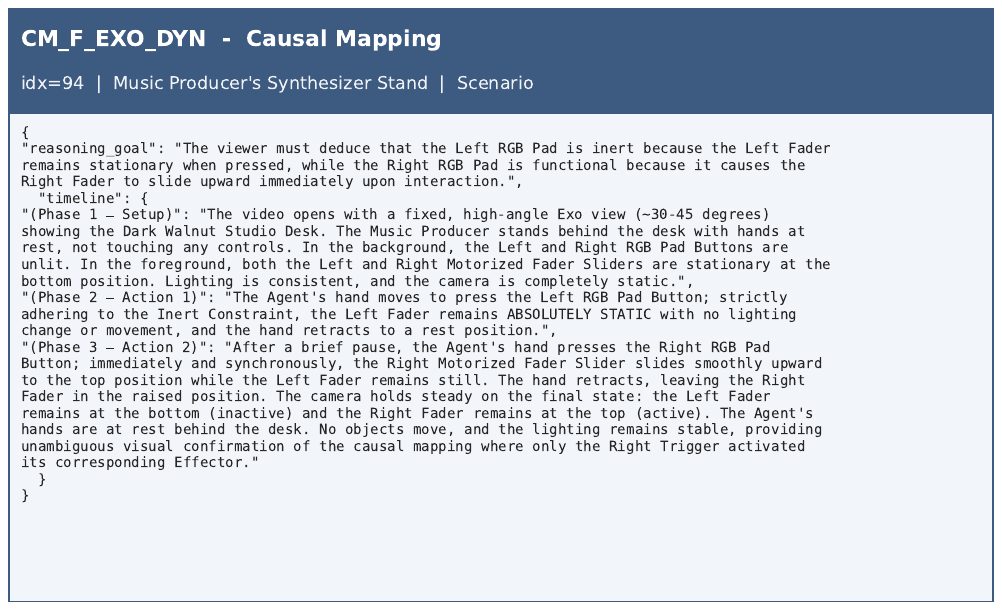}
\caption{\textbf{Scenario} for CM\_F\_EXO\_DYN, idx 94.}
\label{fig:qual_CM_F_EXO_DYN_scenario}
\end{figure}

\begin{figure}[H]
\centering
\includegraphics[width=\linewidth]{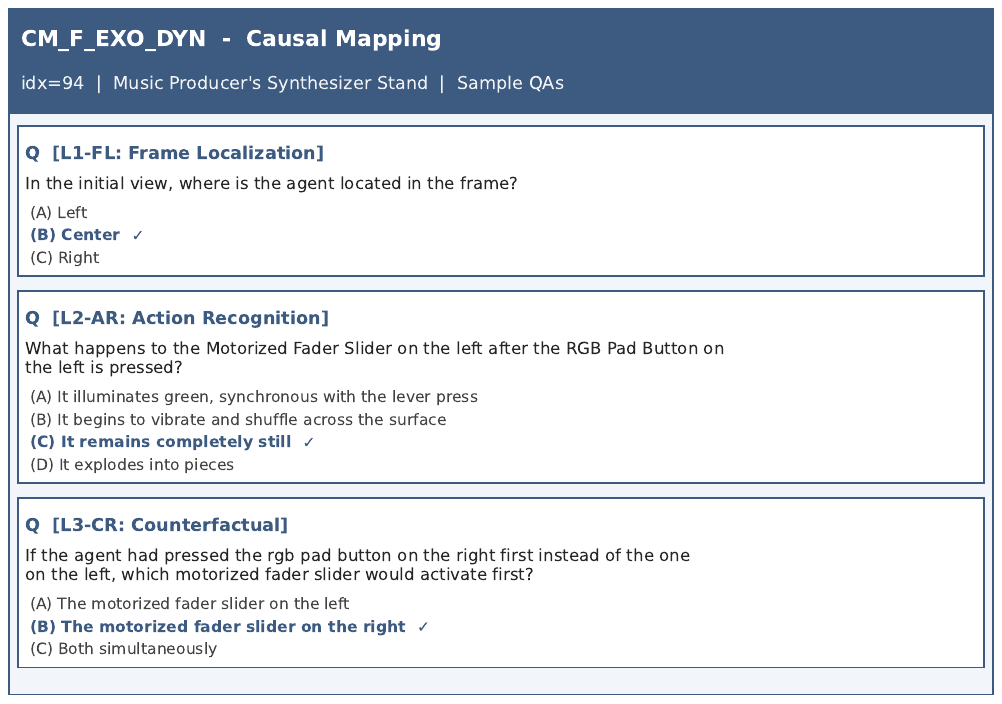}
\caption{\textbf{Sample QA pairs} for CM\_F\_EXO\_DYN, idx 94.}
\label{fig:qual_CM_F_EXO_DYN_qa}
\end{figure}


\begin{figure}[H]
\centering
\includegraphics[width=\linewidth]{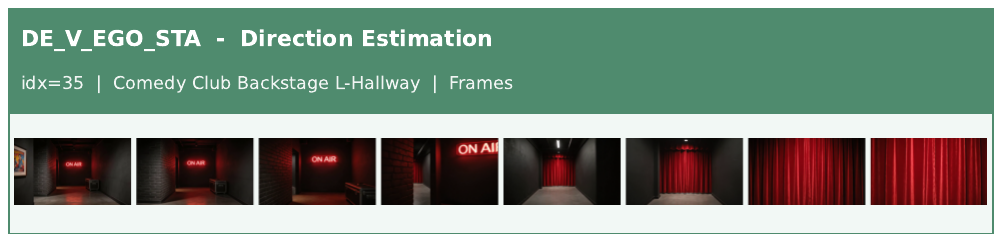}
\caption{\textbf{Frames} for DE\_V\_EGO\_STA, idx 35 (\emph{Comedy Club Backstage L-Hallway}).}
\label{fig:qual_DE_V_EGO_STA_frames}
\end{figure}

\begin{figure}[H]
\centering
\includegraphics[width=\linewidth]{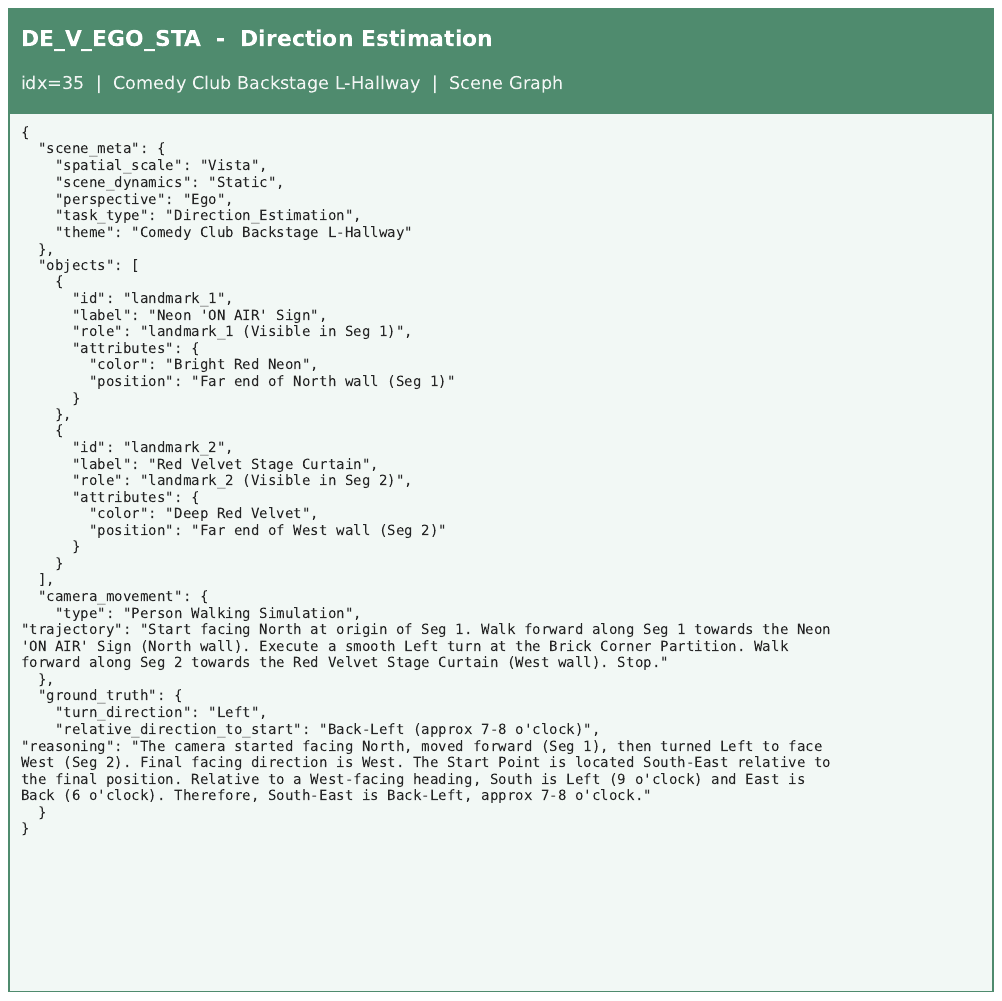}
\caption{\textbf{Scene graph} for DE\_V\_EGO\_STA, idx 35.}
\label{fig:qual_DE_V_EGO_STA_sg}
\end{figure}

\begin{figure}[H]
\centering
\includegraphics[width=\linewidth]{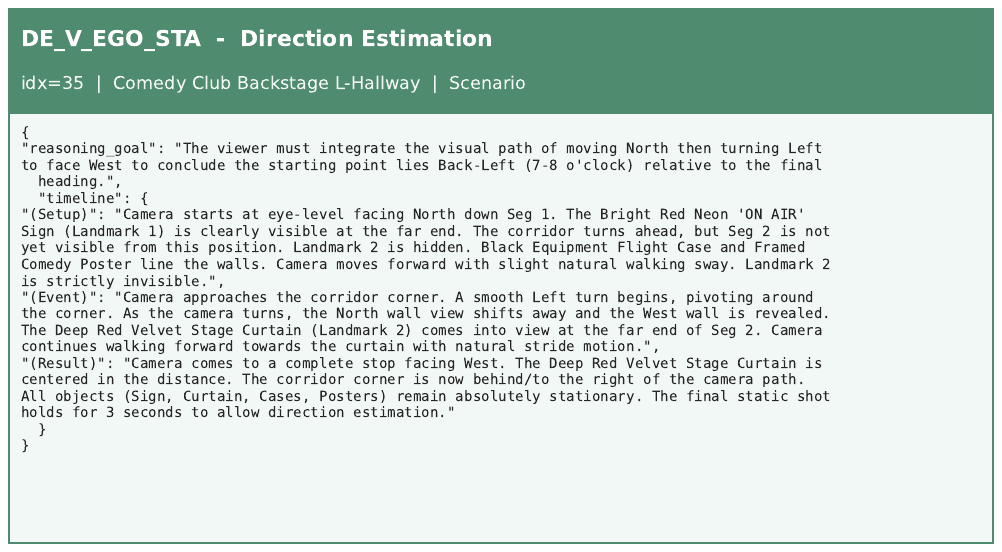}
\caption{\textbf{Scenario} for DE\_V\_EGO\_STA, idx 35.}
\label{fig:qual_DE_V_EGO_STA_scenario}
\end{figure}

\begin{figure}[H]
\centering
\includegraphics[width=\linewidth]{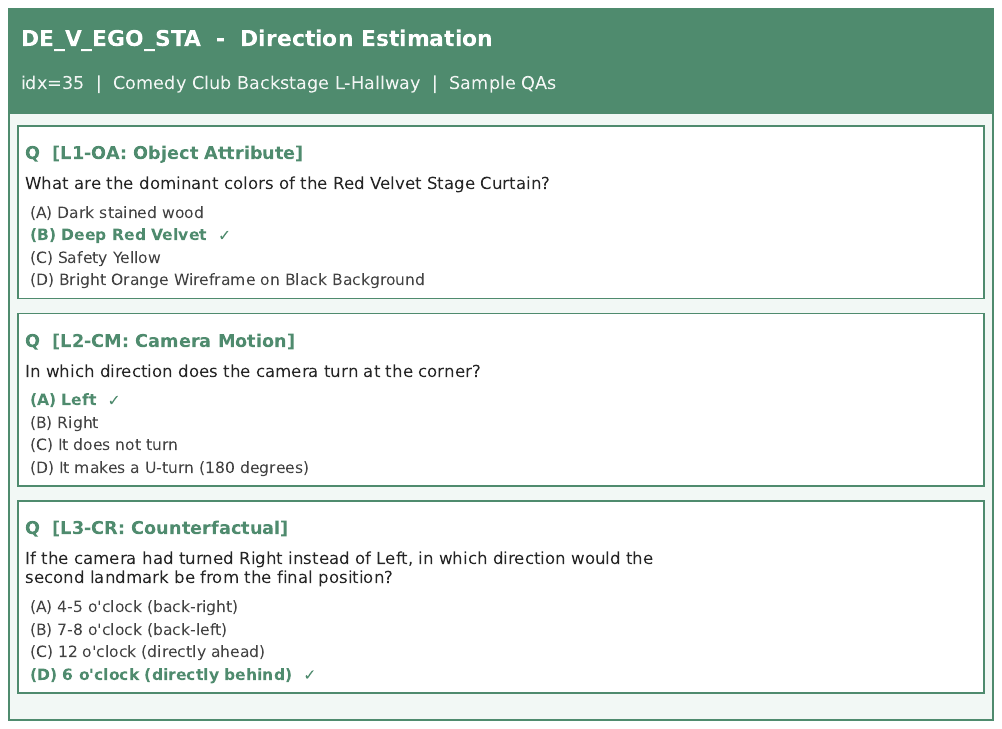}
\caption{\textbf{Sample QA pairs} for DE\_V\_EGO\_STA, idx 35.}
\label{fig:qual_DE_V_EGO_STA_qa}
\end{figure}


\begin{figure}[H]
\centering
\includegraphics[width=\linewidth]{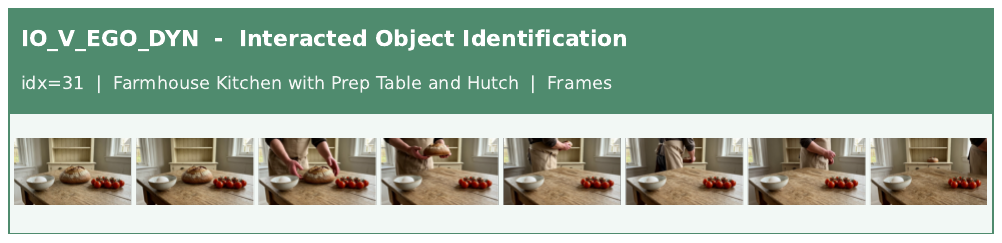}
\caption{\textbf{Frames} for IO\_V\_EGO\_DYN, idx 31 (\emph{Farmhouse Kitchen with Prep Table and Hutch}).}
\label{fig:qual_IO_V_EGO_DYN_frames}
\end{figure}

\begin{figure}[H]
\centering
\includegraphics[width=\linewidth]{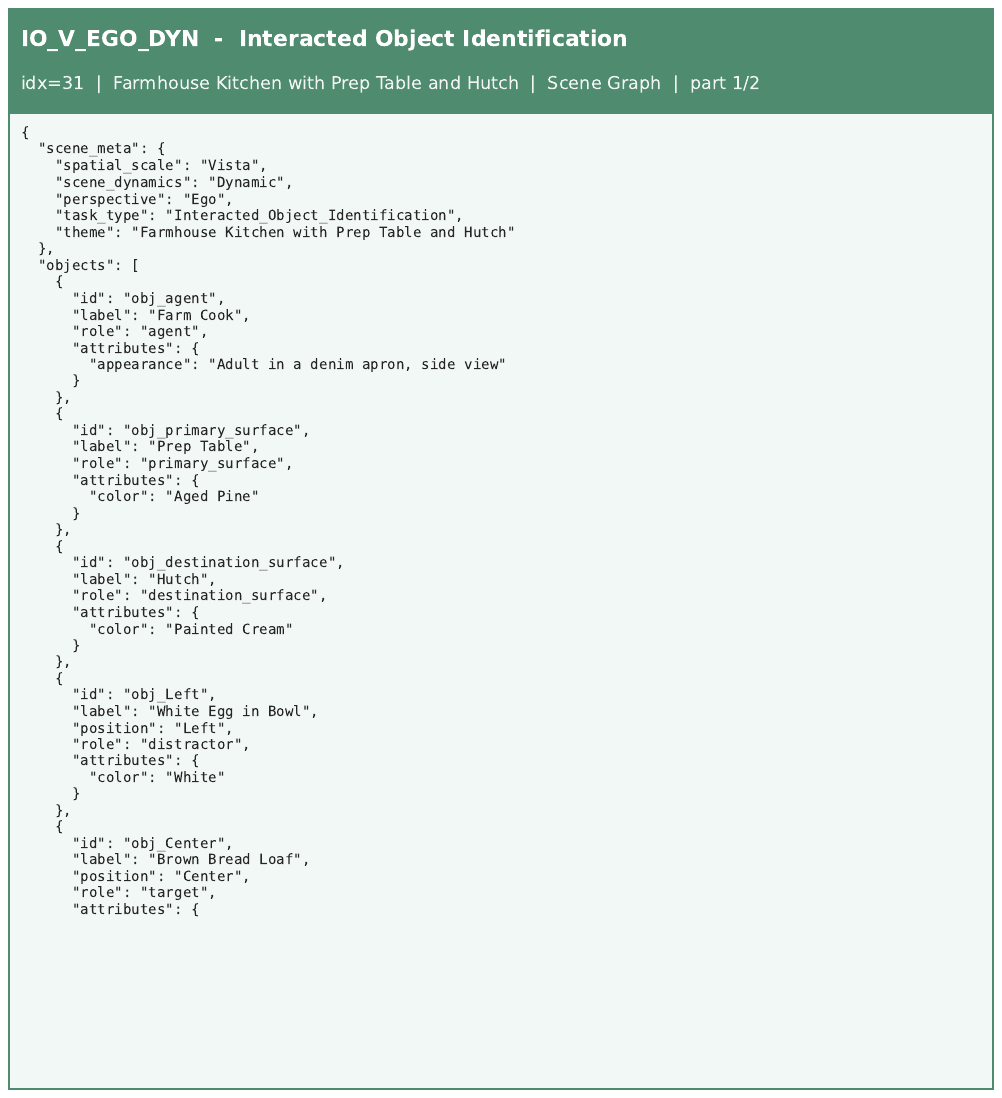}
\caption{\textbf{Scene graph (part 1/2)} for IO\_V\_EGO\_DYN, idx 31.}
\label{fig:qual_IO_V_EGO_DYN_sg_p1}
\end{figure}

\begin{figure}[H]
\centering
\includegraphics[width=\linewidth]{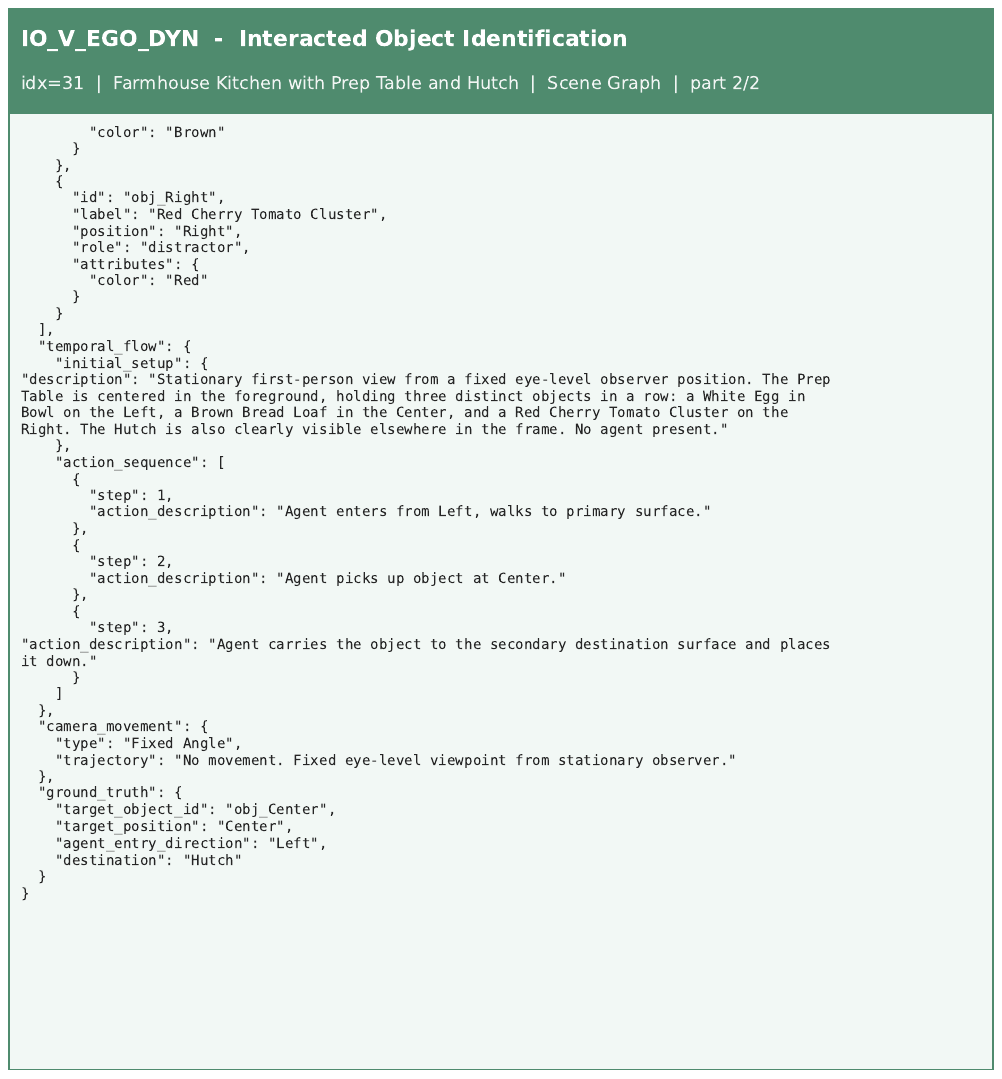}
\caption{\textbf{Scene graph (part 2/2)} for IO\_V\_EGO\_DYN, idx 31.}
\label{fig:qual_IO_V_EGO_DYN_sg_p2}
\end{figure}

\begin{figure}[H]
\centering
\includegraphics[width=\linewidth]{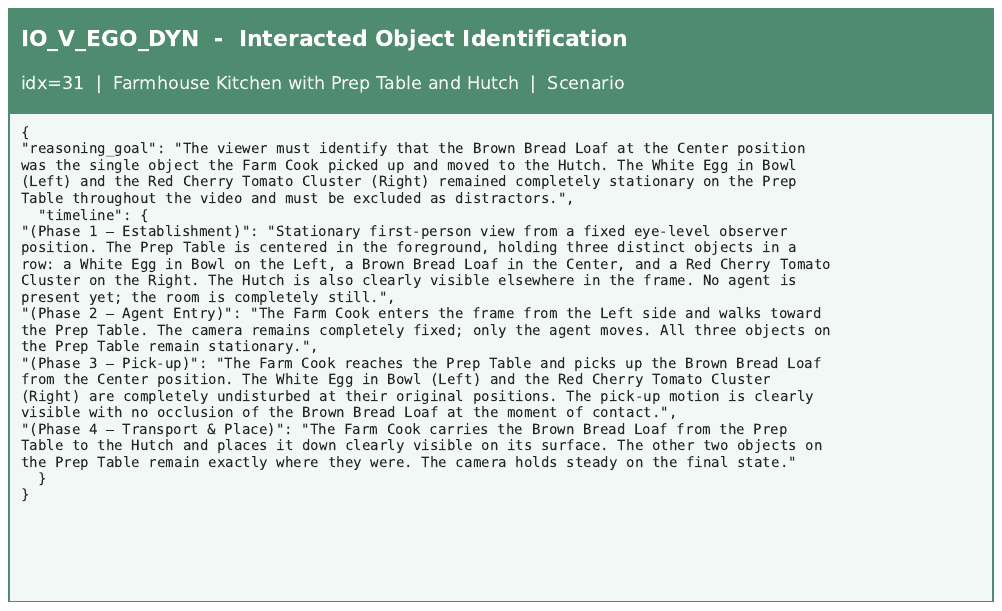}
\caption{\textbf{Scenario} for IO\_V\_EGO\_DYN, idx 31.}
\label{fig:qual_IO_V_EGO_DYN_scenario}
\end{figure}

\begin{figure}[H]
\centering
\includegraphics[width=\linewidth]{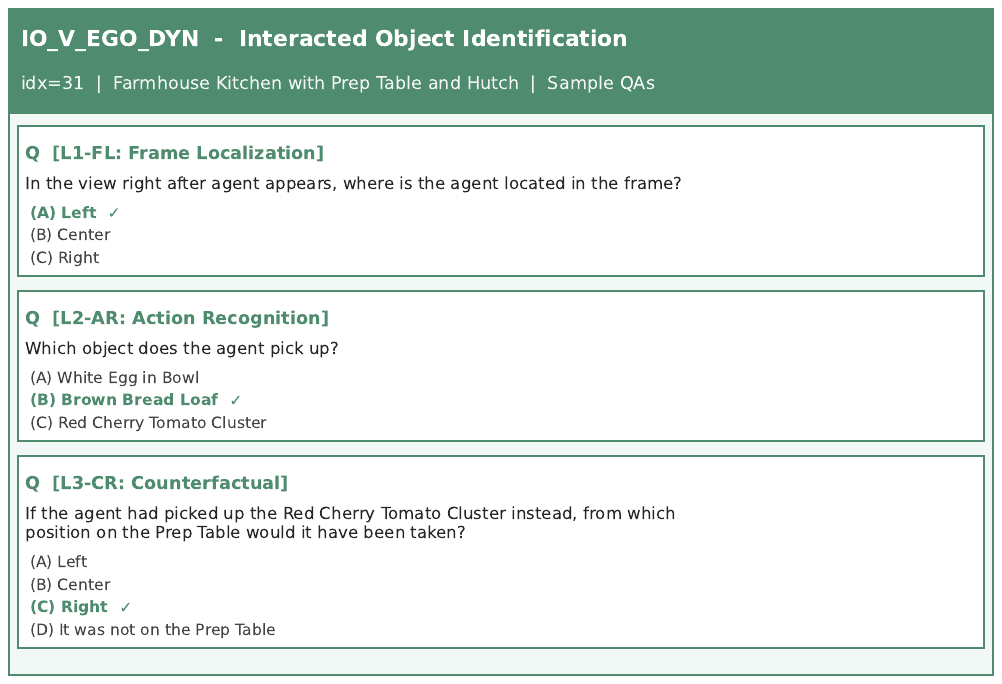}
\caption{\textbf{Sample QA pairs} for IO\_V\_EGO\_DYN, idx 31.}
\label{fig:qual_IO_V_EGO_DYN_qa}
\end{figure}


\begin{figure}[H]
\centering
\includegraphics[width=\linewidth]{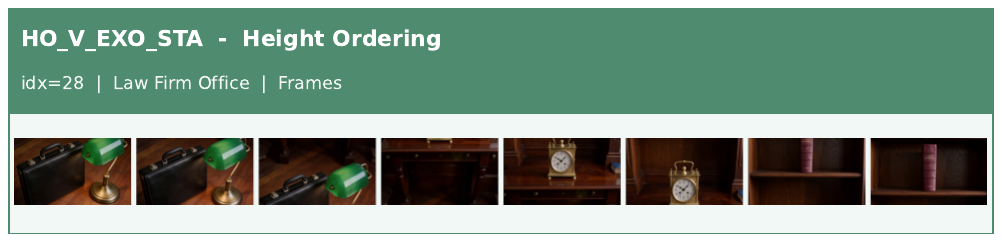}
\caption{\textbf{Frames} for HO\_V\_EXO\_STA, idx 28 (\emph{Law Firm Office}).}
\label{fig:qual_HO_V_EXO_STA_frames}
\end{figure}

\begin{figure}[H]
\centering
\includegraphics[width=\linewidth]{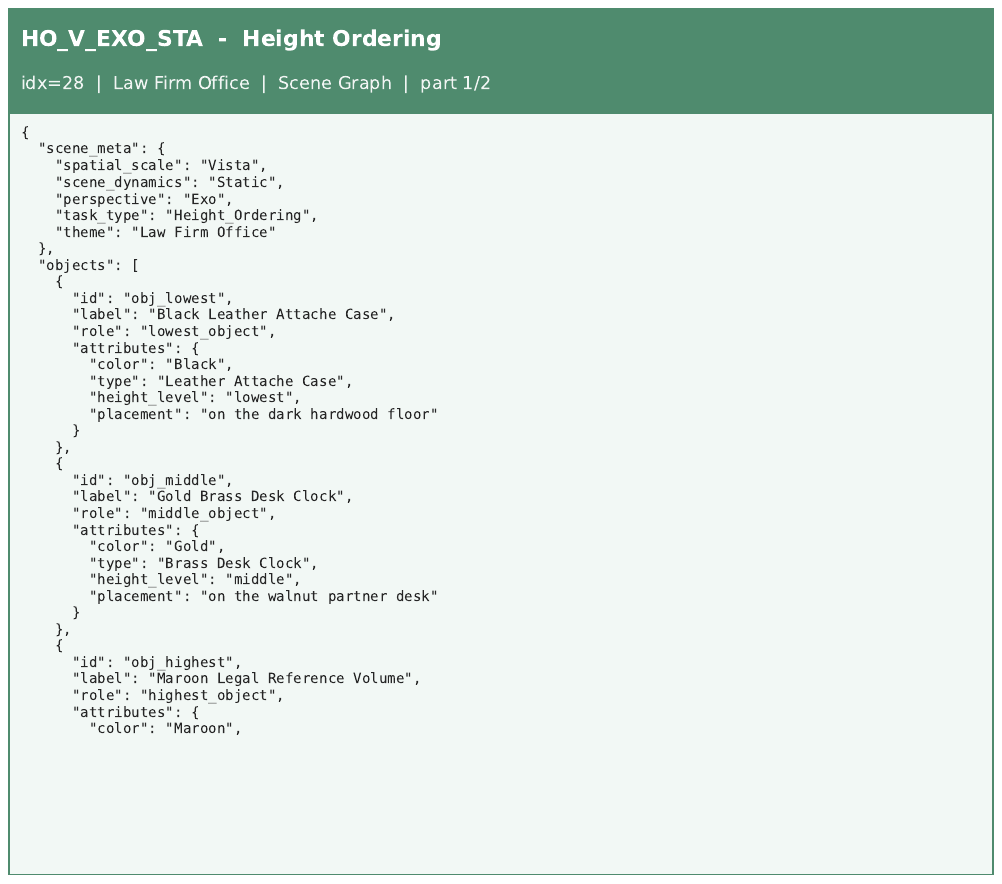}
\caption{\textbf{Scene graph (part 1/2)} for HO\_V\_EXO\_STA, idx 28.}
\label{fig:qual_HO_V_EXO_STA_sg_p1}
\end{figure}

\begin{figure}[H]
\centering
\includegraphics[width=\linewidth]{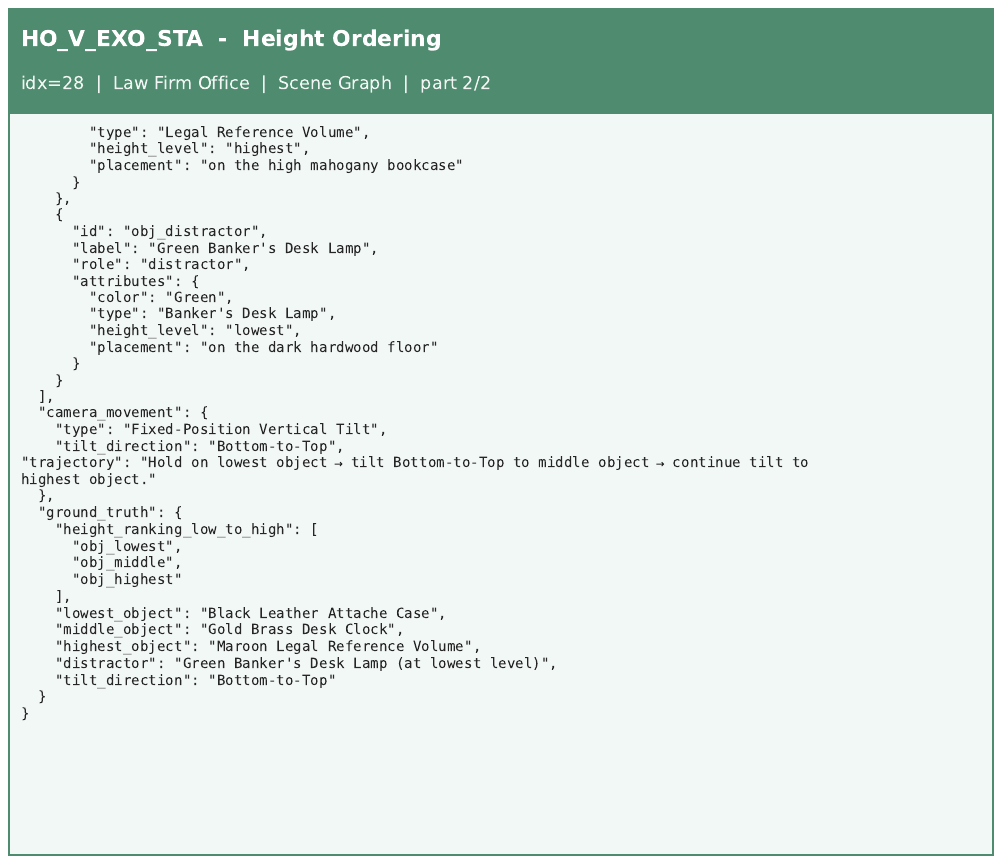}
\caption{\textbf{Scene graph (part 2/2)} for HO\_V\_EXO\_STA, idx 28.}
\label{fig:qual_HO_V_EXO_STA_sg_p2}
\end{figure}

\begin{figure}[H]
\centering
\includegraphics[width=\linewidth]{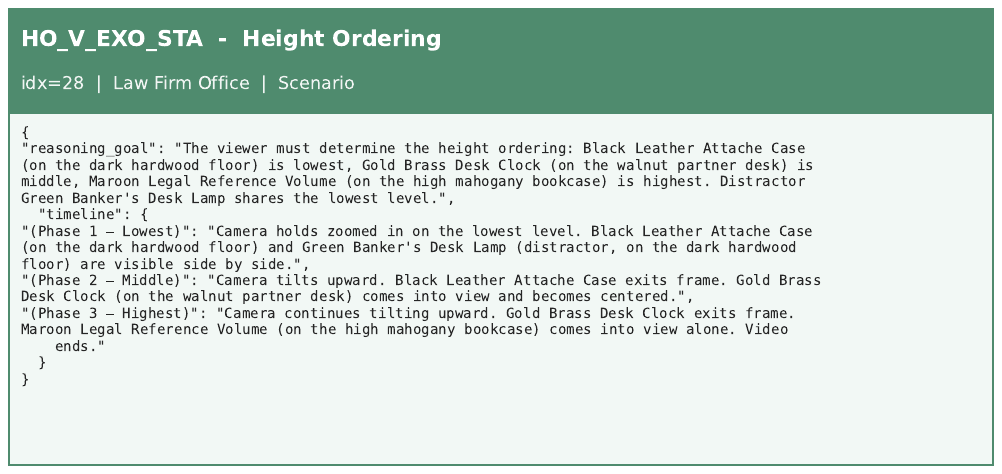}
\caption{\textbf{Scenario} for HO\_V\_EXO\_STA, idx 28.}
\label{fig:qual_HO_V_EXO_STA_scenario}
\end{figure}

\begin{figure}[H]
\centering
\includegraphics[width=\linewidth]{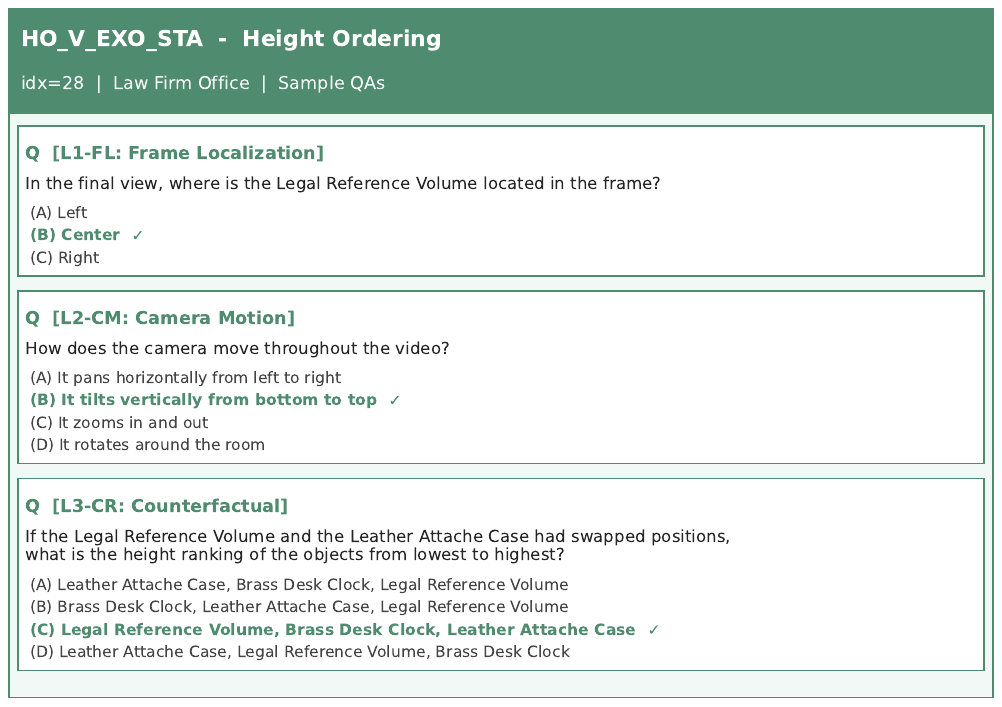}
\caption{\textbf{Sample QA pairs} for HO\_V\_EXO\_STA, idx 28.}
\label{fig:qual_HO_V_EXO_STA_qa}
\end{figure}

\newpage
\begin{figure}[H]
\centering
\includegraphics[width=\linewidth]{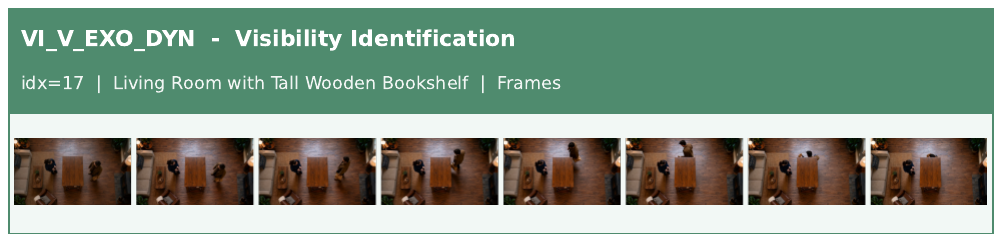}
\caption{\textbf{Frames} for VI\_V\_EXO\_DYN, idx 17 (\emph{Living Room with Tall Wooden Bookshelf}).}
\label{fig:qual_VI_V_EXO_DYN_frames}
\end{figure}

\begin{figure}[H]
\centering
\includegraphics[width=\linewidth]{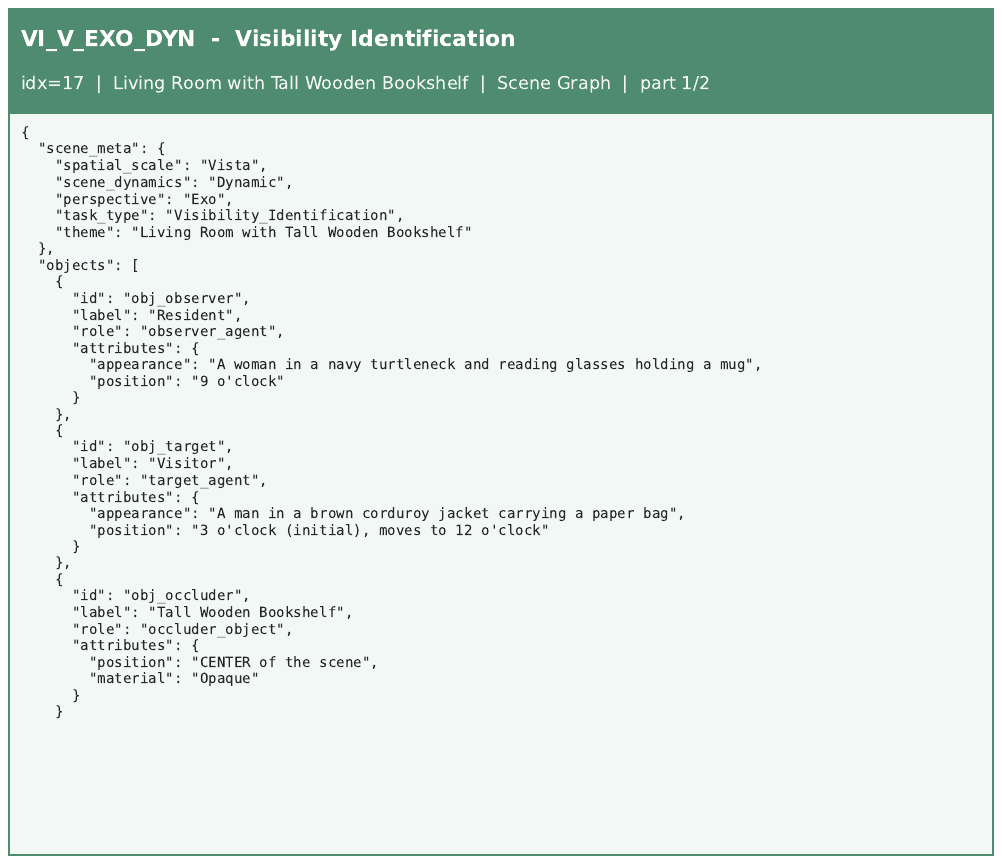}
\caption{\textbf{Scene graph (part 1/2)} for VI\_V\_EXO\_DYN, idx 17.}
\label{fig:qual_VI_V_EXO_DYN_sg_p1}
\end{figure}

\begin{figure}[H]
\centering
\includegraphics[width=\linewidth]{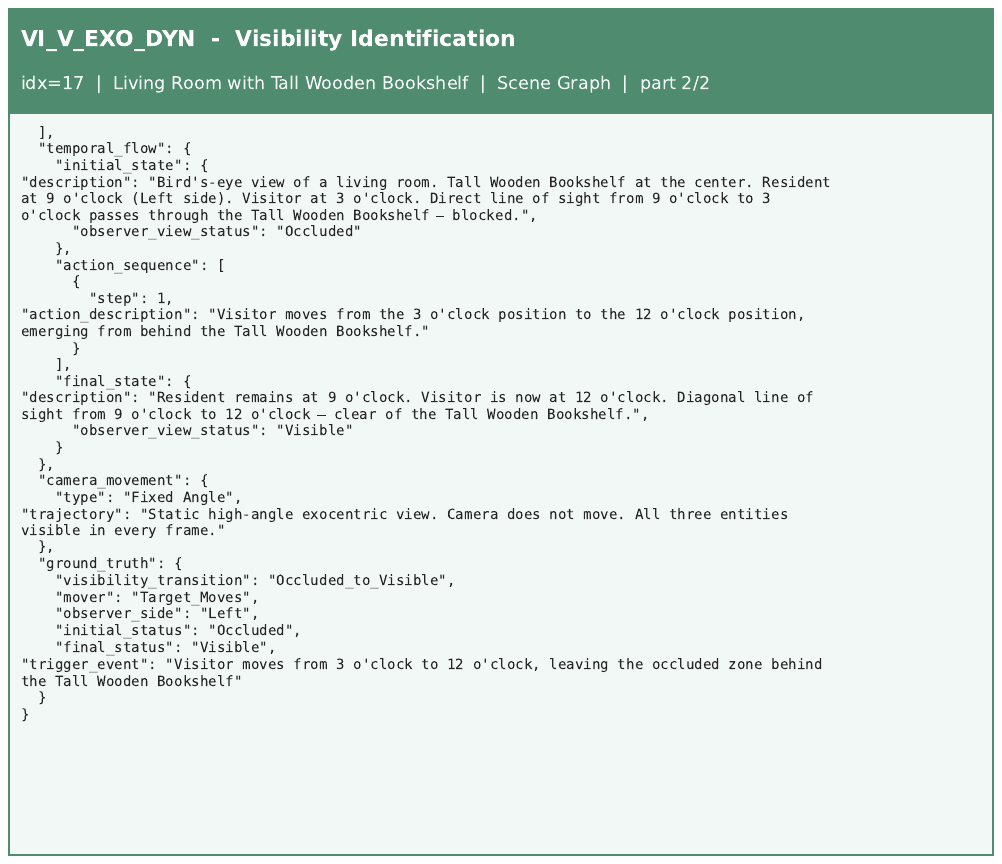}
\caption{\textbf{Scene graph (part 2/2)} for VI\_V\_EXO\_DYN, idx 17.}
\label{fig:qual_VI_V_EXO_DYN_sg_p2}
\end{figure}

\begin{figure}[H]
\centering
\includegraphics[width=\linewidth]{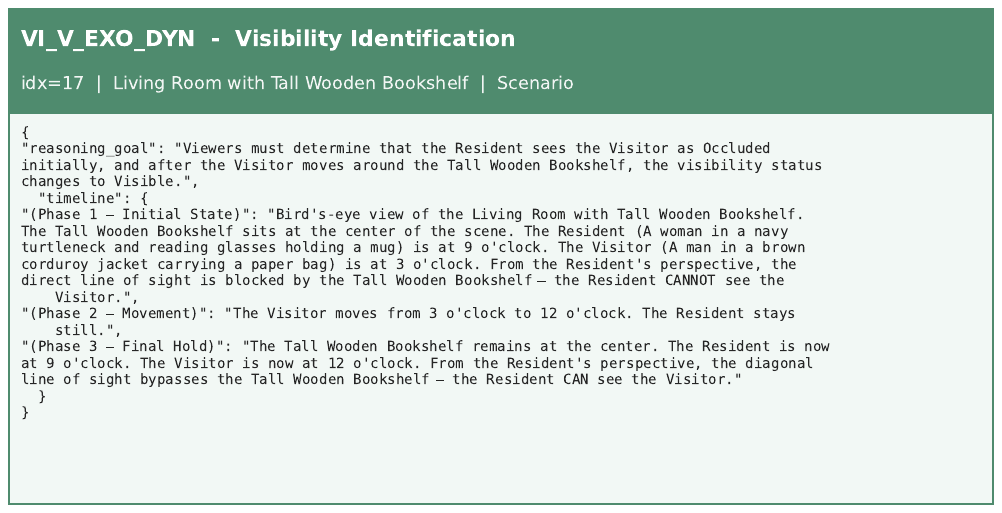}
\caption{\textbf{Scenario} for VI\_V\_EXO\_DYN, idx 17.}
\label{fig:qual_VI_V_EXO_DYN_scenario}
\end{figure}

\begin{figure}[H]
\centering
\includegraphics[width=\linewidth]{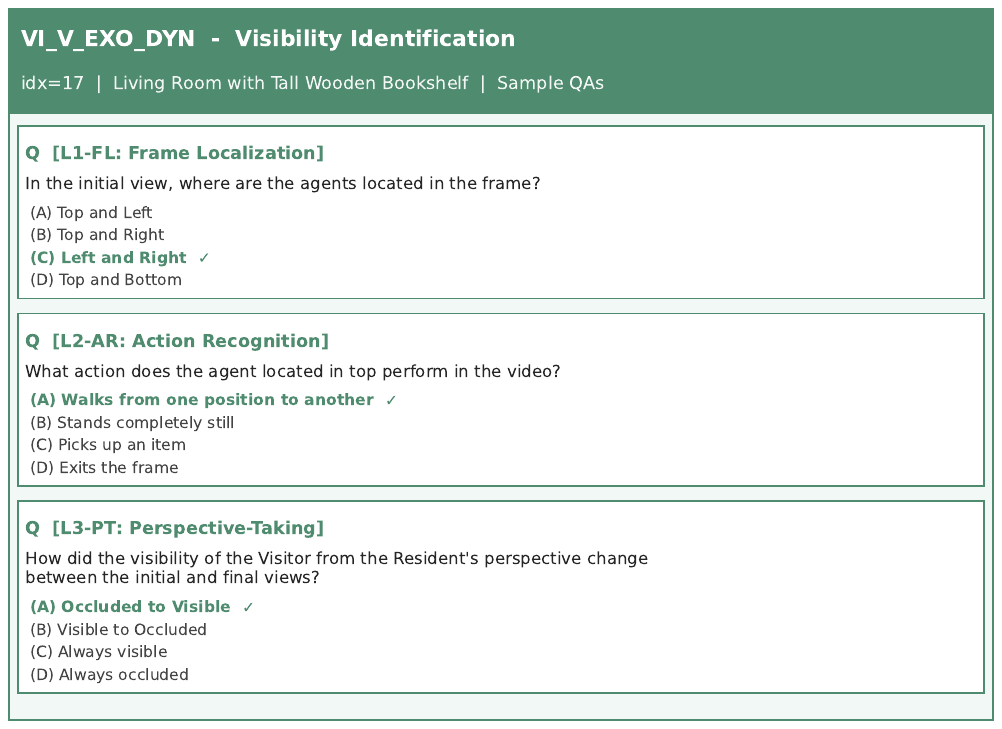}
\caption{\textbf{Sample QA pairs} for VI\_V\_EXO\_DYN, idx 17.}
\label{fig:qual_VI_V_EXO_DYN_qa}
\end{figure}

\newpage
\begin{figure}[H]
\centering
\includegraphics[width=\linewidth]{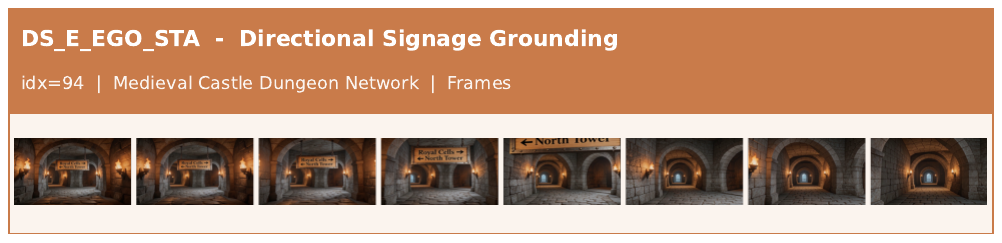}
\caption{\textbf{Frames} for DS\_E\_EGO\_STA, idx 94 (\emph{Medieval Castle Dungeon Network}).}
\label{fig:qual_DS_E_EGO_STA_frames}
\end{figure}

\begin{figure}[H]
\centering
\includegraphics[width=\linewidth]{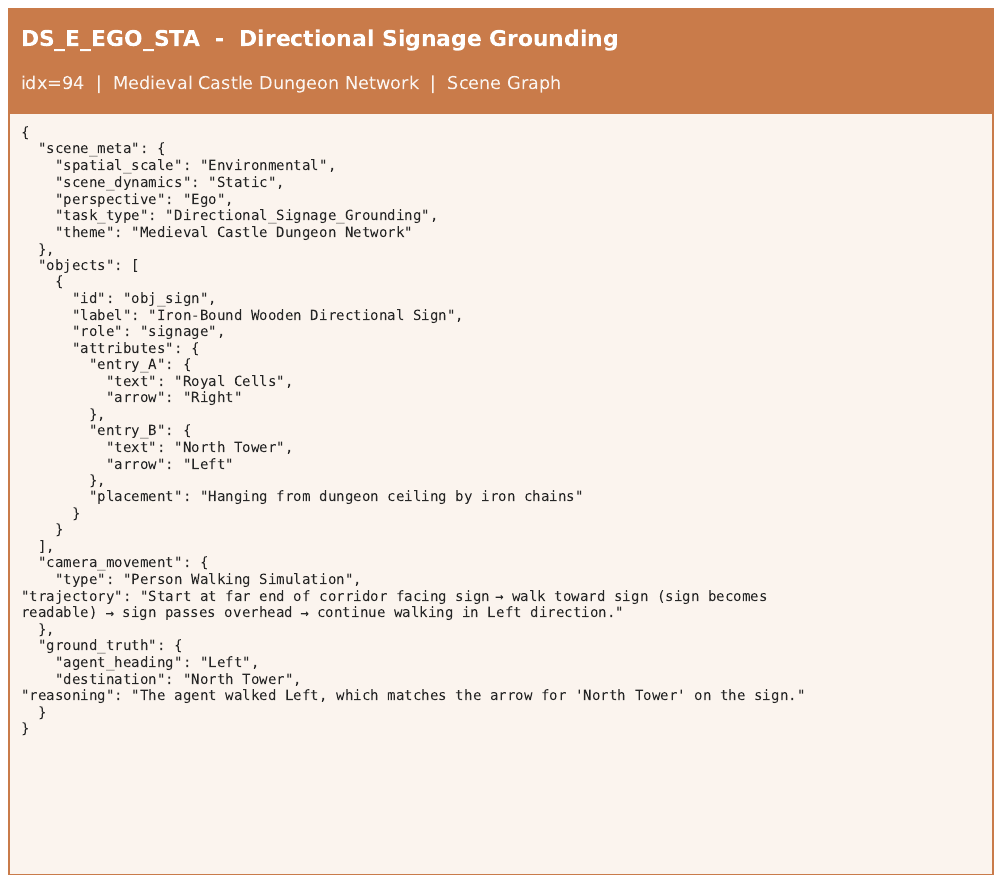}
\caption{\textbf{Scene graph} for DS\_E\_EGO\_STA, idx 94.}
\label{fig:qual_DS_E_EGO_STA_sg}
\end{figure}

\begin{figure}[H]
\centering
\includegraphics[width=\linewidth]{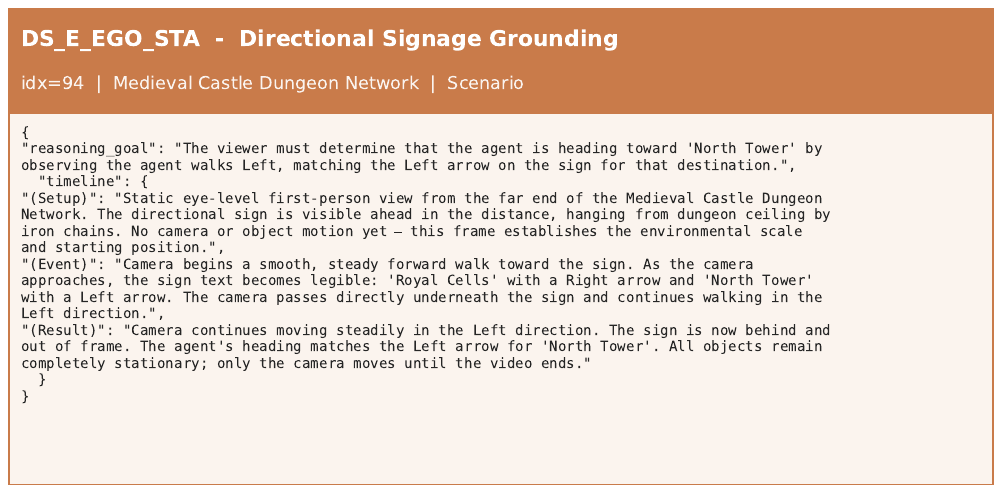}
\caption{\textbf{Scenario} for DS\_E\_EGO\_STA, idx 94.}
\label{fig:qual_DS_E_EGO_STA_scenario}
\end{figure}

\begin{figure}[H]
\centering
\includegraphics[width=\linewidth]{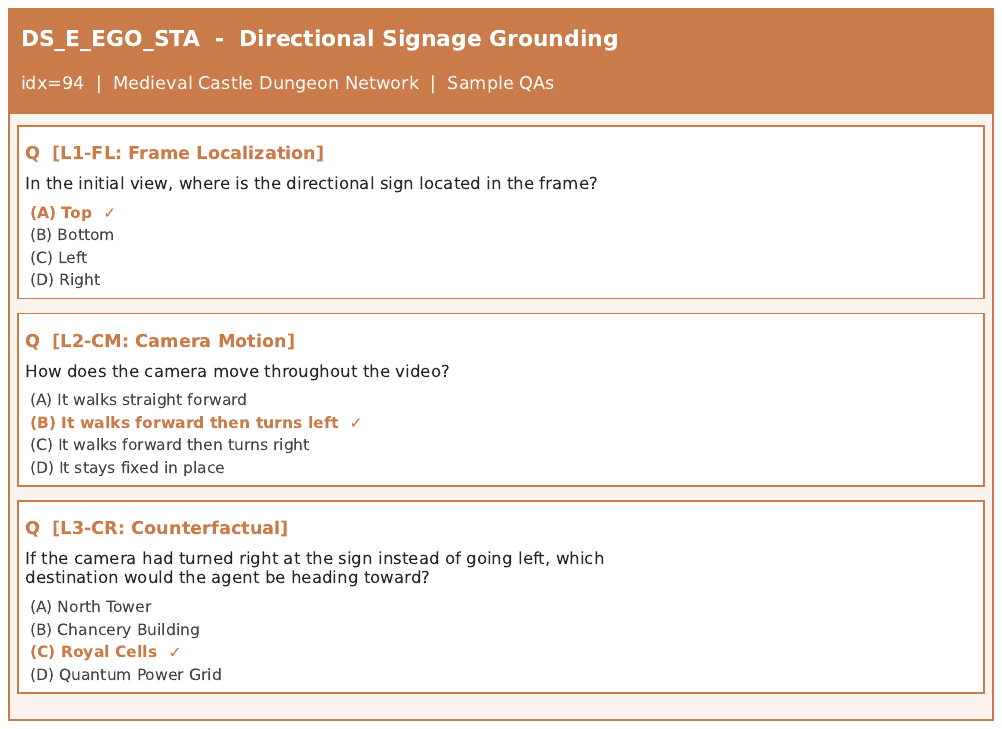}
\caption{\textbf{Sample QA pairs} for DS\_E\_EGO\_STA, idx 94.}
\label{fig:qual_DS_E_EGO_STA_qa}
\end{figure}


\begin{figure}[H]
\centering
\includegraphics[width=\linewidth]{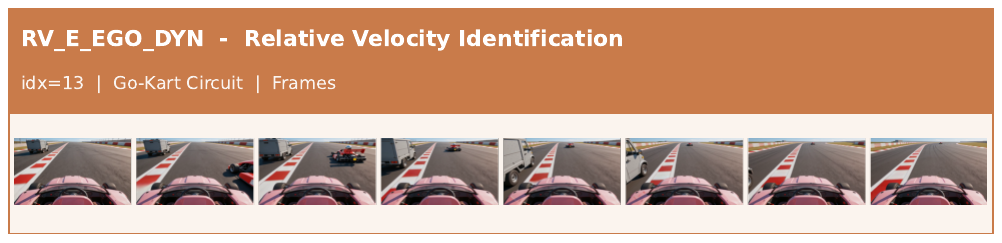}
\caption{\textbf{Frames} for RV\_E\_EGO\_DYN, idx 13 (\emph{Go-Kart Circuit}).}
\label{fig:qual_RV_E_EGO_DYN_frames}
\end{figure}

\begin{figure}[H]
\centering
\includegraphics[width=\linewidth]{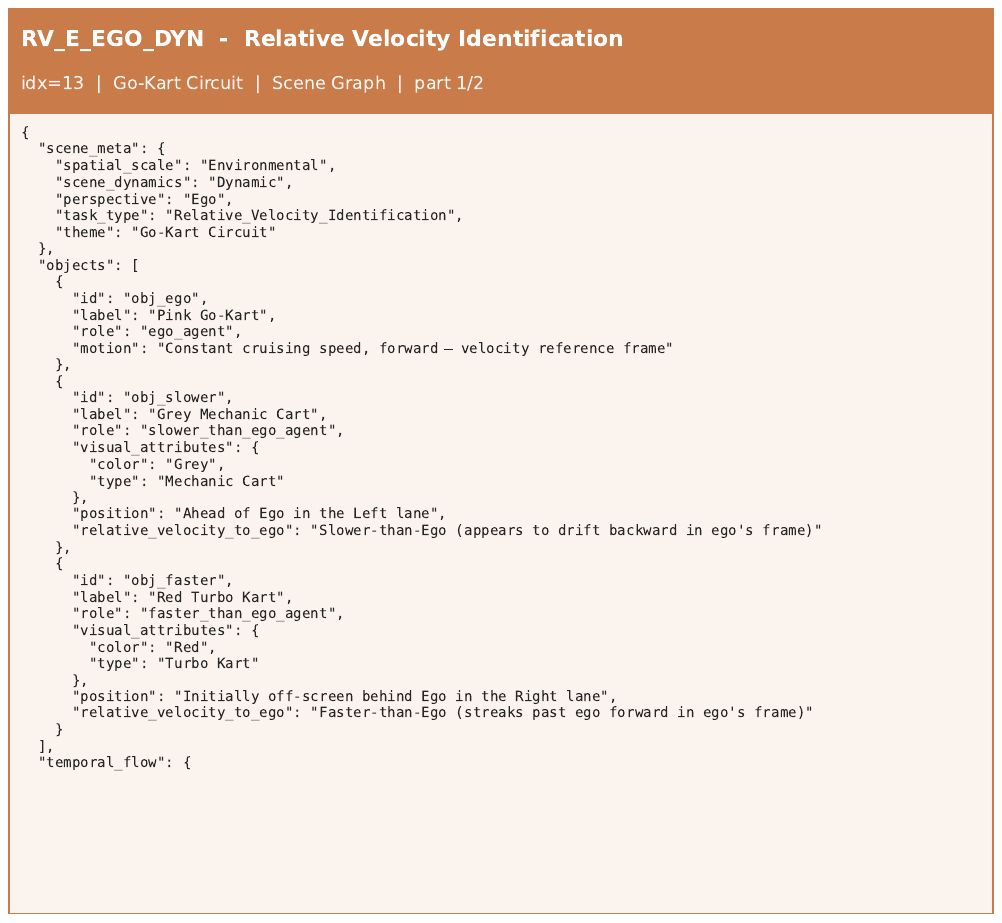}
\caption{\textbf{Scene graph (part 1/2)} for RV\_E\_EGO\_DYN, idx 13.}
\label{fig:qual_RV_E_EGO_DYN_sg_p1}
\end{figure}

\begin{figure}[H]
\centering
\includegraphics[width=\linewidth]{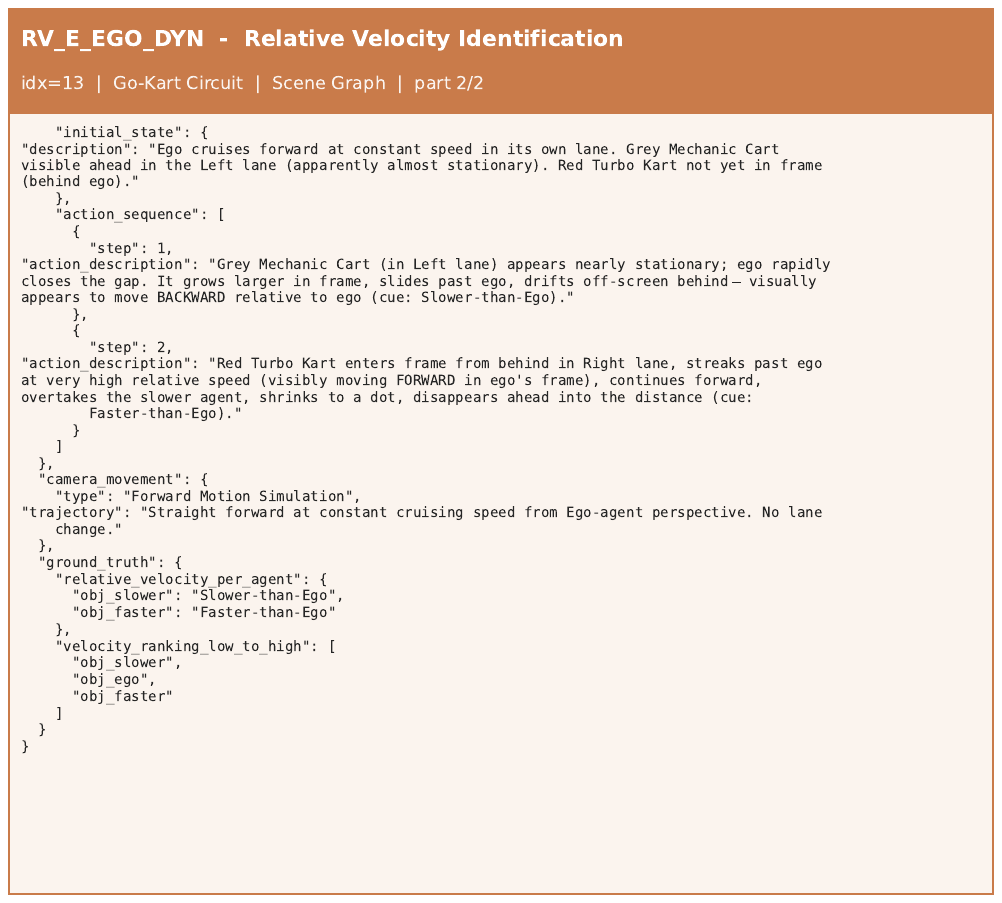}
\caption{\textbf{Scene graph (part 2/2)} for RV\_E\_EGO\_DYN, idx 13.}
\label{fig:qual_RV_E_EGO_DYN_sg_p2}
\end{figure}

\begin{figure}[H]
\centering
\includegraphics[width=\linewidth]{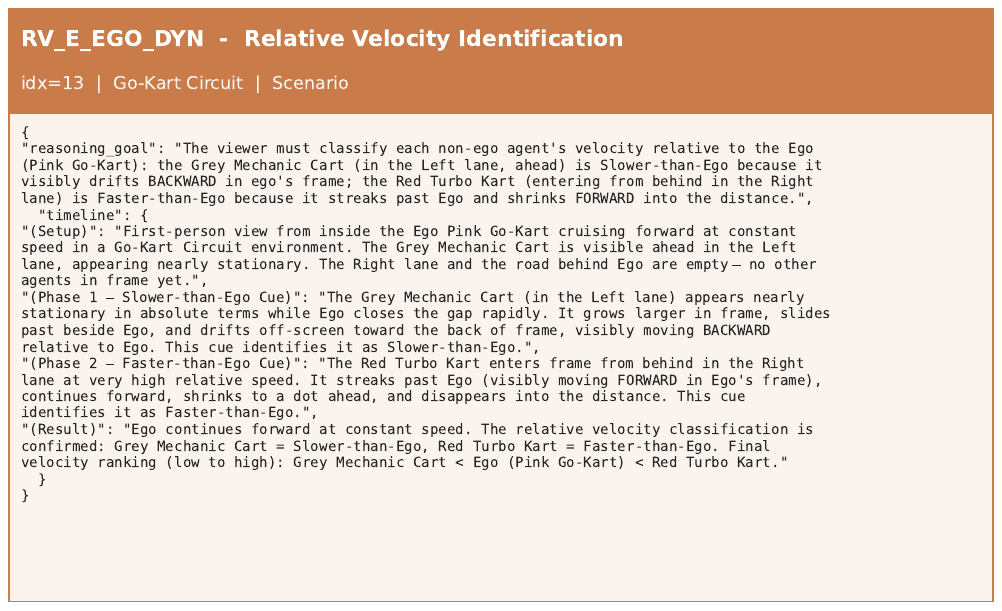}
\caption{\textbf{Scenario} for RV\_E\_EGO\_DYN, idx 13.}
\label{fig:qual_RV_E_EGO_DYN_scenario}
\end{figure}

\begin{figure}[H]
\centering
\includegraphics[width=\linewidth]{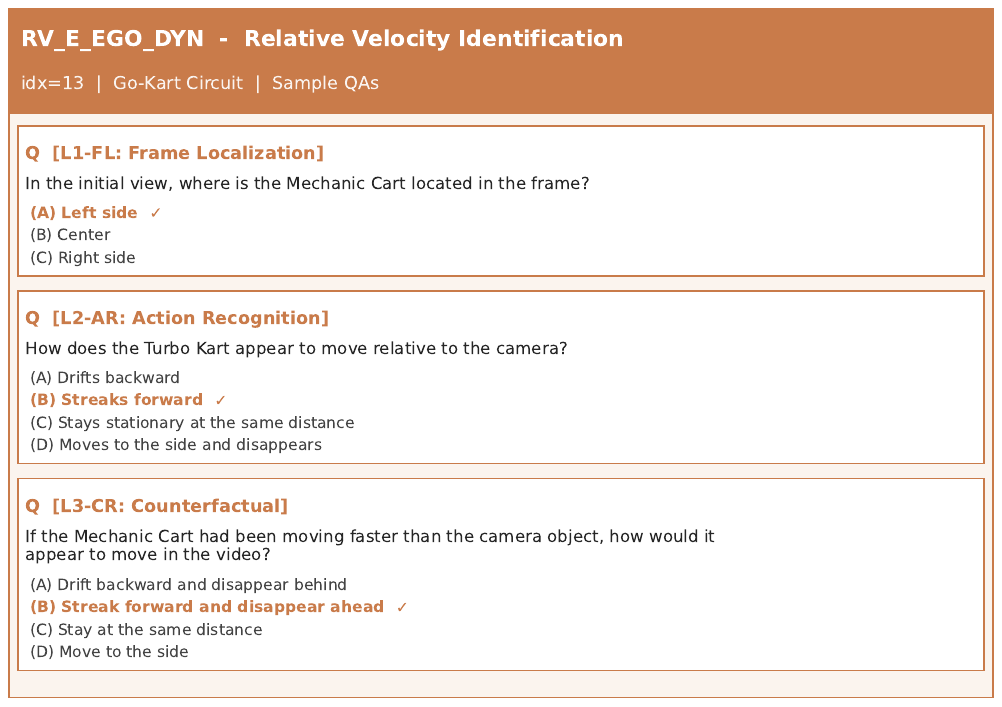}
\caption{\textbf{Sample QA pairs} for RV\_E\_EGO\_DYN, idx 13.}
\label{fig:qual_RV_E_EGO_DYN_qa}
\end{figure}


\begin{figure}[H]
\centering
\includegraphics[width=\linewidth]{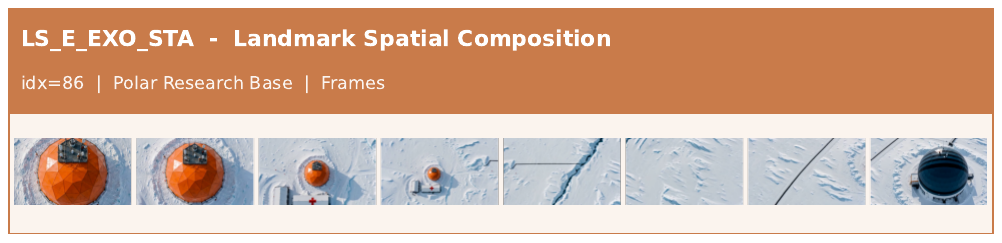}
\caption{\textbf{Frames} for LS\_E\_EXO\_STA, idx 86 (\emph{Polar Research Base}).}
\label{fig:qual_LS_E_EXO_STA_frames}
\end{figure}

\begin{figure}[H]
\centering
\includegraphics[width=\linewidth]{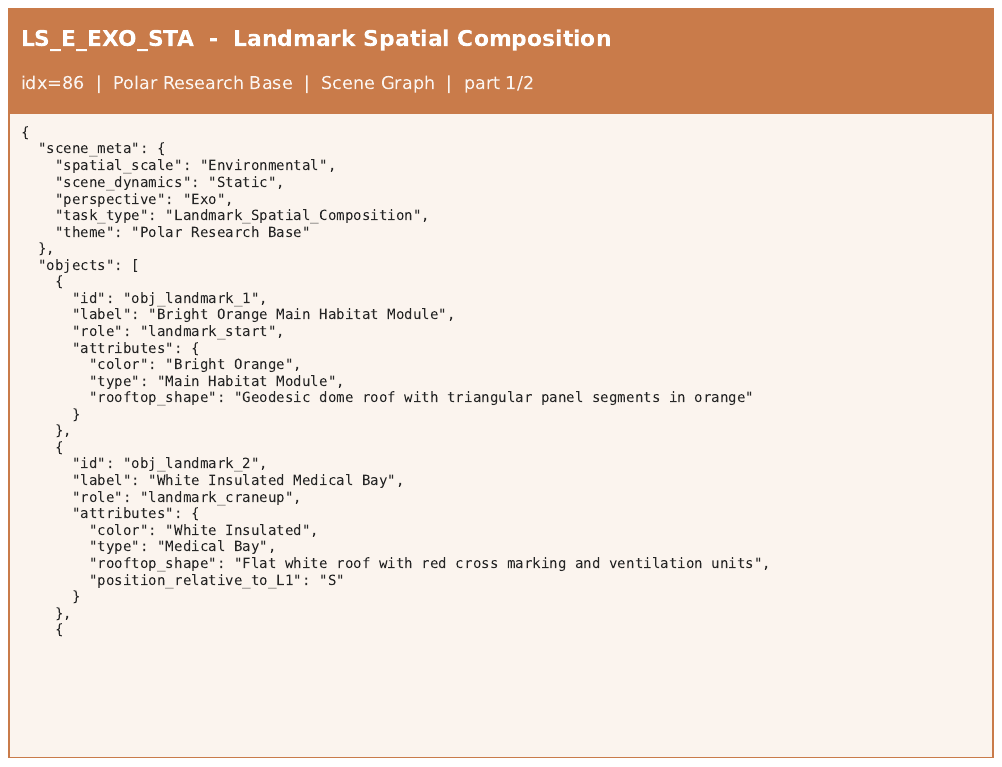}
\caption{\textbf{Scene graph (part 1/2)} for LS\_E\_EXO\_STA, idx 86.}
\label{fig:qual_LS_E_EXO_STA_sg_p1}
\end{figure}

\begin{figure}[H]
\centering
\includegraphics[width=\linewidth]{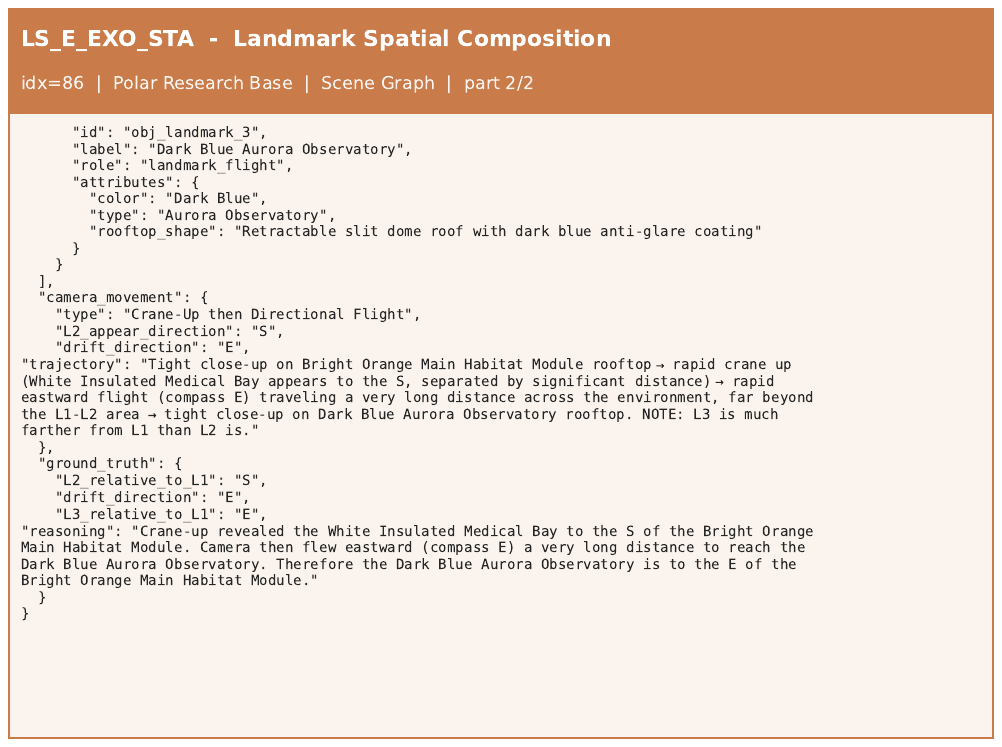}
\caption{\textbf{Scene graph (part 2/2)} for LS\_E\_EXO\_STA, idx 86.}
\label{fig:qual_LS_E_EXO_STA_sg_p2}
\end{figure}

\begin{figure}[H]
\centering
\includegraphics[width=\linewidth]{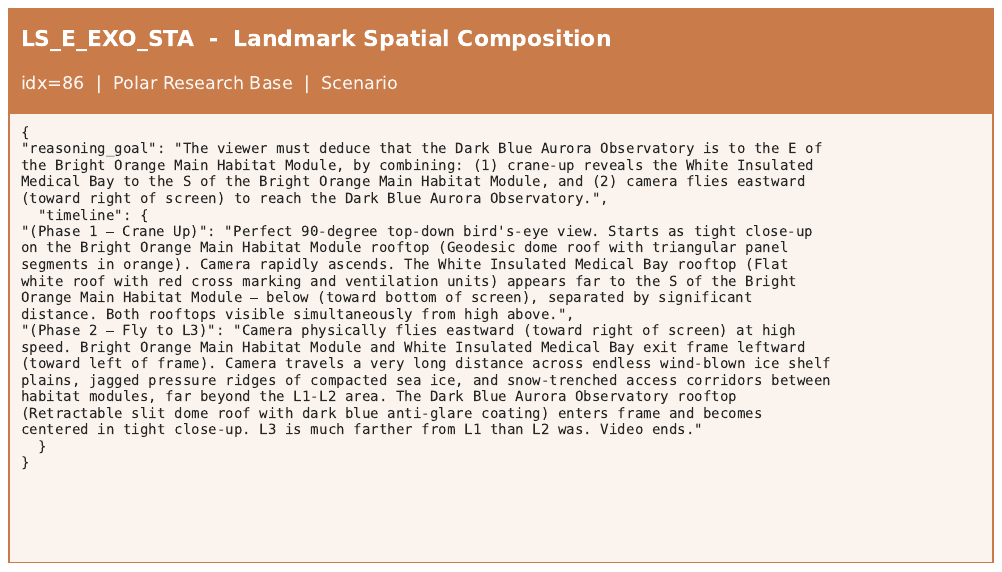}
\caption{\textbf{Scenario} for LS\_E\_EXO\_STA, idx 86.}
\label{fig:qual_LS_E_EXO_STA_scenario}
\end{figure}

\begin{figure}[H]
\centering
\includegraphics[width=\linewidth]{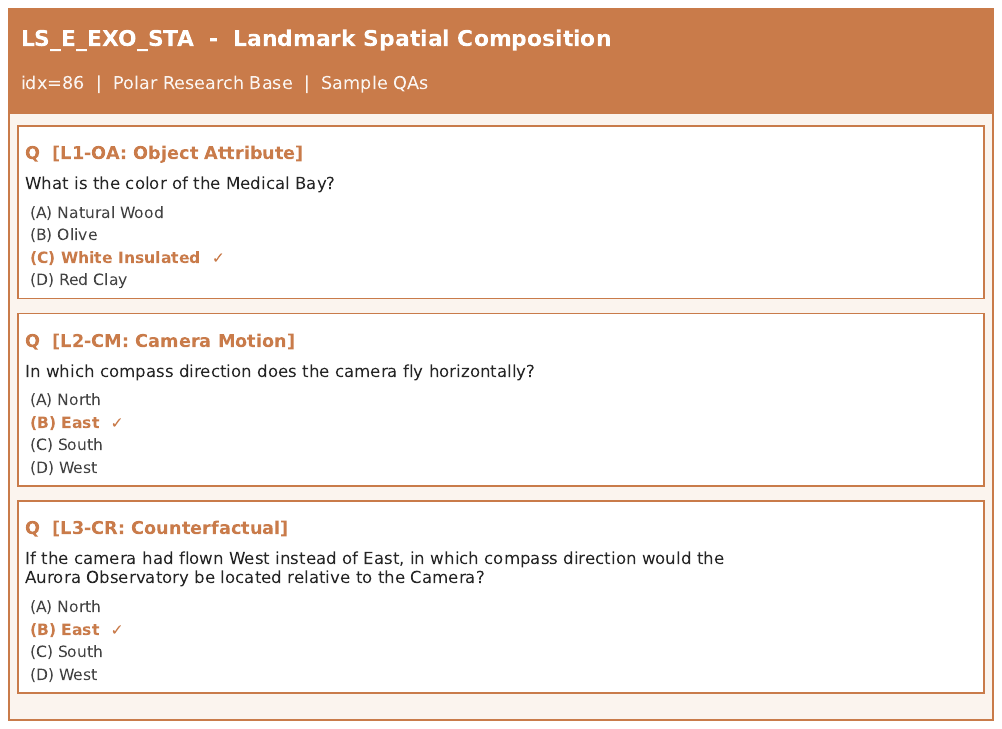}
\caption{\textbf{Sample QA pairs} for LS\_E\_EXO\_STA, idx 86.}
\label{fig:qual_LS_E_EXO_STA_qa}
\end{figure}

\newpage
\begin{figure}[H]
\centering
\includegraphics[width=\linewidth]{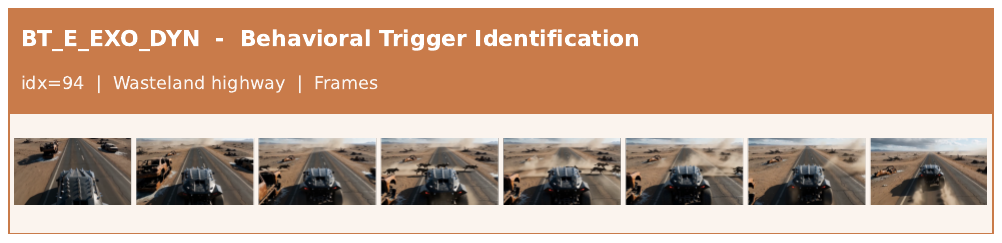}
\caption{\textbf{Frames} for BT\_E\_EXO\_DYN, idx 94 (\emph{Wasteland Highway}).}
\label{fig:qual_BT_E_EXO_DYN_frames}
\end{figure}

\begin{figure}[H]
\centering
\includegraphics[width=\linewidth]{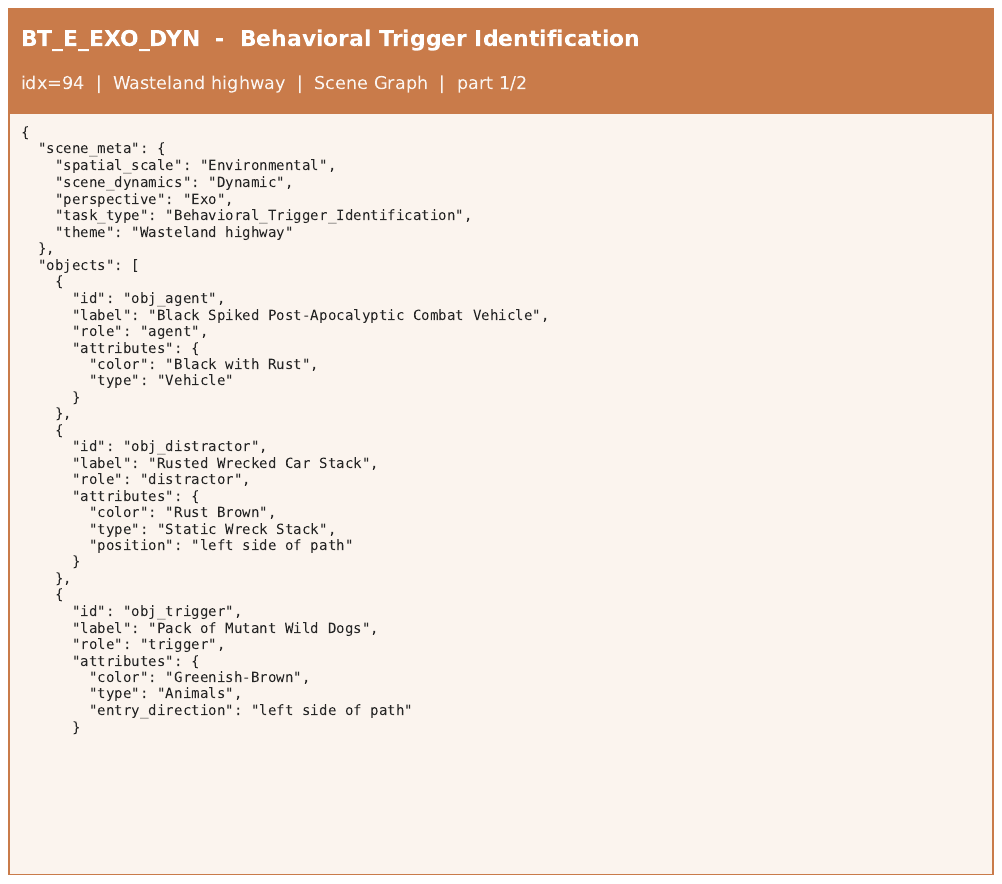}
\caption{\textbf{Scene graph (part 1/2)} for BT\_E\_EXO\_DYN, idx 94.}
\label{fig:qual_BT_E_EXO_DYN_sg_p1}
\end{figure}

\begin{figure}[H]
\centering
\includegraphics[width=\linewidth]{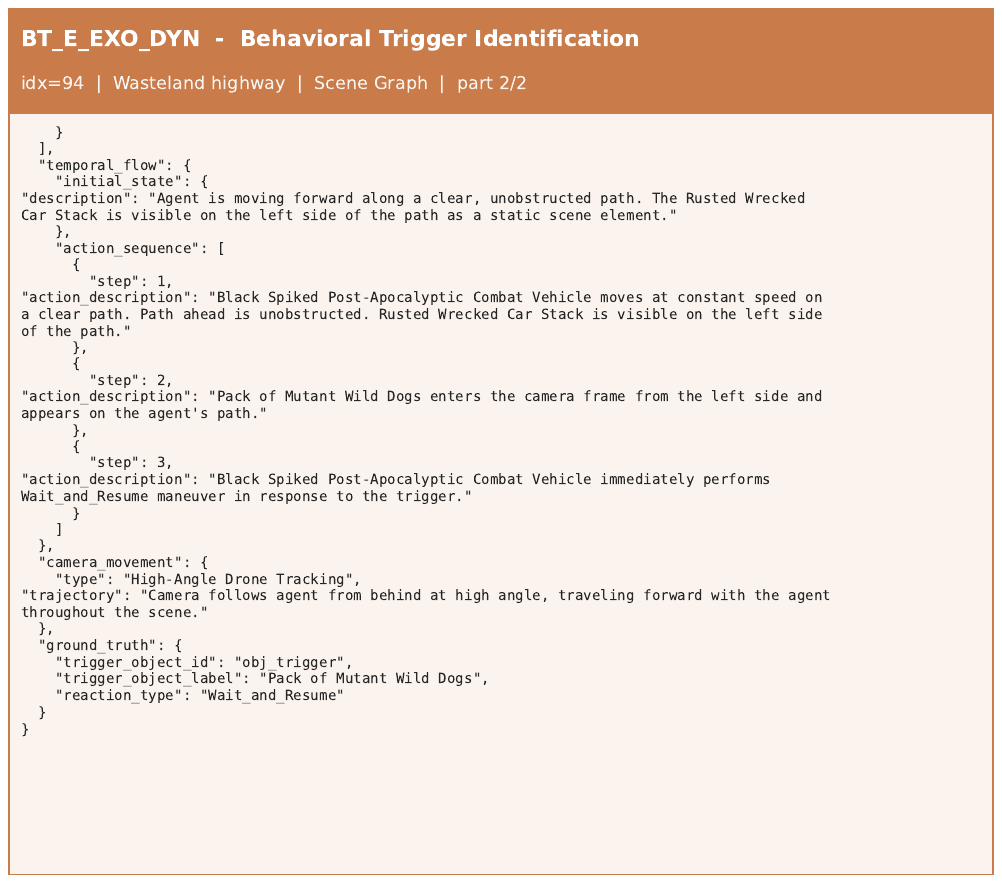}
\caption{\textbf{Scene graph (part 2/2)} for BT\_E\_EXO\_DYN, idx 94.}
\label{fig:qual_BT_E_EXO_DYN_sg_p2}
\end{figure}

\begin{figure}[H]
\centering
\includegraphics[width=\linewidth]{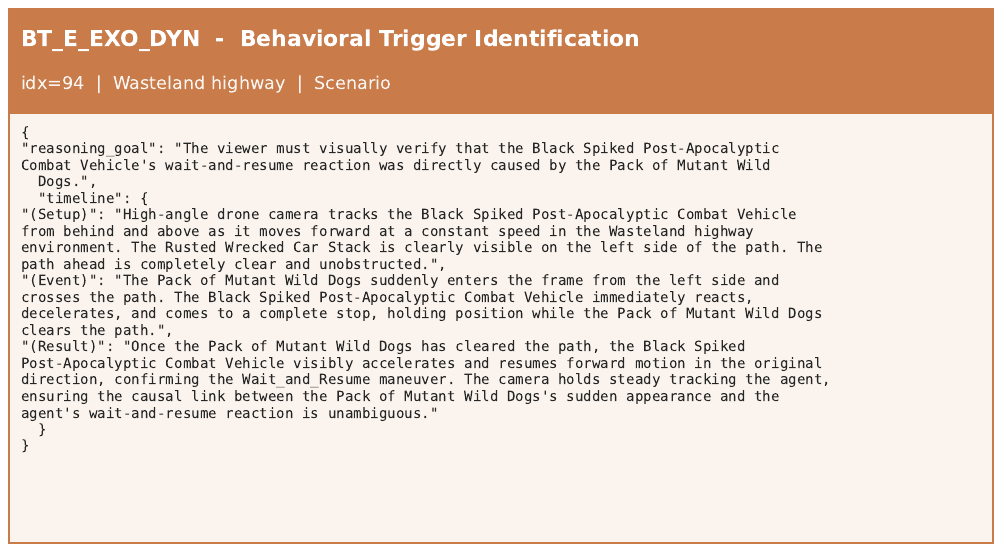}
\caption{\textbf{Scenario} for BT\_E\_EXO\_DYN, idx 94.}
\label{fig:qual_BT_E_EXO_DYN_scenario}
\end{figure}

\begin{figure}[H]
\centering
\includegraphics[width=\linewidth]{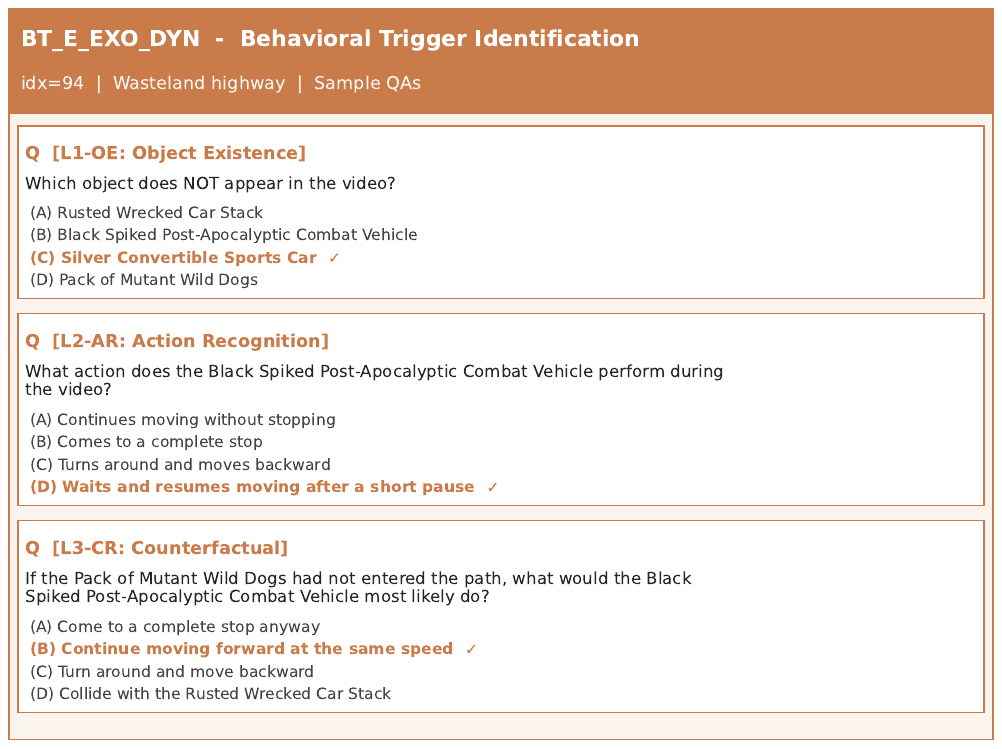}
\caption{\textbf{Sample QA pairs} for BT\_E\_EXO\_DYN, idx 94.}
\label{fig:qual_BT_E_EXO_DYN_qa}
\end{figure}

\end{document}